\documentclass[12pt]{article}
\usepackage{amsmath, amssymb,bm,amsthm}
\usepackage{geometry}

\usepackage{algorithm}
\usepackage{algpseudocode}

\usepackage{hyperref}
\usepackage{url}
\usepackage{comment}
\usepackage{graphicx}
\usepackage{caption}
\usepackage{mathtools}
\usepackage{subcaption}

\usepackage[dvipsnames]{xcolor}

\usepackage[most]{tcolorbox}
\usepackage{booktabs}

\usepackage{pgfplots,tikz}

\numberwithin{equation}{section}

\usepackage[backend=biber, style=ieee, sorting=none]{biblatex}

\AtEveryBibitem{%
  \clearfield{urldate}%
}

\renewbibmacro*{url+urldate}{%
  \printfield{url}%
}

\title{Adaptive Path Integral Diffusion: AdaPID}
\author{Michael (Misha) Chertkov and Hamidreza Behjoo\\  Program in Applied Mathematics \& Department of Mathematics,\\ University of Arizona, Tucson, AZ 87521}
\date{\today}

\begin{document}

\maketitle

\begin{abstract}
Diffusion-based samplers -- Score Based Diffusions, Schrödinger bridges and \emph{Path Integral Diffusion} (PID) -- match a target at terminal time, but the real leverage comes from \emph{choosing the schedule} that governs the intermediate-time dynamics. We develop a pathwise schedule -- selection framework for Harmonic PID with a time–varying stiffness $\beta_t$, exploiting Piece-Wise–constant (PWC) parameterizations and a simple hierarchical refinement. We introduce schedule-sensitive \emph{Quality-of-Sampling} diagnostics -- terminal and time-resolved Wasserstein and cross-entropy, cost-to-go (kinetic/potential) decompositions, velocity- (optimal-control-) gradient sensitivity  based metric, drift–diffusion balance, Langevin mismatch, auto-correlations, energy efficiency, and a speciation transient (reliable decision time). Assuming a Gaussian-Mixture (GM) target (or GM approximation), we retain closed-form Green’s ratios and numerically stable, NN-free oracles for predicted-state maps and scores. Experiments in 2D show that QoS-driven PWC schedules consistently improve early-exit fidelity, tail accuracy, conditioning of the dynamics, and speciation (label-selection) timing at fixed integration budgets.

\end{abstract}

\section{Introduction}\label{sec:intro}
Diffusion models in AI~\cite{ho_denoising_2020,song_score-based_2021} offer a powerful route to sampling from data implicitly modeled by continuous distributions. Most are score-based diffusions (SBDs): they learn time-marginal scores $\nabla_x \log p_t(x)$ and then integrate a reverse Stochastic Differential Equantion (SDE) to transport a simple reference to the target in finite time. This finite-horizon, non-equilibrium view ties modern practice to statistical-mechanics ideas such as Boltzmann generators and related normalizing-flow constructions~\cite{anderson_reverse-time_1982,sohl-dickstein_deep_2015,noe_boltzmann_2019,midgley_flow_2023,zheng_towards_2023,blessing_beyond_2024,sendera_improved_2024,klein_transferable_2024}, where one learns a map (or velocity) that pushes a tractable density to the data.

By contrast, classical Markov Chain Monte Carlo (MCMC)~\cite{metropolis_equation_1953,rosenbluth_genesis_2003} designs time-homogeneous (autonomous) Markov chains with the target density as an invariant law, yielding asymptotic correctness but with mixing-time limitations in multimodal landscapes.

These two lines from tractable density to data and sampling from target density now converge. Recent SBD pipelines~\cite{akhound-sadegh_iterated_2024,he_no_2025} rival MCMC empirically while remaining intrinsically non-autonomous (state- and time-dependent drift). 

In parallel, diffusion bridge versions of SBDs making it even richer mathematically -- linking to Schrödinger bridges~\cite{schrodinger_sur_1932}, Feynman–Kac path integrals~\cite{feynman_space-time_1948,kac_distributions_1949}, and also (with earlier reference to non-autonomous dynamics) to fluctuation theorems (e.g., Jarzynski~\cite{jarzynski_equilibrium_1997,chernyak_dynamical_2005}), therefore situating SBDs within Stochastic Optimal Transport (SOT) paradigm~\cite{peluchetti_non-denoising_2021,behjoo_space-time_2024} where the sampling path is an explicitly designed finite-time transport rather than a stationary chain.

In this paper we build on this diffusion bridge line within the GenAI/Diffusion modeling by continuing our Path Integral Diffusion (PID) program~\cite{behjoo_harmonic_2025}. PID starts from Path Integral Control (PIC)~\cite{kappen_path_2005,kappen_adaptive_2016} and its linearly solvable Markov Decision Process (MDP) generalizations~\cite{e_todorov_linearly-solvable_2007,e_todorov_general_2008,dvijotham_unifying_2012,dvijotham_linearly_2013}. In these settings a class of objectives is \emph{linearly solvable}: the optimal drift admits an integral representation as a ratio of conjugate Green’s functions for a forward/backward Kolmogorov–Fokker–Planck pair. The Harmonic PID (H-PID) of~\cite{behjoo_harmonic_2025} specializes to quadratic  potentials, yielding Gaussian Green’s functions and closed-form expressions; the optimal control becomes a convolution of the target with the Green-function ratio. (See also related \cite{tzen_theoretical_2019} where the case of zero potential was introduced and analyzed and \cite{teter_schrodinger_2024} with a complementary Hermit polynomials based analysis.)

A key insight is that in the limit of infinitely many particles ($M\!\to\!\infty$), exact sampling of a given target is achieved by running a controlled SDE on $[0,1]$ (Section~\ref{sec:setting-stage}). The SDE solves an SOT problem parameterized by a potential, a drift, and a gauge. Here we work in the simplest H-PID case (zero drift and gauge) but allow the harmonic potential to vary in time via a stiffness schedule $\beta_t$ (Eq.~\eqref{eq:L-SOT}). While target densities produced by all such schedules agree in the $M\!\to\!\infty$ and $T\!\to\!\infty$ limits at $t\!=\!1$, they can differ markedly at finite $M$ and time discretization $T$, and -- crucially -- in their transient behavior on $t\in(0,1)$.

This raises the central question of the paper: \emph{How should we evaluate, compare, and ultimately \underline{choose} a schedule $\beta_{0\to 1}$ at finite budgets?} We answer it in two steps. First, Section~\ref{sec:QoS} defines a set of schedule-dependent Quality-of-Sampling (QoS) diagnostics. Second, Section~\ref{sec:experiments} empirically tests these diagnostics. We evaluate protocols using:

\begin{itemize}
\item {Wasserstein metrics} (Section~\ref{sec:Wasserstein}): terminal $W_2$; time-resolved $W_2(t)$ for early-exit quality; tail-restricted $W_2^{(q)}$; and $r$-localized $W_2$.
\item {Cross-entropy family} (Section~\ref{sec:CE}): path cross-entropy for current and predicted states, with conditional (energy-thresholded) variants.
\item {Cost-to-go} (Section~\ref{sec:ODCG}): decomposition into kinetic/potential shares and their time profiles.
\item {Sensitivity of the velocity field} (Section~\ref{sec:grad}): statistics of the optimal-control gradient $\Omega_t$ (op-norm, trace, extremal eigenvalues) and radial scalars.
\item {Drift–diffusion balance} (Section~\ref{sec:drift-diff-balance}): $\kappa^{(\mathrm{s})}$, $\kappa^{(\mathrm{ms})}$, and $\kappa^{(\mathrm{align})}$ time series.
\item {Langevin mismatch} (Section~\ref{sec:langevin-mismatch}): symmetric mismatch $\rho_{\mathrm{sym}}$, cosine similarity, and magnitude ratio versus time.
\item  {Auto-correlations} (Section~\ref{sec:auto-corr}): $A(t)$ and $\hat A(t)$ between intermediate and terminal states.
\item  {Energy efficiency} (Section~\ref{sec:energy-eff}): energy-along-path with a timing regularizer.
\item  {Speciation transient} (Section~\ref{sec:Speciation}): accuracy/risk curves and the CDF of reliable decision times.
\end{itemize}

To keep the analysis transparent we make three pragmatic choices. 
(i) We generalize H-PID by allowing a time-dependent stiffness $\beta_t$ (Sections~\ref{sec:setting-stage},\ref{sec:PID}); 
(ii) we emphasize piece-wise-constant (PWC) protocols for $\beta_t$, which preserve the analytic structure and keep oracle evaluation essentially as cheap as in the constant-$\beta$ case (up to the number of PWC segments); 
(iii) we adopt a Gaussian-mixture (GM) target (or GM approximation), granting a closed-form score and a numerically stable “GM oracle” for validation (Appendix~\ref{sec:G-Mix}). Finally, to validate, calibrate, and visualize the methodology, Section~\ref{sec:experiments} concentrates on two-dimensional targets.%

\section{Setting the Stage: Problem Formulation}
\label{sec:setting-stage}

The Harmonic Path–Integral Diffusion (H-PID) framework of \cite{behjoo_harmonic_2025} seeks a controlled Itô process 
\begin{equation}
\label{eq:x-SODE}
   d x_t \;=\; u_t(x_t)\,dt + dW_t,
   \qquad x_{0}=0^{d}, \qquad t\in[0\to 1],
\end{equation}
where $x_t\in \mathbb{R}^d$; whose terminal law matches a prescribed target density
\begin{equation}\label{eq:tar}
p_{1}(y)=p^{(\text{tar})}(y)=\left(Z^{(\text{\tiny tar})}\right)^{-1} \exp(-E(y)),\quad Z^{(\text{\tiny tar})}\doteq \int \exp(-E(y'))dy',
\end{equation}
where $E(y)$ is the target energy function and $Z^{(\text{\tiny tar})}$ is the target partition function. In a data-driven regime $p^{(\text{tar})}$ is given only via $K$ ground-truth samples $\{y^{(k)}=x^{(k)}_{1}\}_{k=1}^{K}$. However, in this paper we will mainly focus on sampling from a probability distribution density represented via the target energy function and without access to the target partition function. The SODE (\ref{eq:x-SODE}) can also be viewed as representing probability (distribution) density of a path $x_{0\to 1}$, within a path integral framework:
\begin{equation}\label{eq:PI}
p(x_{0\to 1}|u_{0\to 1}(\cdot))\propto \exp\left(-\frac{1}{2}\int_0^1 dt \left(\dot{x}_t-u_t(x_t)\right)^2\right).
\end{equation}

The control field $u_{0\to1}(\cdot)$ -- a.k.a.\ the score function -- is obtained in \cite{behjoo_harmonic_2025} by solving a linearly solvable Stochastic Optimal Transport (SOT) problem that includes a quadratic potential $V(x)=\beta x^2/2$, where $\beta$ was a stiffness constant. A residual degree of freedom in the choice of $\beta$ was noted but was not explored in \cite{behjoo_harmonic_2025}.

In this manuscript we extend H-PID allowing the harmonic potential to vary in time 
\begin{equation}\label{eq:V-t-beta-t}
V_t(x)=\beta_t x^2/2,
\end{equation}
via the stiffness protocol $\beta_t$.

\subparagraph{Adaptive stiffness protocol.} Working within the time-dependent H-PID family of models parameterized by $\beta_{0\to1}\doteq \{\beta_t\}_{t\in[0,1]}$, we show that the Stochastic Optimal Transport (SOT) problem
\begin{align}
\label{eq:L-SOT}
u^{*}_{t}(x;\beta_{0\to 1}) =\text{arg}\min_{u_{0\to1}}
\int_{0}^{1}
\mathbb E
\!\Bigl[
      \tfrac12\|u_t\|^{2} + \frac{\beta_t}{2}\|x_t\|^{2} \Big|\text{Eqs.~(\ref{eq:x-SODE},\ref{eq:tar})}\Bigr]
\,dt,
\end{align}
allows the closed-form solution 
\begin{align}\label{eq:u*}
  u^{*}_{t}(x;\beta_{0\to 1}) & =\nabla_{x} \log\!\left( Z^{(\text{\tiny probe})}(t;x;\beta_{0\to 1})\right),\\ \label{eq:Z}
  Z^{(\text{\tiny probe})}(t;x;\beta_{0\to 1}) & \doteq \int_{\mathbb R^{d}} q^{(\text{\tiny probe})}(y|t;x;\beta_{0\to 1})dy,\\ 
 \label{eq:cond} 
 q^{(\text{\tiny probe})}(y|t;x;\beta_{0\to 1}) &\doteq p^{(\text{\tiny tar})}(y) \exp(-\Delta(t;x;y;\beta_{0\to 1})),\\ 
 \label{eq:Delta} 
 \Delta(t;x;y;\beta_{0\to 1}) & \doteq -\log \frac{G^{(-)}_{t}(x\mid y;\beta_{0\to 1})}{G^{(+)}_{1}(y\mid 0;\beta_{0\to 1})} +C(t;x;\beta_{0\to 1}),
\end{align}
for any $\beta_{0\to 1}$. Here in Eq.~(\ref{eq:cond}) $q^{(\text{\tiny probe})}(y|t;x;\beta_{0\to 1})$ may be interpreted as an un-normalized density of getting $y$ as a target state at $t=1$ conditioned to observing state $x$ at the earlier time $t\leq 1$; and $G^{(\pm)}_{t}$ denote the forward/backward Green Functions (GF) of auxiliary second-order linear PDEs of the Schrödinger equation type (however the dynamics are in real time, not in imaginary time as in quantum mechanics) with the potential $V_t(x)$ parameterized by $\beta_{0\to1}$ -- see Eqs.~(\ref{eq:gen-GF-},\ref{eq:gen-GF+}). The GF equations yield solutions which are Gaussians in  space variables -- $x,y$ -- with time-dependent coefficients which are governed by the Riccati system of Eqs.~(\ref{eq:abcpm}). We also added a $C(t;\beta_{0\to 1})$ function to the definition of $\Delta(t;x;y;\beta_{0\to 1})$ in Eq.~(\ref{eq:Delta}), which makes $\exp(-\Delta(t;x;y;\beta_{0\to 1}))$ normalizable to unity:
\[ \int \exp(-\Delta(t;x;y;\beta_{0\to 1})) dy =1.\]
We discuss details of the PID theory in Section \ref{sec:PID}.

\section{Quality of Sampling (QoS) diagnostics}\label{sec:QoS}

In this section we introduce several schedule–dependent diagnostics for assessing the \emph{quality of sampling} (QoS) under a given $\beta_{0\to 1}$ protocol. Each diagnostic can be read as a candidate metric for comparing protocols.

Conceptually, the diagnostics fall into two groups. The first group consists of \emph{distributional–fidelity diagnostics}, including Wasserstein distance (Section~\ref{sec:Wasserstein}) and cross– entropy– based measures (Section~\ref{sec:CE}). These tests examine both the end-time fit and the manner in which the target distribution emerges over time.

The second group focuses on \emph{dynamical} aspects beyond end-time fidelity: an optimal-transport cost-to-go (Section~\ref{sec:ODCG}); sensitivity of the velocity (optimal control) gradient (Section~\ref{sec:grad}); drift–diffusion balance (Section~\ref{sec:drift-diff-balance}); and \emph{Langevin mismatch}, quantifying departures of PID dynamics from detailed-balance Langevin baselines (Section~\ref{sec:langevin-mismatch}). We also report auto-correlations between intermediate and terminal states (Section~\ref{sec:auto-corr}), an energy-efficiency proxy for dynamical effort (Section~\ref{sec:energy-eff}), and, finally, a \emph{speciation} analysis that asks when component labels in the GM become reliably identifiable (Section~\ref{sec:Speciation}).

\subsection{Wasserstein Distance}\label{sec:Wasserstein}

We compare, \emph{at the terminal time}, the law of the AdaPID terminal state \(x_1\) under the optimal drift induced by a given schedule \(\beta_{0\to 1}\) with the target density \(p^{(\mathrm{tar})}\). Let \(p_1 \equiv p_{X_1}(\cdot\mid u^*;\beta_{0\to 1})\) denote the terminal marginal of the AdaPID dynamics, and let \(p_2 \equiv p^{(\mathrm{tar})}\).
The 2-Wasserstein distance is
\begin{equation}\label{eq:W2}
W_2(p_1,p_2)
\;=\;
\left(\inf_{\pi\in\Pi(p_1,p_2)}\int \|y - y'\|^2 \, d\pi(y,y')\right)^{\!1/2}\!,
\end{equation}
where \(\Pi(p_1,p_2)\) is the set of couplings with marginals \(p_1\) and \(p_2\).

\paragraph{Empirical estimator.}
Let \(\{x^{(m)}_1\}_{m=1}^M \stackrel{\text{i.i.d.}}{\sim} p_1\) (AdaPID terminal samples) and \(\{y^{(m)}\}_{m=1}^M \stackrel{\text{i.i.d.}}{\sim} p_2\) (exact target draws), and define
\(
\hat p_1^{(M)}=\tfrac1M\sum_{m=1}^M \delta_{x^{(m)}_1},\;
\hat p_2^{(M)}=\tfrac1M\sum_{m=1}^M \delta_{y^{(m)}}.
\)
The empirical 2-Wasserstein distance computes the minimum-cost bipartite assignment between 
the two point sets
\begin{equation}\label{eq:W2-emp-balanced}
\hat W_2^2\!\left(\hat p_1^{(M)},\hat p_2^{(M)}\right)
\;=\;
\min_{\sigma\in\mathfrak{S}_M}\;\frac1M \sum_{m=1}^M \bigl\|x^{(m)}_1 - y^{(\sigma(m))}\bigr\|^2,
\end{equation}
where $\mathfrak{S}_M$ denotes the symmetric group on $M$ elements, i.e.,  the set of all permutations $\sigma : \{1,\dots,M\}\to\{1,\dots,M\}$, and thus Eq.~(\ref{eq:W2-emp-balanced}) defines the minimum-cost bipartite matching (Hungarian assignment) with cost \(\|x^{(m)}_1-y^{(k)}\|^2\). Equivalently, in transport-plan form,
\begin{equation}\label{eq:W2-emp-LP}
\hat W_2^2
\;=\;
\min_{\pi\in\mathbb{R}_+^{M\times M}}
\sum_{m,k} \pi_{mk}\,\bigl\|x^{(m)}_1-y^{(k)}\bigr\|^2
\quad\text{s.t.}\quad
\sum_k \pi_{mk}=\frac1M,\;\;
\sum_m \pi_{mk}=\frac1M.
\end{equation}
Our goal in evaluating $\widehat W_2$ across a range of stiffness values $\{\beta_\ell\}$ is to compare the effect of different schedules on the terminal  sampling quality. To reduce variance across protocols, we draw a single reference set $\{y^{(m)}\}_{m=1}^M \sim p_{\mathrm{tar}}$ and reuse it for all $\beta_\ell$ (common–random–numbers). This allows direct, consistent schedule-to-schedule comparison.

\paragraph{Sinkhorn algorithm.} When an exact assignment solver is unavailable or for speed, we use entropically regularized OT with uniform marginals \(r=c=\tfrac{1}{M}\mathbf{1}\) and cost matrix \(C_{mk}=\|x^{(m)}_1-y^{(k)}\|^2\):
\[
\min_{\Pi\ge 0}\;\langle \Pi, C\rangle + \varepsilon\,\mathrm{KL}\!\left(\Pi\,\middle\|\, r c^\top\right)
\quad\text{s.t.}\quad
\Pi\mathbf{1}=r,\;\Pi^\top\mathbf{1}=c.
\]
Writing \(K=\exp(-C/\varepsilon)\), Sinkhorn scaling iterates \(u \leftarrow r/(K v),\; v \leftarrow c/(K^\top u)\) (component-wise division), and returns \(\Pi=\mathrm{diag}(u)\,K\,\mathrm{diag}(v)\). We report \(\hat{W}_{2,\varepsilon}^2=\langle \Pi, C\rangle\) and \(\hat{W}_{2,\varepsilon}=\sqrt{\max\{\hat{W}_{2,\varepsilon}^2,0\}}\). When an exact Hungarian routine is available and \(M\) matches, we switch to it automatically. 

\paragraph{Protocol across $\beta_t$}. Fix a grid \(\{\beta_\ell\}\). Draw once a target reference set \(\{y^{(m)}\}_{m=1}^M\sim p^{(\mathrm{tar})}\) and reuse it for all \(\beta_\ell\) (variance reduction by common random numbers). For each \(\beta_\ell\), simulate \(\{x^{(m)}_1(\beta_\ell)\}_{m=1}^M\) and compute \(\hat W_2\bigl(\hat p_1^{(M)}(\beta_\ell),\hat p_2^{(M)}\bigr)\). We average \(\hat W_2\) over independent simulator seeds and report mean \(\pm\) 95\% confidence intervals. 

\subsubsection{Early-exit Wasserstein} \label{sec:Wasserstein-early}

To quantify anytime sampling quality, we monitor the Wasserstein distance at intermediate times. Let \(p_{X_t}\) be the law of \(x_t\) under AdaPID with schedule \(\beta_{0\to 1}\); define
\begin{equation}\label{eq:W2-time}
W_2(t;\beta)\;\doteq\;W_2\!\bigl(p_{X_t},\,p^{(\mathrm{tar})}\bigr),\qquad t\in(0,1],
\end{equation}
and estimate it empirically as in \eqref{eq:W2-emp-balanced} with \(x^{(m)}_t\) (particles recorded on a fixed midpoint grid) matched to the \emph{same} reference \(\{y^{(m)}\}_{m=1}^M\) for all \(t\) and all \(\beta_t\). 
Two normalizations are useful:
\[
\widetilde{W}_2(t;\beta)\;\doteq\;\frac{W_2(t;\beta)}{W_2(1;\beta)}\quad\text{(shape only)},\qquad
\mathrm{AUC}_{[0,t_\star]}(\beta)\;\doteq\;\int_0^{t_\star} W_2(t;\beta)\,dt\quad\text{(early-exit area)}.
\]
The first compares how \emph{quickly} schedules approach their own terminal quality; the second ranks schedules by overall closeness to the target on \([0,t_\star]\) for a chosen cutoff \(t_\star<1\). In discrete time (midpoint grid \(t_n=(n+\tfrac12)/T\)) we use trapezoidal sums. By default we evaluate strictly inside \((0,1)\) to avoid boundary singularities; our simulator enforces \(x(0)=0\) and uses midpoint drift for stability.

\noindent
\emph{Remark.} Optionally, one may also track \(W_2\!\big(\hat y(t;X_t),\,p^{(\mathrm{tar})}\big)\) to assess how the predicted map \(\hat y\) concentrates toward the target along the flow. 

\subsubsection{Quantile–restricted Wasserstein}\label{sec:Wasserstein-quantile}

To probe fidelity specifically in the low–probability regions of the target, we define a tail domain
\begin{equation*}
\Omega_q \;=\; \bigl\{x:\; p_2(x)\le\tau_q\bigr\},
\qquad
\text{where }\;\tau_q\;\text{ satisfies }\;
\mathbb{P}_{Y\sim p_2}\!\bigl(p_2(Y)\le\tau_q\bigr)=q,
\end{equation*}
i.e., $\tau_q$ is the $q$–quantile of the target density evaluated on a target sample. Given i.i.d. samples
$\{x^{(m)}_1\}_{m=1}^M\sim p_1$ and $\{y^{(m)}\}_{m=1}^M\sim p_2$, let
$I_q=\{m:\,p_2(y^{(m)})\le\tau_q\}$ and $J_q=\{m:\,p_2(x^{(m)}_1)\le\tau_q\}$
be the indices that fall inside the common tail domain $\Omega_q$, which is defined \emph{only} via $p_2$.
Set $n_q=\min\{|I_q|,|J_q|\}$. When $n_q\ge 1$, we form balanced empirical measures on $\Omega_q$ by subsampling without replacement $\tilde I_q\subseteq I_q$ and $\tilde J_q\subseteq J_q$ with $|\tilde I_q|=|\tilde J_q|=n_q$, and define the tail–restricted empirical distance
\begin{equation}\label{eq:W2-emp-quant}
\hat W_{2}^{(q)}
\;\doteq\;
\left(
\min_{\sigma\in\mathfrak{S}_{n_q}}\;
\frac{1}{n_q}\sum_{m=1}^{n_q}
\bigl\|x^{(\tilde J_q(m))}_1 - y^{(\tilde I_q(\sigma(m)))}\bigr\|^2
\right)^{\!1/2}.
\end{equation}
This metric reports the transport cost \emph{restricted to} $\Omega_q$ and is thus a localized “tail fidelity” score.

\subsubsection{Localized Wasserstein}\label{sec:Wasserstein-localization}

To isolate the sampler's performance in the core region of the distribution,  we introduce the $r$-localized Wasserstein metric. For $r>0$, define the  ball $B_r=\{x\in\mathbb{R}^d : \|x\|\le r\}$. Given terminal AdaPID samples $X^{(\beta)}_1=\{x^{(m)}_1\}_{m=1}^M$ and target samples $Y=\{y^{(m)}\}_{m=1}^M$, we restrict to the subsets
\[
X^{(\beta)}_1\!\upharpoonright B_r 
  := \{x^{(m)}_1 : \|x^{(m)}_1\|\le r\}, 
\qquad 
Y\!\upharpoonright B_r 
  := \{y^{(m)} : \|y^{(m)}\|\le r\}.
\]
Let $K=\min\bigl(|X^{(\beta)}_1\!\upharpoonright B_r|,|Y\!\upharpoonright B_r|\bigr)$, and downsample both sets uniformly without replacement to size $K$. The \emph{$r$-localized empirical Wasserstein distance} is then
\[
\widehat W^2_{2,r} 
   := \min_{\sigma\in S_K} 
      \frac{1}{K}\sum_{m=1}^K 
      \bigl\|x^{(\sigma(m))}_1 - y^{(m)}\bigr\|^2 .
\]
Algorithmically: (i) filter samples to $B_r$;  (ii) downsample to equal cardinality; (iii) compute exact or entropic OT between the restricted sets. The diagnostic isolates how well the protocol reconstructs the distribution near its high-density core, which often exhibits smaller model-induced variance than the global Wasserstein distance.

\subsection{Path Cross Entropy}\label{sec:CE}

It is sensible to judge the quality of sampling via the Path Cross Entropy (P-CE):
\begin{align}\label{eq:P-CE}
    {\cal H}^{(\text{\tiny P-CE})}(t;\beta_{0\to 1})
    \;\doteq\;
    -\int p^{(\text{\tiny pred-e})}(y \mid t;\beta_{0\to 1})\,\log p^{(\text{tar})}(y)\,dy,
\end{align}
between the predicted density of the expected state \eqref{eq:pred-expected} and the target density \eqref{eq:tar}.
Given a protocol \(\beta_{0\to 1}\) and \(M\) path samples, P-CE is estimated as
\begin{align}\nonumber
    {\cal H}^{(\text{\tiny P-CE})}(t;\beta_{0\to 1})
    &\approx
    \bar{E}^{(\text{\tiny P-CE})}(t;\beta_{0\to 1};M) + \text{const},\\
\bar{E}^{(\text{\tiny P-CE})}(t;\beta_{0\to 1};M)
& \doteq \frac{1}{M}\sum_{m=1}^M E\!\bigl(\hat y(t;x_t^{(m)};\beta_{0\to 1})\bigr),
\label{eq:P-CE-emp}
\end{align}
where \(\hat y(\cdot)\) is defined in \eqref{eq:y-hat}; the unknown \(\log Z^{(\text{\tiny tar})}\) is absorbed into a \(t\)– and \(\beta_{0\to 1}\)–independent constant.

\paragraph{Energy normalization.}
Throughout, we define the energy up to an additive constant as
\[
E(x)\;=\;-\log p^{(\text{\tiny tar})}(x) + c,
\]
with \(c\) chosen so that $E(x_0=0)=0$.

\paragraph{Conditional Cross Entropies.}
To mitigate the mode-focus of CE, we consider conditional versions that retain only samples with energy above a fixed cutoff tied to the terminal mean C-CE for the same protocol. Specifically, for a given \(\zeta\ge 0\) and protocol \(\beta_{0\to 1}\), define the \(\beta\)-specific threshold
\[
E_c(\zeta;\beta_{0\to 1})
\;\doteq\;
\zeta \cdot \bar{E}^{(\text{\tiny C-CE})}(t\!\approx\!1^{-};\beta_{0\to 1};M).
\]
We then use the \emph{same} \(E_c(\zeta;\beta_{0\to 1})\) across all \(t\) and for the two objects (state and predicted), and form the empirical conditional means
\begin{align}\label{eq:bar-E-C-CE-zeta}
\bar{E}^{(\text{\tiny C-CE})}_\zeta(t;\beta_{0\to 1};M)
& \doteq
\frac{1}{\bigl|{\cal D}^{(\text{\tiny C-CE})}_\zeta(t;\beta_{0\to 1};M)\bigr|}
\sum_{x_t^{(m)}\in {\cal D}^{(\text{\tiny C-CE})}_\zeta(t;\beta_{0\to 1};M)} E\!\bigl(x_t^{(m)}\bigr),\\[-0.25em]
\nonumber
{\cal D}^{(\text{\tiny C-CE})}_\zeta(t;\beta_{0\to 1};M)
&\doteq \Bigl\{x_t^{(m)}\ \big|\ E(x_t^{(m)}) \ge E_c(\zeta;\beta_{0\to 1})\Bigr\},
\\[0.35em]
\label{eq:bar-E-P-CE-zeta}
\bar{E}^{(\text{\tiny P-CE})}_\zeta(t;\beta_{0\to 1};M)
& \doteq
\frac{1}{\bigl|{\cal D}^{(\text{\tiny P-CE})}_\zeta(t;\beta_{0\to 1};M)\bigr|}
\sum_{x_t^{(m)}\in {\cal D}^{(\text{\tiny P-CE})}_\zeta(t;\beta_{0\to 1};M)}
E\!\bigl(\hat y(t;x_t^{(m)};\beta_{0\to 1})\bigr),\\[-0.25em]
\nonumber
{\cal D}^{(\text{\tiny P-CE})}_\zeta(t;\beta_{0\to 1};M)
&\doteq \Bigl\{x_t^{(m)}\ \big|\ E\!\bigl(\hat y(t;x_t^{(m)};\beta_{0\to 1})\bigr) \ge E_c(\zeta;\beta_{0\to 1})\Bigr\},
\end{align}
Typical choices are \(\zeta\in\{0,1,2\}\): \(\zeta=0\) recovers the unconditional average, while larger \(\zeta\) progressively emphasizes higher-energy (rarer) regions.  
\emph{Remark.} One may also adopt a time-dependent cutoff \(E_c(t;\zeta;\beta_{0\to 1})=\zeta\,\bar{E}^{(\text{\tiny C-CE})}(t;\beta_{0\to 1};M)\); we use the terminal definition above to enable direct comparison across \(t\) and across the two CE variants.

The conditional cross-entropy introduced above is a new diagnostic.  Standard CE is dominated by low-energy (high-density) regions of $p_{\mathrm{tar}}$, which masks schedule-dependent discrepancies in moderate- and high-energy regions. Conditioning on an energy threshold $E \ge E_c(\zeta)$ removes these low-energy samples and forces the metric to probe the harder, tail-sensitive portions of the distribution. This idea parallels restricted-likelihood and rare-event diagnostics used in statistics and physics. It therefore provides complementary schedule-sensitivity that the unconditional CE lacks.

\subsection{Cost-to-Go: Potential \& Kinetic Components}\label{sec:ODCG}

Let us define the average Dynamic Cost-to-Go (DCG), also splitting it into potential and kinetic terms
\begin{align}\label{eq:DCG}
   {\cal C}(t;\beta_{0\to 1}) & \doteq {\cal C}^{(\text{\tiny pot})}(t;\beta_{0\to 1})+{\cal C}^{(\text{\tiny kin})}(t;\beta_{0\to 1}),\\ \label{eq:DCG-pot}
   {\cal C}^{(\text{\tiny pot})}(t;\beta_{0\to 1}) & \approx\frac{1}{2 M}\sum_{m=1}^M \int_0^t \beta_{t'} \|x_{t'}^{(m)}\|^{2}dt',\\ \label{eq:DCG-kin} 
    {\cal C}^{(\text{\tiny kin})}(t;\beta_{0\to 1}) & \approx\frac{1}{2 M}\sum_{m=1}^M \int_0^t \|u^{*}_{t'}(x_{t'}^{(m)};\beta_{0\to 1})\|^{2} dt',
\end{align}
where $u^{*}_t(x;\beta_{0\to 1})=b_t^{(-)}\hat{y}(t;x;\beta)-a_t^{(-)}$ (according to Eq.~(\ref{eq:u*-low})); dependence of $a_t^{(-)}, b_t^{(-)}$  on $\beta_{0\to 1}$ is according to Eqs.~(\ref{eq:Kabc-const}) in the case of the const-$\beta$-schedule (see also Section \ref{sec:PWC} for the case of the PWC schedule); and $M$ path samples $x_t^{(m)}$ are generated according to Eq.~(\ref{eq:x-SODE}) with $u$ replaced by $u^*$. We are interested to analyze time dependence and $\beta_{0\to 1}$ dependence of the normalized quantities:
\begin{align}\label{eq:DCG-norm}
   \varsigma (t;\beta_{0\to 1}) & \doteq \varsigma ^{(\text{\tiny pot})}(t;\beta_{0\to 1})+\varsigma ^{(\text{\tiny kin})}(t;\beta_{0\to 1}),\\ \nonumber  \varsigma ^{(\text{\tiny pot})}(t;\beta_{0\to 1}) & =\frac{{\cal C}^{(\text{\tiny pot})}(t;\beta_{0\to 1})}{{\cal C}(t;\beta_{0\to 1})},\quad
   \varsigma ^{(\text{\tiny kin})}(t;\beta_{0\to 1})=\frac{{\cal C}^{(\text{\tiny kin})}(t;\beta_{0\to 1})}{{\cal C} (t;\beta_{0\to 1})},
\end{align}
which we may call kinetic and potential {\bf shares}.

\subsection{Statistics of Optimal Sensitivity (Velocity Gradient)}\label{sec:grad}

As argued in \cite{tsimpos_optimal_2025}, the spatial Lipschitz constant of the velocity field is an important characteristics of the diffusion processes which controls distributional stability and affects numerical approximation/integration error. It is thus prudent to minimize the time-averaged squared gradient of the velocity. According to \cite{tsimpos_optimal_2025}, and follow up stochastic interpolants literature \cite{chen_lipschitz-guided_2025,kunkel_distribution_2025}, schedules which optimizes the objective decelerate or accelerate motion to achieve mass splitting or to handle concentration of the source or target measures, therefore improving the quality of generated images. 

Therefore, projecting these considerations to PID and acting by analogy with statistical hydrodynamics we analyze Lagrangian summaries computed along simulated sample paths $\{x_t\}_{t\in(0,1)}$ driven by the AdaPID dynamics under a chosen $\beta$–protocol. Central to this view is juxtaposition of the state map and  predicted state map visualizing the two components of $x^{(m)}(t)$ and $y(t;x^{(m)}(t))$ respectively.  

It is then natural to introduce the velocity gradient (gradient of optimal control)
\begin{equation}\label{eq:Omega}
    \Omega_t(x)=(\Omega_{t;ij}(x)\doteq \frac{\partial u^{(*)}_i(t;x)}{\partial x_j}|i,j=1,\cdots,d),
\end{equation}
and track its Lagrangian statistics. Specifically, it is of interest to study average norm of $\Omega_t(x_t)$ -- $\mathbb{E}\left[|\Omega_t(x_t)|^2\right]$, its mean trace -- $\mathbb{E}\left[\text{Tr}[\Omega_t(x_t)]\right]$, and its mean max/min eigenvalues.

\paragraph{Exact radial alignment and a scalar gain.} In the case of the low-level integrability discussed in this paper, when $A=f=0$ the optimal control (score function) is a gradient of a scalar function. Then, the optimal sensitivity matrix is symmetric: $\Omega_{ij}=Q_{ji}$. Then  the optimal sensitivity (velocity gradient) matrix, $\Omega_t(x)$, and associated sensitivity matrix of the predicted map, $\hat{\Omega}_t(x)\doteq (\hat{\Omega}_{t;ij}(x)\doteq \frac{\partial \hat{y}_i(t;x)}{\partial x_j}|i,j=1,\cdots,d)$, reduce to scalars
\begin{equation}\label{eq:omega}
\hat{\omega}_t(x)\;=\;\frac{x^\top}{\|x\|}\,
\frac{\partial \hat y(t;x)}{\partial x}\,
\frac{x}{\|x\|},\qquad
\omega_t(x)\;=\;b^{-}_t\,\hat{\omega}_t(x)-a^{-}_t.
\end{equation}

\subsection{Drift–Diffusion Balance}\label{sec:drift-diff-balance}

We work with the unit–diffusion SDE (\ref{eq:x-SODE}
where $u(\cdot,\cdot)$ is substituted by \(u^*(\cdot,\cdot)\) -- the optimal drift induced by the chosen \(\beta_{0\to 1}\)–protocol. All expectations \(\mathbb{E}[\cdot]\) below are with respect to the law of \(x_t\) under this dynamics. Let \(d\) denote the ambient dimension, and let \(\hat y(t;x)\) be the predicted (expected) state map -- given by Eq.~(\ref{eq:hat-x-G-mix-full}) in the case of the Gaussian mixture target density.

We introduce three time-dependent, dimensionless diagnostics that quantify the instantaneous role of the drift relative to diffusion.

\paragraph{(i) Drift to diffusion strength (state space).}
\begin{equation}\label{eq:kappa-state}
\kappa^{({\text{\tiny s})}}(t)\;\doteq\;\sqrt{\frac{\mathbb{E}\,\|u^*(t;x_t)\|^2}{d}}\,.
\end{equation}
The denominator \(d\) is the per–unit–time variance rate contributed by unit Brownian noise across \(d\) coordinates; hence \(\kappa^{({\text{\tiny s})}}(t)\) compares drift speed to the noise scale and is independent of the time step.

\paragraph{(ii) Drift share of mean–square growth (state space).}
By Itô,
\[
\frac{d}{dt}\,\mathbb{E}\|x_t\|^2 \;=\; 2\,\mathbb{E}\!\big[x_t\!\cdot\!u^*(t;x_t)\big] \;+\; d.
\]
Define the drift fraction
\begin{equation}\label{eq:kappa-ms-state}
\kappa^{({\text{\tiny ms})}}(t)\;\doteq\;
\frac{2\,\mathbb{E}[\,x_t\!\cdot\!u^*(t;x_t)\,]}
     {d}\,.
\end{equation}
Values in \((0,1)\) indicate a net positive drift contribution to the instantaneous growth of \(\mathbb{E}\|x_t\|^2\); negative values mean the drift counteracts diffusion.

\paragraph{(iii) Directional efficiency / alignment (state space).}
\begin{equation}\label{eq:kappa-align-state}
\kappa^{({\text{\tiny align})}}(t)\;\doteq\;
\mathbb{E}\!\left[\frac{x_t\!\cdot\!u^*(t;x_t)}{\|x_t\|\,\|u^*(t;x_t)\|}\right],
\end{equation}
with the convention that the ratio is \(0\) if \(\|x_t\|\,\|u^*(t;x_t)\|=0\). This cosine reports how collinear the drift is with the current state: \(1\) (mostly radial), \(0\) (tangential), negative (pulling back).

\subsection{Langevin Mismatch}\label{sec:langevin-mismatch}

We compare the AdaPID drift to the reference (time-homogeneous) Langevin drift built from the target density.  
Let \(p^{(\mathrm{tar})}\) be the target law, and assume the unit–diffusion convention \(dx_t=b(x_t,t)\,dt+dW_t\).  
The AdaPID drift is
\begin{equation}\label{eq:lm-adapid-drift}
u^*(x,t;\beta)\;=\;b_t^{(-)}\,\hat y(t;x)\;-\;a_t^{(-)}\,x,
\end{equation}
while the Langevin drift is
\begin{equation}\label{eq:lm-langevin-drift}
b_L(x)\;\doteq\;\tfrac{1}{2}\,\nabla_x \log p^{(\mathrm{tar})}(x).
\end{equation}
Define the instantaneous discrepancy
\begin{equation}\label{eq:lm-delta}
\Lambda_t(x;\beta)\;\doteq\;u^*(x,t;\beta)\;-\;b_L(x).
\end{equation}

For a given \(\beta\) and time \(t\in(0,1)\), with \(x_t\) distributed as the AdaPID state at time \(t\), we consider the following relative alignment metrics (expectations taken w.r.t.\ \(x_t\)). A small numerical constant \(\varepsilon>0\) is used to avoid division by zero.

\paragraph{(i) Symmetric mismatch.}
\begin{equation}\label{eq:lm-symm}
\rho_{\mathrm{sym}}(t;\beta)
\;\doteq\;
\mathbb{E}\!\left[
\frac{\left\|\Lambda_t(x_t;\beta)\right\|}{\|u^*(x_t,t;\beta)\|+\|b_L(x_t)\|+\varepsilon}
\right]\;\in[0,1].
\end{equation}

\paragraph{(ii) Cosine similarity.}
\begin{equation}\label{eq:lm-cos}
\mathrm{cos}(t;\beta)
\;\doteq\;
\mathbb{E}\!\left[
\frac{u^*(x_t,t;\beta)\cdot b_L(x_t)}{\|u^*(x_t,t;\beta)\|\,\|b_L(x_t)\|+\varepsilon}
\right]\;\in[-1,1].
\end{equation}

\paragraph{(iii) Magnitude ratio.}
\begin{equation}\label{eq:lm-ratio}
r_{\mathrm{mag}}(t;\beta)
\;\doteq\;
\mathbb{E}\!\left[
\frac{\|u^*(x_t,t;\beta)\|}{\|b_L(x_t)\|+\varepsilon}
\right].
\end{equation}

\noindent
Interpretation: \(\rho_{\mathrm{sym}}\) quantifies relative mismatch (0 is best); \(\mathrm{cos}\) measures directional alignment (1 is best); \(r_{\mathrm{mag}}\) compares magnitudes (values near 1 indicate comparable strengths).  
If a different diffusion convention \(dx_t=\sqrt{2}\,dW_t\) is used, \eqref{eq:lm-langevin-drift} should be adjusted accordingly.

\subsection{Auto–Correlation Functions}\label{sec:auto-corr}

Following \cite{behjoo_harmonic_2025}, we monitor two auto–correlations:
\begin{align}
\label{eq:A}
\mathcal{A}(t;\beta_{0\to 1})
&\doteq
\frac{\mathbb{E}_{p^{(\text{\tiny path})}}\!\left[x_t^\top x_1\right]}
     {\mathbb{E}_{p^{(\text{\tiny path})}}\!\left[\|x_1\|^2\right]}
=
\frac{\sum_{m=1}^M \bigl(x^{(m)}_t\bigr)^\top x^{(m)}_1}
     {\sum_{m=1}^M \|x^{(m)}_1\|^2},\\[3pt]
\label{eq:A-hat}
\hat{\mathcal{A}}(t;\beta_{0\to 1})
&\doteq
\frac{\mathbb{E}_{p^{(\text{\tiny path})}}\!\left[\hat{y}\!\left(t;x(t);\beta_{0\to 1}\right)^\top x_1\right]}
     {\mathbb{E}_{p^{(\text{\tiny path})}}\!\left[\|x_1\|^2\right]}
=
\frac{\sum_{m=1}^M \hat{y}\!\left(t;x^{(m)};\beta_{0\to 1}\right)^\top x^{(m)}_1}
     {\sum_{m=1}^M \|x^{(m)}_1\|^2}.
\end{align}

As reported in \cite{behjoo_harmonic_2025}, both \(\mathcal{A}(t;\beta_{0\to 1})\) and \(\hat{\mathcal{A}}(t;\beta_{0\to 1})\) typically evolve monotonically from \(0\) to \(1\) as \(t\) increases, with \(\hat{\mathcal{A}}\) systematically leading \(\mathcal{A}\) (i.e., \(\hat{\mathcal{A}}\) rises earlier in time). In high dimensions, \(\hat{\mathcal{A}}(t;\beta_{0\to 1})\) was observed to undergo a comparatively abrupt change, motivating its interpretation as an order parameter for a dynamic phase transition.

In the low–dimensional setting studied here (\(d=2\) with Gaussian mixture targets), the transition is gradual rather than abrupt. Even so, \(\hat{\mathcal{A}}(t;\beta_{0\to 1})\) remains a practical early indicator of the emergence of target structure and is useful for schedule design -- e.g., as a constraint or regularizer that encourages the transition to occur within a desired time window.

\subsection{Energy Efficiency}\label{sec:energy-eff}

It may be of interest in many physical applications to use the freedom in the schedule \(\beta_{0\to 1}\) to minimize the target energy accumulated along the path, where \(E(x)\doteq -\log p^{(\text{\tiny tar})}(x)\) and it can thus be as well called "Negative Log-Likelihood" (NLL):
\begin{equation}
\label{eq:path-energy}
\beta^{(*;\mathrm{EE})}_{0\to 1}
\;=\;
\arg\min_{\beta_{0\to 1}}
\underbrace{\int_{0}^{1} \mathbb{E}_{p^{(\text{\tiny path})}}\!\left[\,E\!\bigl(x_t\bigr)\,\right]\,dt}_{\displaystyle J_E(\beta_{0\to 1})}
\;\approx\;
\arg\min_{\beta_{0\to 1}}
\frac{1}{T}\sum_{n=0}^{T-1}\frac{1}{M}\sum_{m=1}^{M} E\!\bigl(x^{(m)}_{t_n}\bigr).
\end{equation}

Experimenting with the unregularized criterion \(J_E\) we observed that is minimized by schedules that delay dynamics, favoring largest $\beta$
thus delaying departure from $x=0$  and reducing the objective without yielding meaningful energy efficiency.

To avoid such degenerate schedules, we couple the energy objective to the predicted auto-correlation \(\hat{\mathcal A}(t;\beta_{0\to 1})\) defined in Eq.~\eqref{eq:A-hat}.  Let \(t^\ast(\beta_{0\to 1})\) denote the first time at which \(\hat{\mathcal A}(t;\beta_{0\to 1})\) reaches a target level \(\mathcal A^\ast\in(0,1)\) (we use \(\mathcal A^\ast=\tfrac12\) in our experiments), which serves as a scalar proxy for progress of the predicted map along the flow.  We then penalize overly late transitions relative to a prescribed deadline \(t^{(\mathrm{trans})}\in(0,1)\) 
and minimize the regularized objective
\begin{equation}
\label{eq:energy-regularized}
\beta^{(*;\mathrm{EE\!-\!reg})}_{0\to 1}
\;=\;
\arg\min_{\beta_{0\to 1}}
\Bigl\{
J_E(\beta_{0\to 1})
\;+\;
\lambda_{\mathrm{time}}\bigl(t^\ast(\beta_{0\to 1})-t^{(\mathrm{trans})}\bigr)^2
\Bigr\}.
\end{equation}
This regularizer discourages schedules that merely freeze motion to obtain low energy efficiency, and instead favors energy-efficient schedules that still achieve timely progress as measured by \(\hat{\mathcal A}\).

\subsection{Speciation Transient}\label{sec:Speciation}

As our target is multi-modal, it is natural to ask not only how to sample but also when a trajectory can be reliably assigned a component label. Early in the dynamics such an assignment is ill-defined, but at later times it can emerge abruptly. Following \cite{biroli_dynamical_2024}, we refer to this sharp onset of labelability as the speciation transition. The phenomenon was also observed via the U-turn diagnostic in our earlier work on standard score-based diffusion (SBD) \cite{behjoo_u-turn_2025} and in the H-PID study \cite{behjoo_harmonic_2025}, where we also saw that in low dimensions it appears as a transient rather than a bona fide transition.

Below we extend this analysis by probing the speciation transient sensitivity to the choice of the $\beta_t$ protocol and by searching for schedules that optimize the speciation transient -- i.e., that make reliable labeling occur earlier and more decisively under our verification criteria.

\paragraph{Early label indicator.}
Let the target be a $K$-component mixture \(p^{(\mathrm{tar})}(y)=\sum_{k=1}^K \pi_k\,\mathcal{N}(y\mid\mu_k,\Sigma_k)\). Given a partial state \(x_t\in\mathbb{R}^d\) and its prediction \(\hat y(t;x_t)\), we define the \emph{prediction-based} responsibilities
\begin{equation*} 
r_k^{(\mathrm{pred})}(t;x_t)
\;=\;
\frac{\pi_k\,\mathcal{N}\!\bigl(\hat y(t;x_t)\,\big|\,\mu_k,\Sigma_k\bigr)}
     {\sum_{j=1}^K \pi_j\,\mathcal{N}\!\bigl(\hat y(t;x_t)\,\big|\,\mu_j,\Sigma_j\bigr)}\,,
\qquad \sum_{k=1}^K r_k^{(\mathrm{pred})}=1.
\end{equation*}
From \(r^{(\mathrm{pred})}\) we use three scalars:
\begin{align*}
c(t;x_t)   &\doteq \max_k r_k^{(\mathrm{pred})}(t;x_t), \\
\Lambda(t;x_t) &\doteq r_{(1)}^{(\mathrm{pred})}(t;x_t)-r_{(2)}^{(\mathrm{pred})}(t;x_t), \\
H(t;x_t)   &\doteq -\sum_{k=1}^K r_k^{(\mathrm{pred})}(t;x_t)\,\log r_k^{(\mathrm{pred})}(t;x_t),
\end{align*}
where \(r_{(1)}\ge r_{(2)}\ge\cdots\) are the order statistics. The predicted label is \[\hat\ell_t(x_t)\doteq\arg\max_k r_k^{(\mathrm{pred})}(t;x_t)\].

\paragraph{Reliable decision time.} Fix thresholds \(c_\star\in(0,1)\), \(\Lambda_\star\in(0,1)\), \(H_\star>0\), a stability window \(\tau_w>0\), and a minimum decision time \(\tau_{\min}\in(0,1)\). The reliable decision time for a trajectory is
\[
\hat t^{\,\mathrm{rel}}(x_{\cdot})
\;=\;
\inf\Bigl\{t\ge \tau_{\min}:\;
c\!\ge\!c_\star,\ \Upsilon\!\ge\!\Upsilon_\star,\ H\!\le\!H_\star,\
\hat\ell_s \text{ is constant for } s\in[t-\tau_w,t]\Bigr\}.
\]
If the criteria never hold, we set \(\hat t^{\,\mathrm{rel}}=1\).

\paragraph{Verification protocol.}
The terminal oracle label is \(\ell_1(x_1)\doteq\arg\max_k r_k^{(\mathrm{tar})}(x_1)\)
using the mixture posterior at \(t=1\).
For each constant-\(\beta\) schedule and model, over many trajectories we report:
(i) \textbf{Accuracy vs time}:
\(\mathrm{accuracy}(t)=\mathbb{P}(\hat\ell_t=\ell_1)\);
(ii) \textbf{Risk vs time}:
\(\mathrm{risk}(t)=1-\mathbb{E}[\max_k r_k^{(\mathrm{pred})}(t;x_t)]\);
(iii) \textbf{Decision-time distribution}:
the empirical CDF of \(\hat t^{\,\mathrm{rel}}\).
All figures enforce the minimum decision time \(\tau_{\min}\) and stability window \(\tau_w\).

\section{Path–Integral Diffusion (PID)}
\label{sec:PID}

Main part of this section summarizes the linearly–solvable Path–Integral Diffusion (PID) framework introduced in \cite{chertkov_sampling_2025}; Sec.~\ref{sec:PID-beta-t} then extends it to the time–dependent quadratic potentials required by our adaptive protocol.

\paragraph{From sampling to stochastic optimal transport.} Let $x_t\in\mathbb R^{d}$ evolve according to the controlled SDE (\ref{eq:x-SODE}) and denote by $p_t$ its marginal density. We prescribe $p_0(x)=\delta(x)$ and $p_1(x)\propto \exp[-E(x)]$, and pose the question of infering the score field $u_{0\to1}(\cdot)$. PID casts the search for $u$ as the \emph{stochastic optimal–transport} (SOT) problem Eq.~(\ref{eq:L-SOT} subject to the dynamics~\eqref{eq:x-SODE} and the terminal constraint $p_1\propto e^{-E}$. Throughout this section we focus on the low‐integrability setting of \cite{behjoo_harmonic_2025}, where the gauge and drift vector fields vanish ($A=f=0$) and the potential is isotropic and quadratic, $V_t(x)=\beta x^{2}/2$ with \(\beta>0\).

\paragraph{Linearly solvable structure.} Problem~\eqref{eq:L-SOT} becomes tractable because the optimal score field can be expressed in closed form once two linear PDEs are solved:
\begin{align}
\label{eq:gen-GF-}
-\partial_t G^{(-)}_t + V_t(x)\,G^{(-)}_t &= \tfrac12\Delta_x G^{(-)}_t,
\quad
G^{(-)}_{1}(x;y)=\delta(x-y),\\
\label{eq:gen-GF+}
\partial_t G^{(+)}_t + V_t(x)\,G^{(+)}_t &= \tfrac12\Delta_x G^{(+)}_t,
\quad
G^{(+)}_{0}(x;y)=\delta(x-y).
\end{align}
With $G^{(\pm)}$ at hand we arrive at Eq.~(\ref{eq:u*}). Because the control is recovered from linear PDEs, the problem is said to be linearly solvable \cite{e_todorov_linearly-solvable_2007}.

\paragraph{Constant stiffness: analytic benchmark.} For $V_t(x)=\frac12\beta x^{2}$ with non-negative constant $\beta$ the Green functions are Gaussian and Eqs.~(\ref{eq:gen-GF-}–\ref{eq:gen-GF+}) yield an explicit analytic score \cite{behjoo_harmonic_2025} (see Section \ref{sec:PID-beta-t} for details):
\begin{align}
\label{eq:u*-low} u^{*}_t(x;\beta) &=b_t^{(-)}\hat{y}(t;x;\beta)-a_t^{(-)} x=-a_t^{(-)} \left(x-\tilde y(t;x;\beta)\right),\\
\label{eq:y-hat-const} 
      \hat y(t;x;\beta) & = \mathbb{E}_{y\sim p^{(\text{\tiny probe})}(\cdot;x;\beta)}\left[y\right],\quad p^{(\text{\tiny probe})}(y;x;\beta)\propto \exp\left(-E(y)-\Delta(t;x;y;\beta)\right),
\\ \label{eq:Delta-low}
\Delta(t;x;y;\beta) &=\frac{{\cal K}_t}{2}\left\|y-\frac{b_t^{(-)}}{{\cal K}_t} x\right\|^2-\frac{d}{2}\log\left(\frac{{\cal K}_t}{2\pi}\right),\quad \text{where}
\quad {\cal K}_t  \doteq c^{(-)}_t - a^{(+)}_1,\\  
a_t^{(+)} & = \sqrt{\beta}\coth(t\sqrt{\beta}),\  
a_t^{(-)} \!=\! c_t^{(-)}\!=\! \sqrt{\beta}\coth\!\bigl((1-t)\sqrt{\beta}\bigr),\  
b_t^{(-)} \!=\! \frac{\sqrt{\beta}}{\sinh((1-t)\sqrt{\beta})}. \label{eq:Kabc-const}
\end{align}

Notice that at $\beta\to 0$ Eq.~(\ref{eq:Kabc-const}) becomes
\begin{equation}\label{eq:abc-0}
a_t^{(+)} = \frac{1}{t},\quad  a_t^{(-)} \!=\! c_t^{(-)}\!= b_t^{(-)} \!=\! \frac{1}{1-t}.
\end{equation}

\paragraph{Why time–dependent stiffness?} A fixed $\beta$ cannot simultaneously encourage exploration at early times and enforce contraction near $t=1$. Allowing $\beta_t$ to vary introduces exactly this flexibility. Section~\ref{sec:PID-beta-t} derives the Gaussian Green functions for a general schedule $\beta_{0\to1}$.

\subsection{Time-Dependent Quadratic Potential}\label{sec:PID-beta-t}

Here we continue to discuss the case with $A=f=0$ and isotropic quadratic/harmonic but {\bf time-dependent} potential (\ref{eq:V-t-beta-t}), where $\beta_t\geq 0$. (Some it generalizes to partially negative $\beta_t$ -- see Apendix \ref{sec:negative-beta}.) This case is a bit more general than the "low"-level integrability setting discussed in \cite{behjoo_harmonic_2025} and also briefly reviewed above.  This case is of interest to study because of the intuition gained  on the  benefit of different but time-independent $\beta$\footnote{The intuition -- applied to the goal of building the target distribution discussed in \cite{behjoo_harmonic_2025}  -- suggests that  smaller $\beta$ at the beginning and larger $\beta$ closer to the end may be beneficial for building the target distribution gradually. Indeed, as we start at ${ x}_0=0$ small $\beta$ at early stages will encourage faster spread -- this is the stage of exploration. On the contrary, when you are closer to the end, you want to limit exploration thus increasing $\beta$ and allowing the model to fine-tune -- in this regime the terminal term bias, guaranteeing that we are getting to the target distribution any case will be even stronger. Actually, if true this logic suggests that the largest $t$ do not matter much,  but we need to increase $\beta$ at some intermediate times -- to stabilize memorization --- locking to the final image at the point of phase transition which should probably happen somewhere in the middle of the time interval.}

In this regime of an isotropic but time–dependent quadratic potential the differential relations from Appendix B of \cite{behjoo_harmonic_2025} still apply, except that the constant stiffness $\beta$ is now replaced by the schedule $\beta_t$.  We therefore obtain\footnote{In the following, and when not confusing we drop dependence on $\beta$ to simplify notations.}
\begin{align}
\label{eq:Gpm}
G^{(-)}_t(x;y)&\propto \exp\left(-\tfrac{a^{(-)}_t}{2}x^{2}+b^{(-)}_t x^\top y-\tfrac{c^{(-)}_t}{2}y^{2}\right),\quad G^{(+)}_t(y;0)\propto \exp\left(-\tfrac{a^{(+)}_t}{2}y^{2}\right),
\\[4pt] \label{eq:abcpm}
\mp \dot a^{(\pm)}_t + \beta_t &= \bigl(a^{(\pm)}_t\bigr)^{2},\
\dot b^{(-)}_t= a^{(-)}_t\,b^{(-)}_t,\ 
\dot c^{(-)}_t=(b^{(-)}_t)^{2},
\end{align}
where $\propto$ means that we dropped $t$-dependent multiplier. 

Expressions for $u_t^*(x;\beta_{0\to 1})$, $\hat{y}(t;x;\beta_{0\to 1})$ and $\Delta(t;x;y;\beta_{0\to 1})$ -- originally introduced in Eq.~(\ref{eq:u*-low}), Eq.~(\ref{eq:y-hat-const}) and Eq.~(\ref{eq:Delta-low}) for the case of the constant $\beta$ stay exactly the same in the general case of the time-dependent $\beta_{0\to 1}$, provided we substitute original explicit expressions for $a_t^{(\pm)}, b_t^{(-)}$ and $c_t^{(-)}$ set in Eq.~(\ref{eq:Kabc-const}) by generally an implicit representation via Eqs.~(\ref{eq:abcpm}).

Recall that the Green functions must satisfy the boundary conditions -- \(G^{(-)}_{1}(x;y)=\delta(x-y)\) and \(G^{(+)}_{0}(y;0)=\delta(y)\) -- which results in the following asymptotic requirements: 
\begin{align}\label{eq:-asympt}
    t\to 1^-: & \quad a^{(-)}_t,b^{(-)}_t,c^{(-)}_t\to \frac{1}{1-t},\\ \label{eq:+asympt}
    t\to 0^+: & \quad a^{(+)}_t\to \frac{1}{t}. 
\end{align}

Note that combining Eq.~(\ref{eq:u*-low}) with the $t\to 0$ asymptotics (\ref{eq:-asympt}) and accounting for the fact that $x_0=0$, we estimate that at sufficiently small $t$, $u^{*}_t(x;\beta) \approx b_0^{(-)}\bar{y}$, where $\bar{y}=\mathbb{E}p_1[y]$ is the first moment of the target distribution (which we may assume known).

\subsubsection{Consequences of (time–dependent) self–adjoint generators}
Each instantaneous operator 
\(
\hat L_t=\tfrac12\nabla^2 - V_t(x)
\)
is self–adjoint on $L^2(\mathbb R^d)$. However, when $t\mapsto \hat L_t$ varies in time, the evolution family $U(1,t)$ solving 
$-\partial_t U(1,t)+\hat L_t U(1,t)=0$ with $U(1,1)=I$ is a \emph{time–ordered} product of self–adjoint operators and is not itself self–adjoint unless $\{\hat L_t\}_{t\in[0,1]}$ \emph{commute} (e.g., $\beta_t\doteq\beta$ is constant). Consequently, the backward Green function $G^{(-)}_t(x;y)=[U(1,t)](x,y)$ is, in general, \emph{not} symmetric under $x\leftrightarrow y$.

Under the Gaussian ansatz, this means that one should \emph{not} impose $a^{(-)}_t=c^{(-)}_t$ for a generic time schedule; equality may hold on subintervals where $\beta_t$ and $\nu_t$ are constant (see below), but not globally.

\paragraph{A correct identity for $J_t\doteq \big(a^{(-)}_t\big)^2-\big(b^{(-)}_t\big)^2$.}
From the Riccati system (Eq.~\eqref{eq:abcpm} for the ``$-$'' branch),
\[
\dot a^{(-)}_t=(a^{(-)}_t)^2-\beta_t,\qquad
\dot b^{(-)}_t=a^{(-)}_t\,b^{(-)}_t,\qquad
\dot c^{(-)}_t=(b^{(-)}_t)^2,
\]
one gets
\begin{equation}
\dot J_t
=2\,a^{(-)}_t\big(J_t-\beta_t\big).
\label{eq:J-ode}
\end{equation}
Multiplying \eqref{eq:J-ode} by the integrating factor 
$\exp\!\big(\int_0^t 2a^{(-)}_\tau\,d\tau\big)$
and noting that $\tfrac{d}{dt}\big(\exp(\int_0^t 2a)\,(J-\beta)\big)
=\exp(\int_0^t 2a)\,\big(\dot J - \dot\beta + 2a(J-\beta)\big)$,
we obtain the \emph{correct} linear identity
\begin{equation}
\frac{d}{dt}\,\Bigg(
\exp\!\Big(\!\int_0^t 2a^{(-)}_\tau d\tau\Big)\,
\big(J_t-\beta_t\big)\Bigg)
\;=\;
-\,\dot\beta_t\,
\exp\!\Big(\!\int_0^t 2a^{(-)}_\tau d\tau\Big).
\label{eq:J-intfac}
\end{equation}
Thus,
\begin{equation}
J_t-\beta_t
=
-\,\exp\!\Big(-\!\int_0^t 2a^{(-)}_\tau d\tau\Big)\,
\int_0^t \dot\beta_u\,
\exp\!\Big(\!\int_0^u 2a^{(-)}_\tau d\tau\Big)\,du.
\label{eq:J-solution}
\end{equation}
Two consequences:
\begin{itemize}
\item If $\beta_t$ is \emph{constant on an interval} $I$, then $\dot\beta_t\doteq 0$ on $I$, hence $J_t-\beta$ is constant on $I$. In particular, on the final PWC piece (with constant $\beta_K$ and the terminal asymptotics), one has $J_t\doteq\beta_K$ and therefore $a^{(-)}_t=c^{(-)}_t$ \emph{throughout that last piece}.
\item When $\beta_t$ \emph{changes} (e.g., at a PWC jump), the value of $J_t$ is continuous but the identity $J_t\doteq\beta_t$ \emph{does not persist} into the next piece unless $\beta_{k+1}=\beta_k$. Hence, globally one should not expect $a^{(-)}_t=c^{(-)}_t$.
\end{itemize}

\paragraph{Asymmetry growth.}
It is sometimes convenient to track the difference
$\Delta_t:=a^{(-)}_t-c^{(-)}_t$. From the ODEs above,
\[
\dot\Delta_t
=\dot a^{(-)}_t-\dot c^{(-)}_t
=\big(a^{(-)}_t\big)^2-\beta_t-\big(b^{(-)}_t\big)^2
=J_t-\beta_t.
\]
Combining with \eqref{eq:J-solution} shows explicitly how time-variation in $\beta_t$ \emph{drives} $\Delta_t$ away from zero. On any interval with constant $\beta$, $\dot\Delta_t=0$ and $\Delta_t$ is frozen (equal to its value at the entry to that interval).

\subsubsection{Piece-Wise-Constant $\beta_t$: Analytic Solution}\label{sec:PWC}

This case is advantageous as it allows fully analytic solution. Let us split the time-domain in $K$ equally spaced time intervals and then use the Piece-Wise-Constant (PWC) profile: 
\begin{equation}
    \label{eq:PWC}
    t\in [(k-1)/K,k/K],\ k=1,\cdots,K:\quad \beta_t=\beta_k.
\end{equation}
Then, assuming continuity in time, and taking into account asymptotic behavior of $a_t^{(-)}, b_t^{(-)}, c_t^{(-)}$ at $t\to 1$ documented in Eq.~(\ref{eq:-asympt}) and of $a^{(+)}_t$ at $t\to 0$ documented in Eq.~(\ref{eq:+asympt}) we derive
\begin{align}\label{eq:PWC-K}
& t\in [1-1/K,1]:\  
a_t^{(-)} = c_t^{(-)}= \sqrt{\beta_K}\,\coth\!\bigl((1-t)\sqrt{\beta_K}\bigr),\ 
b_t^{(-)} = \frac{\sqrt{\beta_K}}{\sinh\!\bigl((1-t)\sqrt{\beta_K}\bigr)},\\
\label{eq:PWC-k}
& t\in \bigl[(k-1)/K,\,k/K\bigr],\ k=K-1,\dots,1: \\
\nonumber
&\qquad a_t^{(-)} 
= \sqrt{\beta_k}\,
\frac{\,a_{k/K}^{(-)} + \sqrt{\beta_k}\,\tanh\!\bigl(\sqrt{\beta_k}(k/K - t)\bigr)\,}
{\,\sqrt{\beta_k} + a_{k/K}^{(-)}\,\tanh\!\bigl(\sqrt{\beta_k}(k/K - t)\bigr)\,},\\
\nonumber
&\qquad b_t^{(-)} 
= b_{k/K}^{(-)}\,
\sqrt{\frac{\beta_k - \bigl(a_t^{(-)}\bigr)^2}{\beta_k - \bigl(a_{k/K}^{(-)}\bigr)^2}},\\
\nonumber
&\qquad c_t^{(-)} 
= c_{k/K}^{(-)} 
+ \frac{\bigl(b_{k/K}^{(-)}\bigr)^2}{\beta_k - \bigl(a_{k/K}^{(-)}\bigr)^2}\,
\Bigl(a_{k/K}^{(-)} - a_t^{(-)}\Bigr),\\
\label{eq:PWC-K+}
& t\in [0,1/K]:\quad 
a_t^{(+)} = \sqrt{\beta_1}\,\coth\!\bigl(t\sqrt{\beta_1}\bigr),\\
\label{eq:PWC-k+}
& t\in \bigl[k/K,(k+1)/K\bigr],\ k=1,\dots,K-1: \\
\nonumber
&\qquad a_t^{(+)} 
= \sqrt{\beta_{ k+1}}\,
\frac{\,1 + \rho_{ k+1}(t)\,}{\,1 - \rho_{ k+1}(t)\,},\qquad
\rho_{ k+1}(t)=\exp\!\bigl(-2\sqrt{\beta_{ k+1}}\,(t-k/K)\bigr)\,
\frac{a_{k/K}^{(+)}-\sqrt{\beta_{ k+1}}}{a_{k/K}^{(+)}+\sqrt{\beta_{ k+1}}}.
\end{align}
\emph{Remark on how the formulas are derived and stated.} For the “minus” branch \( \dot a = a^{2}-\beta \), the fractional-linear variable \( u=(a-\sqrt{\beta})/(a+\sqrt{\beta}) \) satisfies \( \dot u = +2\sqrt{\beta}\,u \), which yields the tanh/Möbius form used in \eqref{eq:PWC-k}. For the “plus” branch \( \dot a = \beta-a^{2} \), the same \(u\) obeys \( \dot u = -2\sqrt{\beta}\,u \), giving the exponential Möbius map in \eqref{eq:PWC-k+}.

\section{Numerical Experiments}\label{sec:experiments}

In this section we describe various numerical experiments conducted to understand and leverage the freedom of choosing $\beta_t$ on improving accuracy and efficiency of sampling from the target distribution. The task here is to find characteristics of the scheme which are most sensitive to the choice of the parameters -- primarily $\beta_{0\to 1}$, but also $\nu_{0\to 1}$.

In all the following subsections we will collect statistics by simulating SODE (\ref{eq:x-SODE}). We evaluate all time–dependent quantities at interior midpoints $t_n=(n+\tfrac{1}{2})/T$ to avoid endpoint singularities.

\subsection{Wasserstein Distance}\label{sec:Wasserstein-exp}

\begin{figure}[h!]
    \centering  
\includegraphics[width=\textwidth]{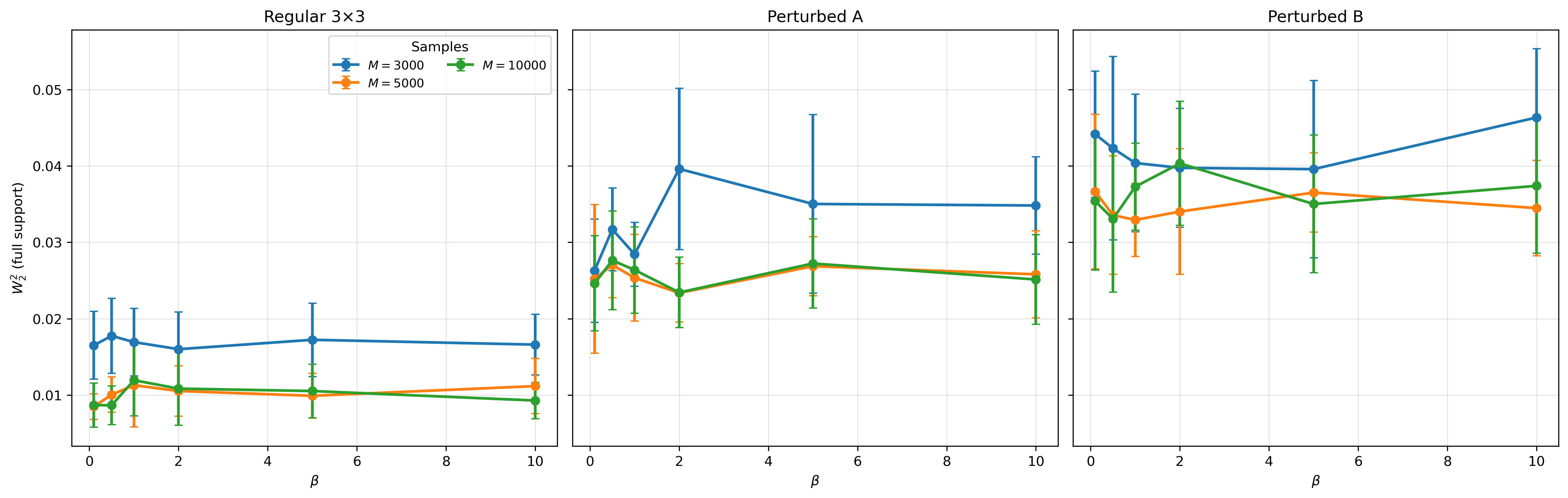}
\includegraphics[width=\textwidth]{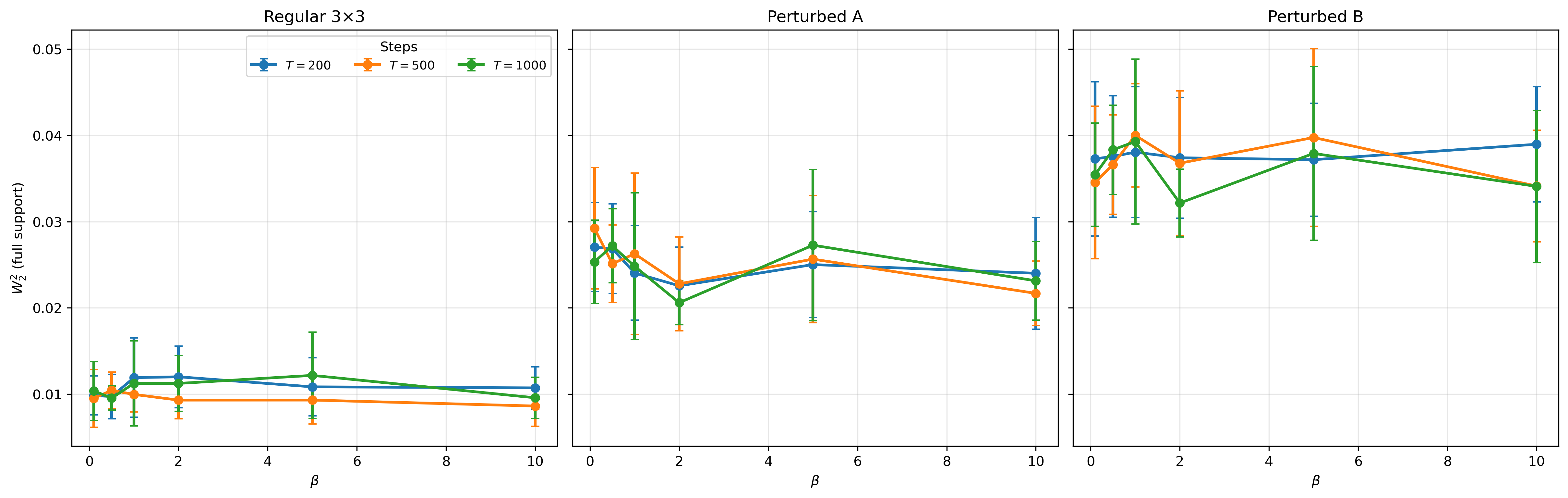}
\includegraphics[width=\textwidth]{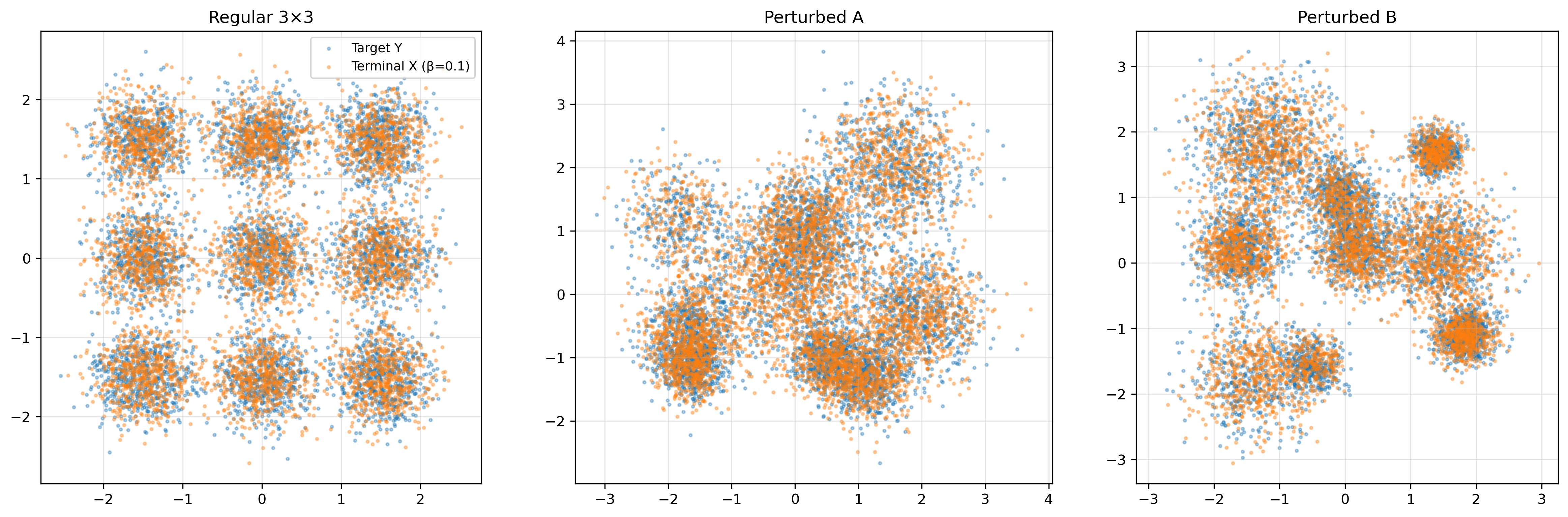}
    \caption{Wasserstein-2 distance (exact earth mover metric) of the AdaPID sampling with const-$\beta$ against the exact sampling  for three models (regular $3\times 3$, perturbed A and perturbed B). We show dependence of the Multi-Seed Mean $\pm$ Standard Deviation (MSM$\pm$SD) over 10 seeds of W2 distance on $\beta$ for fixed $T=500$ (number of time steps) and varying $M=3000, 50000, 10000$ (number of path samples) -- in the top panel --  and vice versa $M=5000$ and $T=200, 500, 1000$ -- in the middle panel. The bottom panel shows scatter plots at $t=1$ -- exact vs AdaPID for the best $\beta$ at $M=5000$ and $T=500$.}
    \label{fig:Wasserstein} 
\end{figure}

Fig.~(\ref{fig:Wasserstein}) shows dependence of Wasserstein-2 distance between AdaPID samples and exact samples in the case of const-$\beta$ on $M$ (number of samples), $T$ (number of time steps) and $\beta$ for experiments with three different Gaussian mixture models (always $9$ modes arranged in a regular 3 $\times 3$ grid, perturbed A and perturbed B in 2D). (See Section \ref{sec:Wasserstein} for relevant definitions and notations.) The object is fragile -- fluctuate a lot from one multi-seed set to another -- thus we show in Fig.~(\ref{fig:Wasserstein}) the Multi-Seed Mean $\pm$ Standard Deviation (MSM$\pm$SD) of the W2 distance -- mean and error bars for SD over 10 seeds. We observed a good quality of the reconstruction --- seen visually in the lower panel of Fig.~(\ref{fig:Wasserstein}), and also in the sufficiently small numbers reported in top and middle panels. The top panel suggests that (a) accuracy and variations decrease with $M$; (b) values vary from model to model (from $\times 2$ to $\times 5$); and (c) best value of $\beta$ -- $\beta^*$ changes from model-to-model and it is also sensitive to the change in both $M$ and $T$, and especially in $M$. Checking the middle panels adds an observation that accuracy does  improve with increase in $T$ the variations, and moreover variations in the Wasserstein distance increases with $T$.

\subsubsection{Tail-Restrictions}\label{sec:tail}

\begin{figure}[h!]
    \centering
\includegraphics[width=\textwidth]{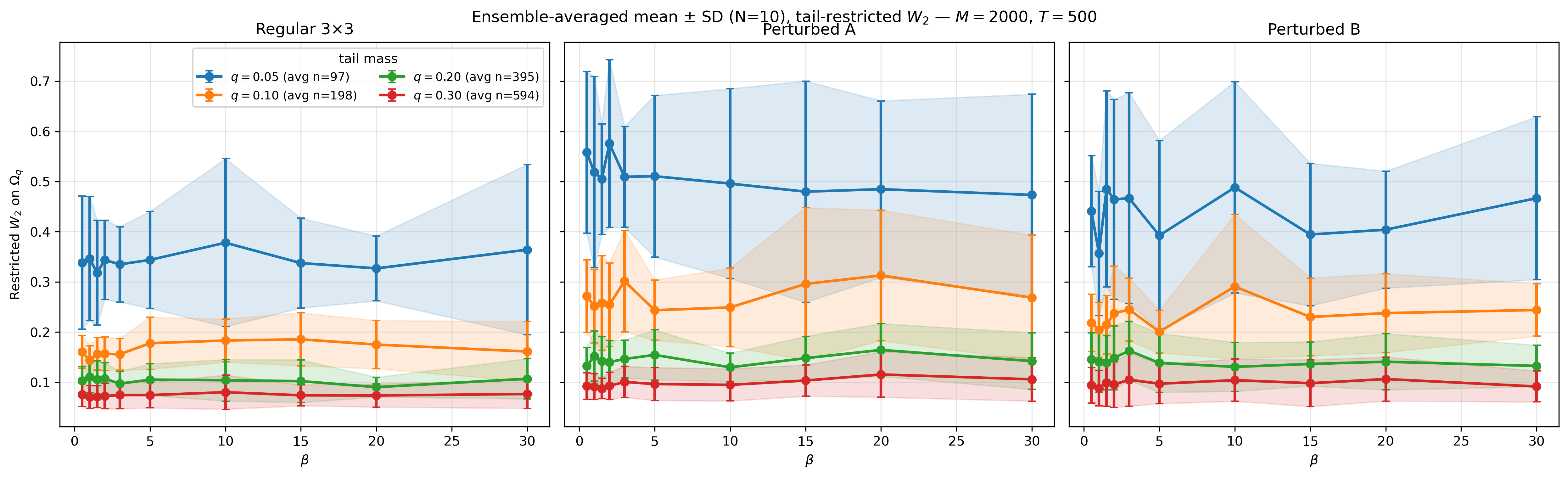}
    \caption{Tail-restricted Wasserstein distance between AdaPID samper with constant $\beta$ and exact sampler for the three GMM models (used across the paper). We plot $\hat W_{2}^{(q)}$ versus $\beta$ for several fixed $q$’s (one curve per $q$), annotated with the average number of matched points $n_q$ -- showing mean and $\pm$ one standard deviation (error bars) showing, as in Fig.~(\ref{fig:Wasserstein}), averaging over (10) multi-seed sets.}
    \label{fig:W2-q}
\end{figure}

We also  experimented with the tail-restricted $W_2^{(q)}$ -- as defined in~\eqref{eq:W2-emp-quant} -- on a grid of $(\beta,q)$ and visualize it in Fig.~(\ref{fig:W2-q}). The figure shows mean and STD over 10 seeds for the regular $3\times 3$ model. We observe, as expected,  that with increase in $q$ -- and thus better statistics (more samples) -- the results stabilize (less fluctuations). We also notice (in the statistics-stabilized regime) some but relatively week dependence on $\beta$.

\subsubsection{Localized Wasserstein}\label{sec:Wasserstein-localization-exp}

Building on the definition in Section \ref{sec:Wasserstein-localization}, we empirically study the \(r\)-localized quadratic Wasserstein distance for \emph{constant} schedules \(\beta_t\equiv\beta\). For a radius \(r>0\), let \(B_r=\{x\in\mathbb{R}^d:\|x\|\le r\}\). Given terminal AdaPID samples \(X_1^{(\beta)}\) and i.i.d.\ target samples \(Y\sim p^{(\mathrm{tar})}\), we form the restricted sets
\(X_1^{(\beta)}\!\upharpoonright B_r=\{x\in X_1^{(\beta)}:\|x\|\le r\}\) and
\(Y\!\upharpoonright B_r=\{y\in Y:\|y\|\le r\}\), down-sample them to a common size \(K\), and compute the empirical \(W_{2,r}^2\) between the two (uniform weights). This realizes the “\(r\)-localized” mismatch posited in Section \ref{sec:Wasserstein-localization}. For comparison we also report the \emph{full} \(W_2^2\) (no radius restriction).

We evaluate three targets: (i) the regular \(3\times 3\) GMM (isotropic, uniform weights), (ii) a perturbed instance A (jittered means, per-component variances, non-uniform weights), and (iii) an independently perturbed instance B. For each model and each \(\beta\in\{0.1,0.5,1,2,5,10\}\), we:
(i) simulate AdaPID to terminal time to obtain \(X_1^{(\beta)}\) (midpoint EM, unit diffusion, \(X(0)=0\)); (ii) draw a large i.i.d.\ target set \(Y\); (iii) for radii \(r\) on a grid, compute \(W_{2,r}^2\bigl(X_1^{(\beta)}\!\upharpoonright B_r,\;Y\!\upharpoonright B_r\bigr)\), provided both restricted sets contain at least a minimum number of points; and (iv) compute the full \(W_2^2\bigl(X_1^{(\beta)},Y\bigr)\). Common random numbers are used across \(\beta\) for fair comparisons.

\begin{figure}[t]
  \centering
  \includegraphics[width=\textwidth]{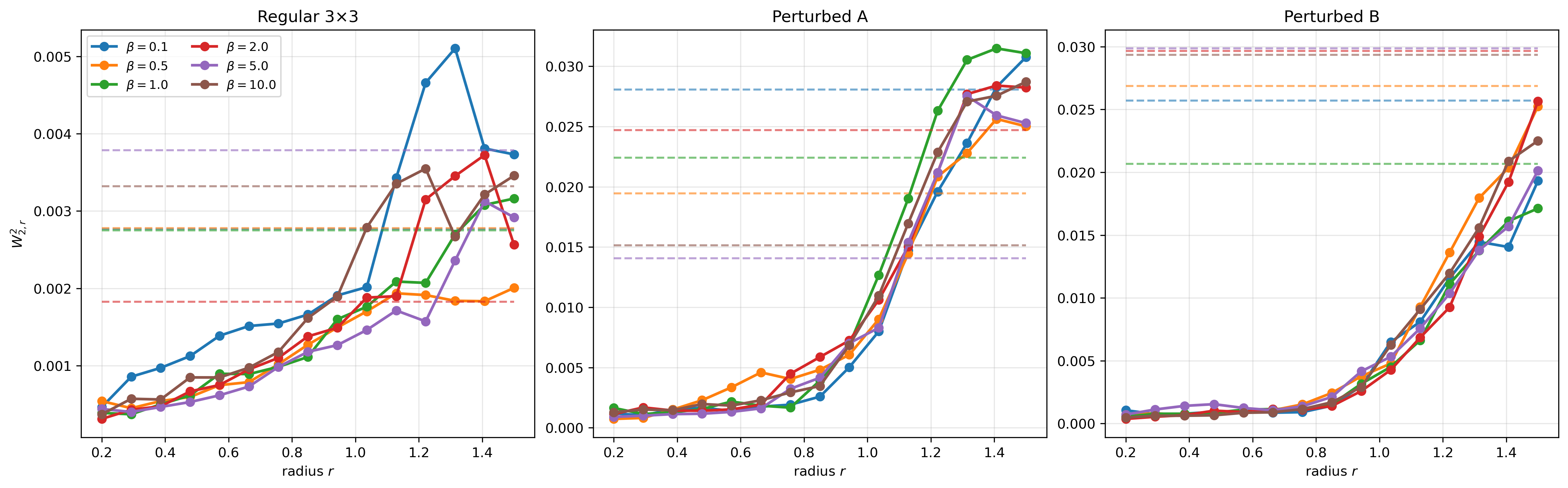}
  \caption{\textbf{\(r\)-localized Wasserstein} (three targets, constant \(\beta\)) with $M=20,000$ (number of sample paths) and $T=2,000$ (number of time steps).  
  Each panel corresponds to one target model (left: Regular \(3\times 3\); middle: Perturbed A; right: Perturbed B).  
  \emph{Solid curves}: \(W_{2,r}^2\) versus radius \(r\) for \(\beta\in\{0.1,0.5,1,2,5,10\}\) (colors).  
  \emph{Dashed horizontal lines}: the corresponding full \(W_2^2\) (no radius restriction) for the same \(\beta\). Construction follows Section \ref{sec:Wasserstein-localization}: at each \(r\) we restrict both AdaPID and target samples to the ball \(B_r\), match them with uniform weights (exact optimal transport), and report the empirical \(W_{2,r}^2\). Missing points indicate insufficient in-ball mass to meet the minimum sample requirement.}
  \label{fig:r-localized-W2-3models}
\end{figure}

Fig.~\ref{fig:r-localized-W2-3models} shows that even for \emph{moderate} radii \(r\), the localized mismatch \(W_{2,r}^2\) is \emph{substantially below} the full \(W_2^2\) and \emph{decreases further} as \(r\) shrinks—consistent with stronger alignment in the distribution’s core. The pattern depends on the target and on \(\beta\), with three robust takeaways: 
\begin{enumerate}
\item \textbf{Reduced \(\beta\)-sensitivity at small \(r\):} curves versus \(\beta\) are markedly flatter inside \(B_r\), indicating greater location stability.
\item \textbf{Different optima:} the constant \(\beta\) minimizing global \(W_2^2\) typically \emph{differs} from the one minimizing \(W_{2,r}^2\).
\item \textbf{Moderate best \(\beta\):} optimal values occur at finite, \(O(1)\) scales—neither the smallest nor the largest tested.
\end{enumerate}
Methodologically, we used larger sample sizes \(M\) to ensure reliable counts inside \(B_r\) and thus probe small (though not vanishing) radii; a side benefit is a smoother, more systematic \(\beta\)-dependence across radii.

\subsection{Cross Entropy} \label{sec:CE-exp}

\begin{figure}[h!]
    \centering
\includegraphics[width=\textwidth]{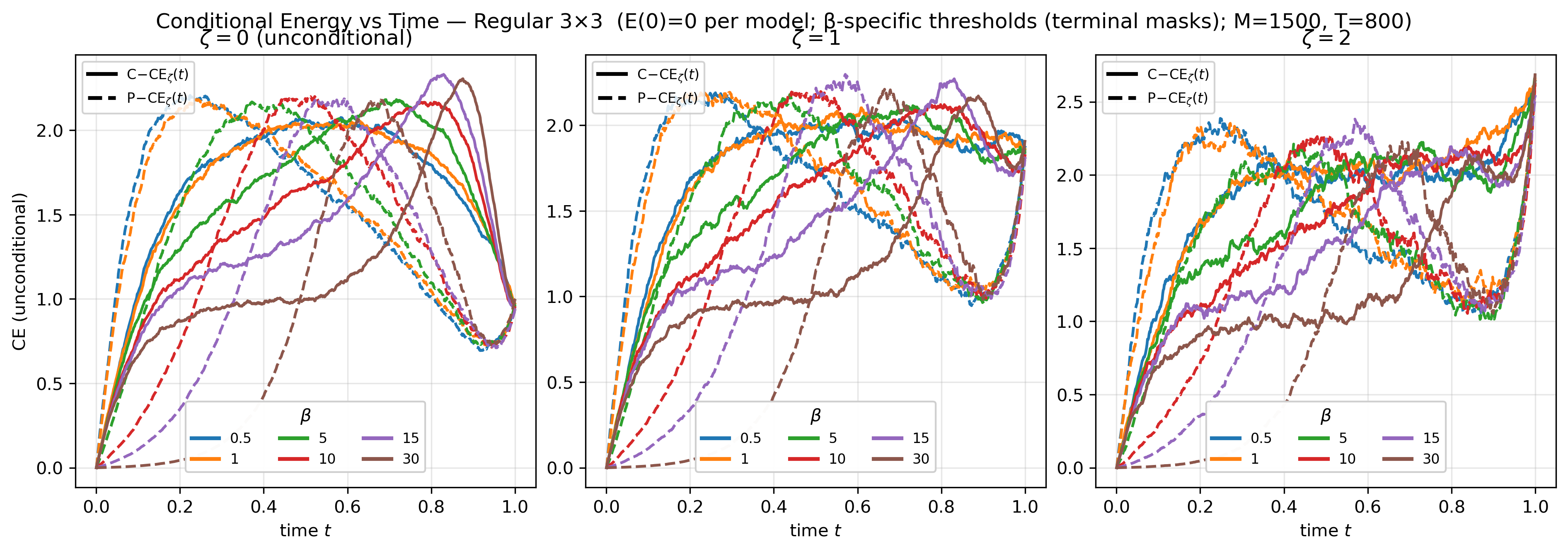}
\includegraphics[width=\textwidth]{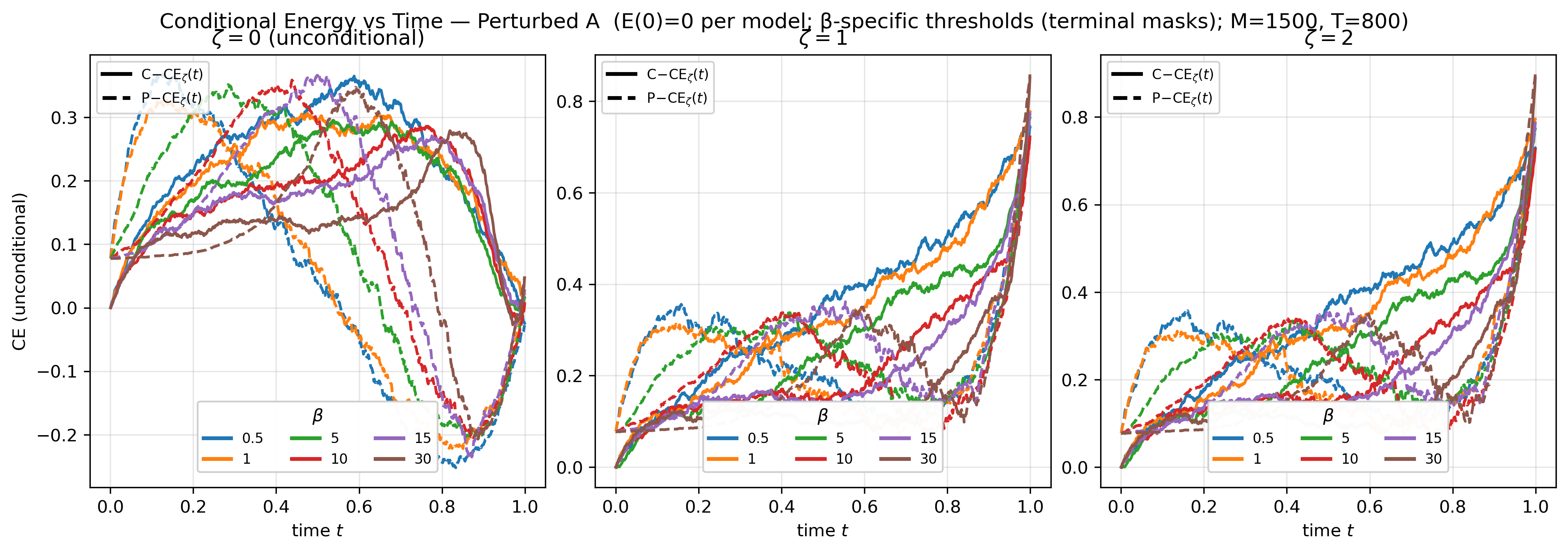}
\includegraphics[width=\textwidth]{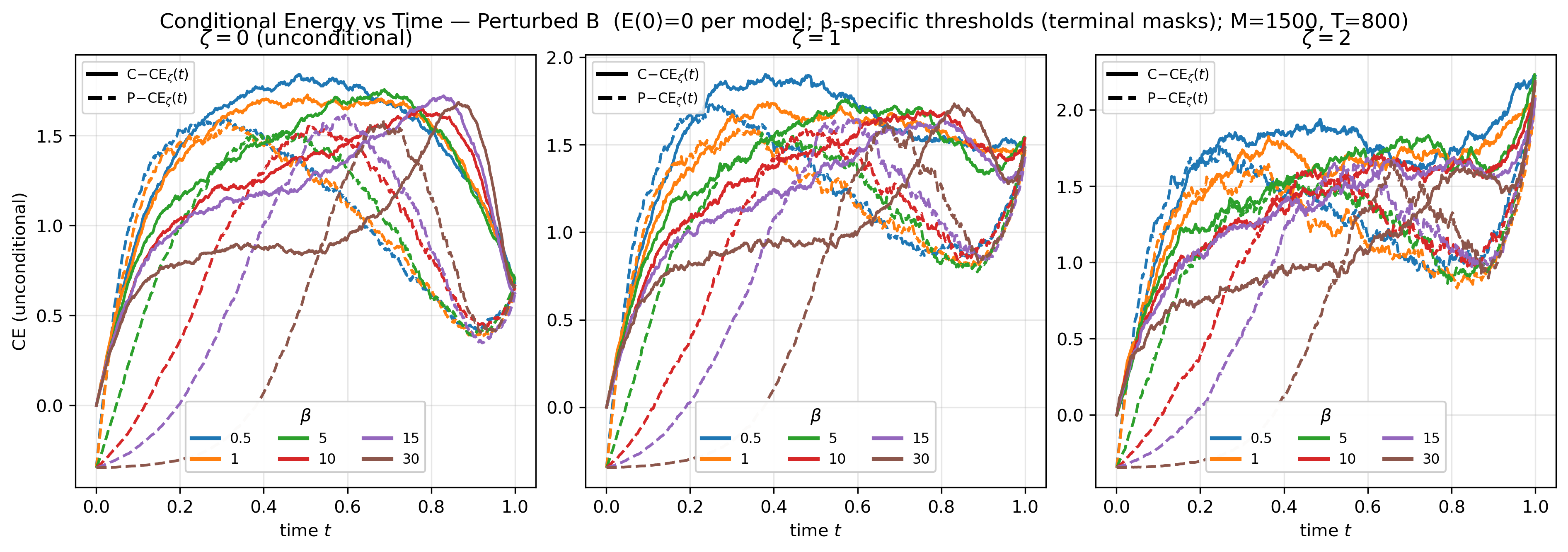}
\caption{Cross–entropy diagnostics with terminal–based conditioning and energy normalization. Three panels correspond to the three GMM models -- regular $3\times 3$, perturbed A and perturbed B. All panels use the normalized energy \(E(x)=-\log p^{(\mathrm{tar})}(x)+c\) with \(c\) chosen so that \(\min_x E(x)=0\). Colors encode the \(\beta\)–protocol; line styles encode the object: solid = \(\mathrm{C\!-\!CE}\), dashed = \(\mathrm{P\!-\!CE}\). Conditioning uses the \emph{terminal} C-CE for each \(\beta\): for a given \(\zeta\ge 0\) we set \(E_c(\zeta;\beta)=\zeta\,\bar{E}^{(\mathrm{C\!-\!CE})}(t\!\approx\!1^{-};\beta)\) and, at every time \(t\), retain only samples whose energy (of the respective object) exceeds \(E_c(\zeta;\beta)\). (a) Unconditional curves (\(\zeta=0\)). (b) Mean-thresholded curves (\(\zeta=1\)). (c) High-energy emphasis (\(\zeta=2\)). Energies are evaluated directly at \(x_t\) and \(\hat y_t\), so the two families (current and predicted) can be compared on the same absolute scale across time and \(\beta\).
}
    \label{fig:CE}
\end{figure}

We observed that the dynamic, i.e. evaluated at $t\in [0,1]$ and not only at $t=1$, Wasserstein distance shows very little sensitivity to the target distribution. This motivates us to return here to testing the dynamic Cross Entropy (CE).

We continue to experiment with the three GMM models, introduced and experimented in the preceding Section, testing their dependence on const-$\beta$. Here we follow the setup of Section~\ref{sec:CE} and study temporal evolution of the CE, evaluated for current $x_t$ and predicted, $\hat{y}(t;x_t)$, states as they develop. Results are shown in Fig.~\ref{fig:CE}: the three panels correspond to the three exemplary GMMs.

Because $x(0)=0$ and $E(0)=0$ (later as our "energy calibration" choice), all the current curves start at zero. As for the predicted curves is concerned, they also start at zero in the case of regular $3\times 3$ model, for which $x=0$ is one of arg-mins of $\min_x E(x)$. However, the latter condition does not hold in the case of perturbed A and perturbed B models,  thus leading to the predicted values showing non-zero energy already at $t=0^+.$

\emph{Unconditioned case:} We observe that generically (across the models) the current cross entropy (C-CE) rises from $0$ to a peak at an intermediate time; both the peak height and its time location increase with $\beta$. For small $\beta$ the C-CE curve is predominantly concave, while for larger $\beta$ an inflection appears and a convex mid–time segment emerges. In contrast, the predicted curves (P-CE) show a different pattern: a rapid early rise, a broad maximum, a shallow late–time dip that is {\bf nearly $\beta$–invariant}, and then approach to the terminal value from below. The  minimum observed in the predicted curves at $t\gtrsim 0.95$ and {\bf nearly $\beta$–invariance} -- collapse of all the different $\beta$-curves -- observed at $t\gtrsim 0.95$, reinforces the observation that an early termination of sampling may be advantageous from the prospective of specific downstream tasks (specific expectations, e.g. expected energy in the case under investigation).  

\emph{Mean–thresholded} and \emph{high–energy} conditioning: Increasing $\zeta$ induces a vertical shift upward (low–energy samples are discarded), while preserving the overall time profiles and the ordering across $\beta$. As expected, variability increases with $\zeta$ because fewer paths/samples contribute to the averages. Qualitative features seen in the unconditioned panel -- such as the mid–time peak of C-CE and the late–time dip of P-CE -- remain visible, merely at elevated levels of $E(\cdot)$.

Overall, we observe that CE is quite sensitive to $\beta$ -- even in the relatively simple const-$\beta$ case the resulting behavior is reach. Some of the special $\beta$-sensitive features -- such as (a) the maximum value of current CE, (b) transition from concave to concave-convex behavior of the current CE with increase of $\beta$, or (c) localization of the predicted CE maximum at a particular time -- can be further utilized to the choice of "optimal" $\beta$.

\subsection{Optimal Dynamic Cost-to-Go} \label{sec:ODCG-exp}

\begin{figure}[h!]
    \centering
\includegraphics[width=\textwidth]{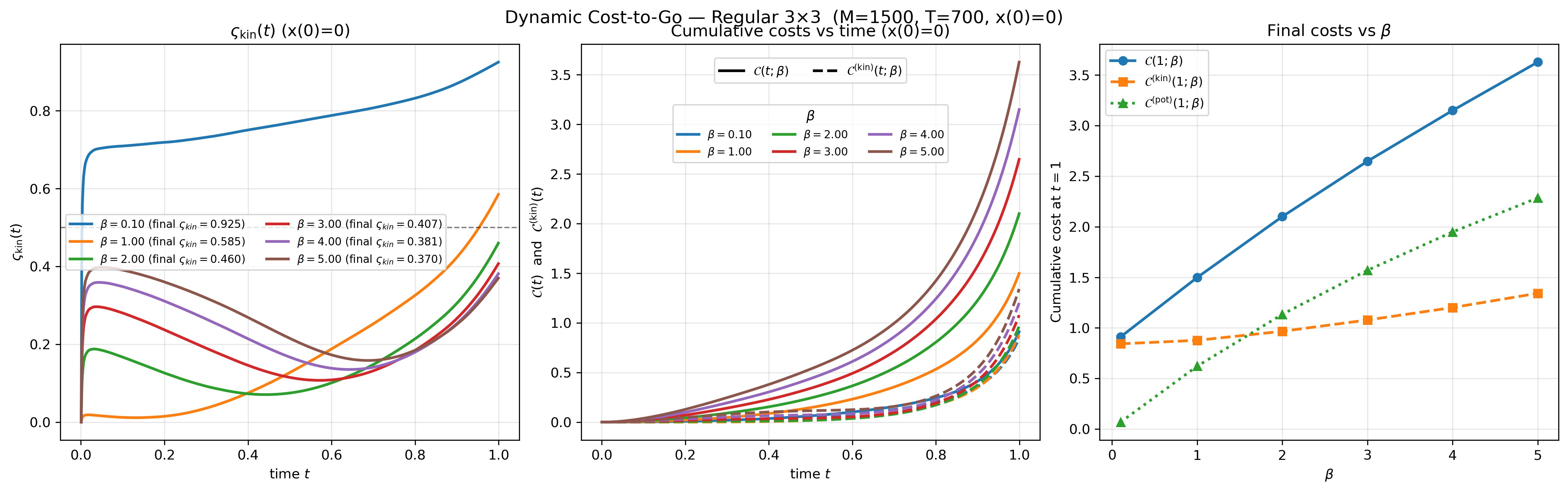}
\includegraphics[width=\textwidth]{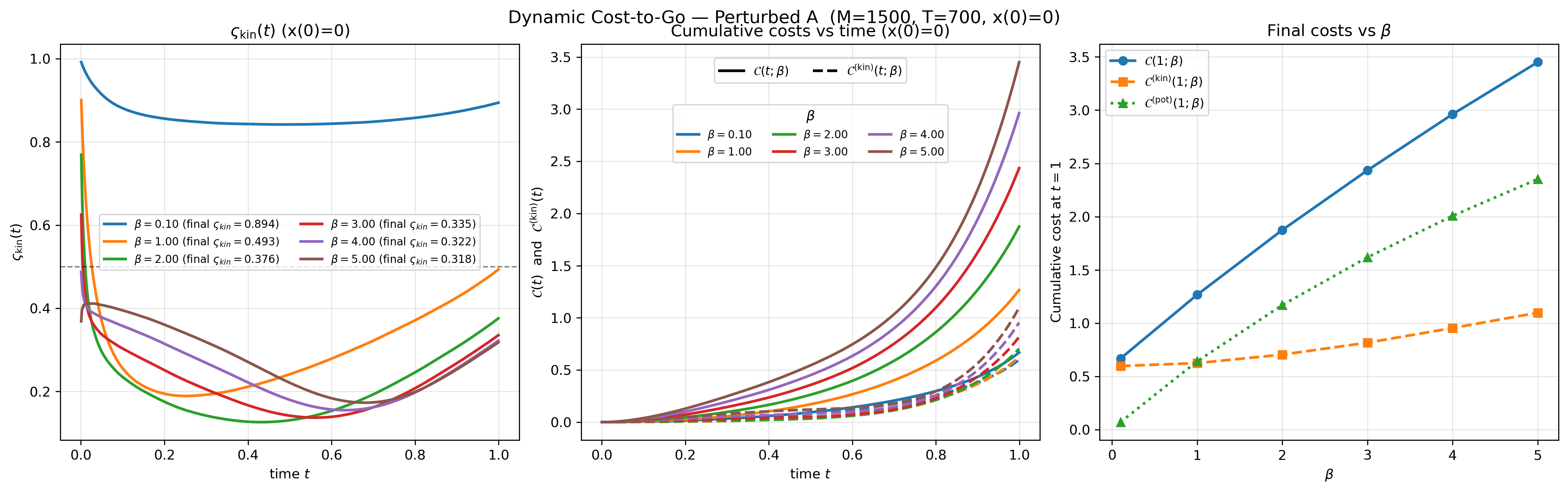}
\includegraphics[width=\textwidth]{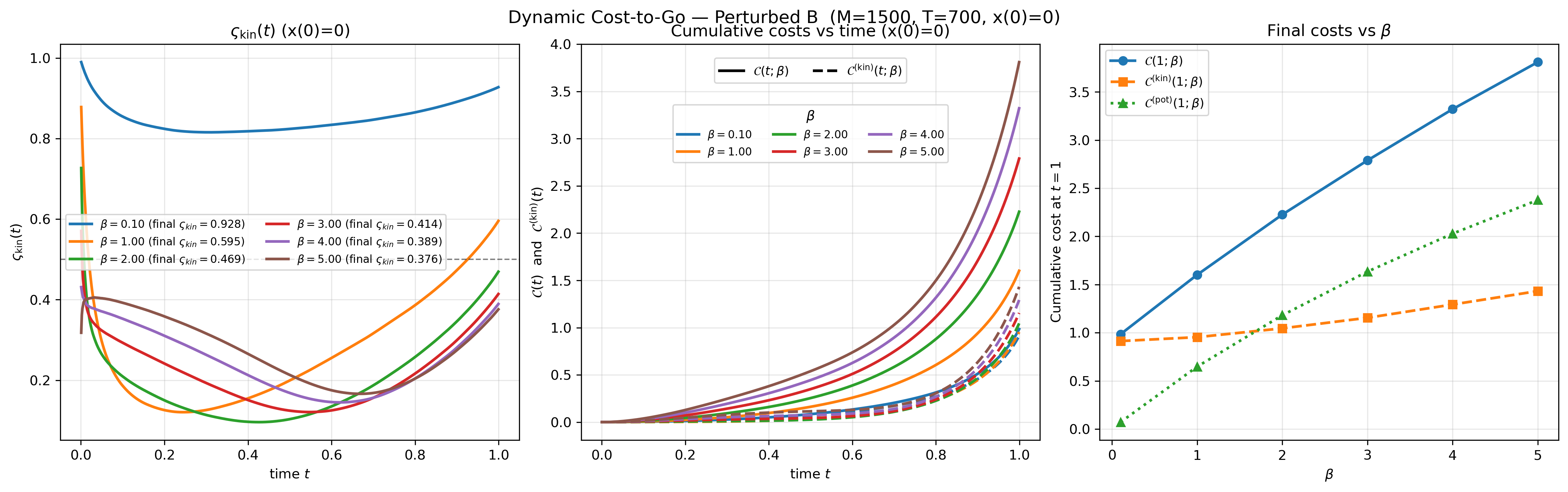}
    \caption{
    Dynamic cost-to-go diagnostics for the Regular 3$\times$3 (top), Perturbed A (middle) and Perturbed B GMMs for varying const-$\beta$. In each row -- Left: kinetic budget share $\varsigma_{\mathrm{kin}}(t)=\dfrac{\mathcal{C}^{(\mathrm{kin})}(t)}{\mathcal{C}^{(\mathrm{kin})}(t)+\mathcal{C}^{(\mathrm{pot})}(t)}$ (Eq.~(\ref{eq:DCG-kin}) across several $\beta$. Center: cumulative costs over time for each $\beta$: total $\mathcal{C}(t;\beta)=\mathcal{C}^{(\mathrm{kin})}(t;\beta)+\mathcal{C}^{(\mathrm{pot})}(t;\beta)$ (solid) and kinetic $\mathcal{C}^{(\mathrm{kin})}(t;\beta)$ (dashed) -- see Eq.~(\ref{eq:DCG}). Right: terminal costs versus $\beta$: $\mathcal{C}(1;\beta)$, $\mathcal{C}^{(\mathrm{kin})}(1;\beta)$, and $\mathcal{C}^{(\mathrm{pot})}(1;\beta)$.}
    \label{fig:varsigma}
\end{figure}

Fig.~(\ref{fig:varsigma}) follows the material of Section \ref{sec:ODCG} (see also Eqs.~(\ref{eq:DCG},\ref{eq:DCG-pot},\ref{eq:DCG-kin},\ref{eq:DCG-norm})) and visualizes dependence of the kinetic part of the re-scaled Dynamic Cost-to-Go (DCG), $\zeta^{(\text{\tiny kin})}$ defined in Eq.~(\ref{eq:DCG-norm}) on time at different values of $\beta$ (for the constant $\beta$ protocol). 

We observe that dependence of the kinetic share $\varsigma_{\mathrm{kin}}(t)$ changes significantly from model-to-model, with $\beta$ and also with time. We observe that at the earliest times and small $\beta$ the kinetic share is the largest. However, even at the earliest times the kinetic share decreases with $\beta$ increase. A fully balanced terminal situation with $\varsigma_{\mathrm{kin}}(1)=1/2$ is achieved at an intermediate and model-dependent $beta$.  In other words -- if we want to neutralize the kinetic-to-potential balance  with the total cost of control, a balanced $\beta$ may be preferable.

On the other hand the terminal cost of ``control" (seen in the right panels of Fig.~(\ref{fig:varsigma})) grows with const-$\beta$ for all the models. This means that -- considered within the const-$\beta$ framework -- the minimal cost is achieved at $\beta=0$. 

The later observation prompt further experiments with Piece-Wise-Constant (PWC) $\beta$ discussed in the following.

\subsubsection{Optimal Kinetic Cost-to-Go for PWC-$\beta$} 

\begin{figure}[t]
  \centering
  \begin{subfigure}[t]{0.34\textwidth}
    \centering
    \includegraphics[width=\linewidth]{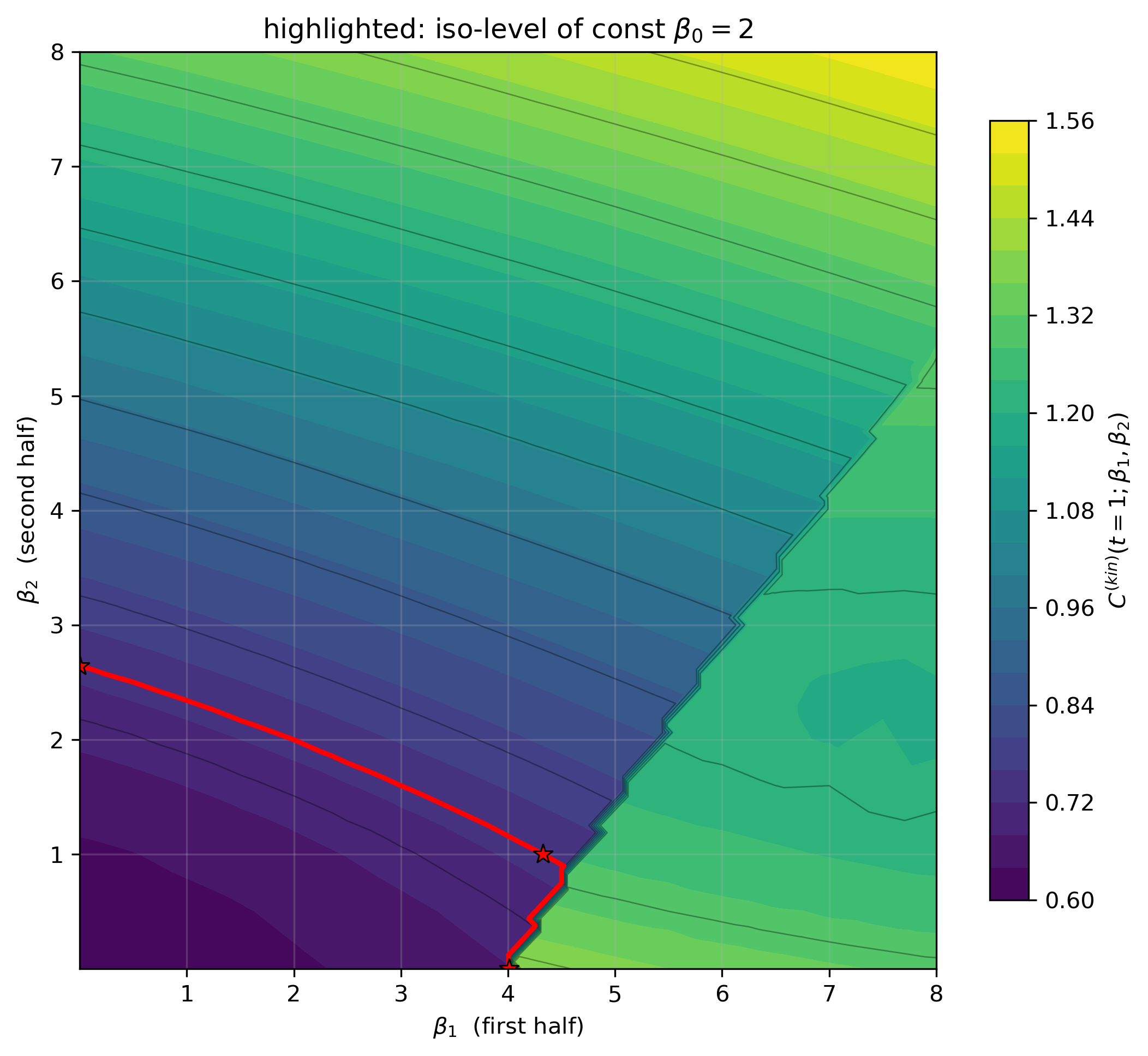}
    \label{fig:isocost-map}
  \end{subfigure}\hfill
  \begin{subfigure}[t]{0.64\textwidth}
    \centering 
    \includegraphics[width=\linewidth]{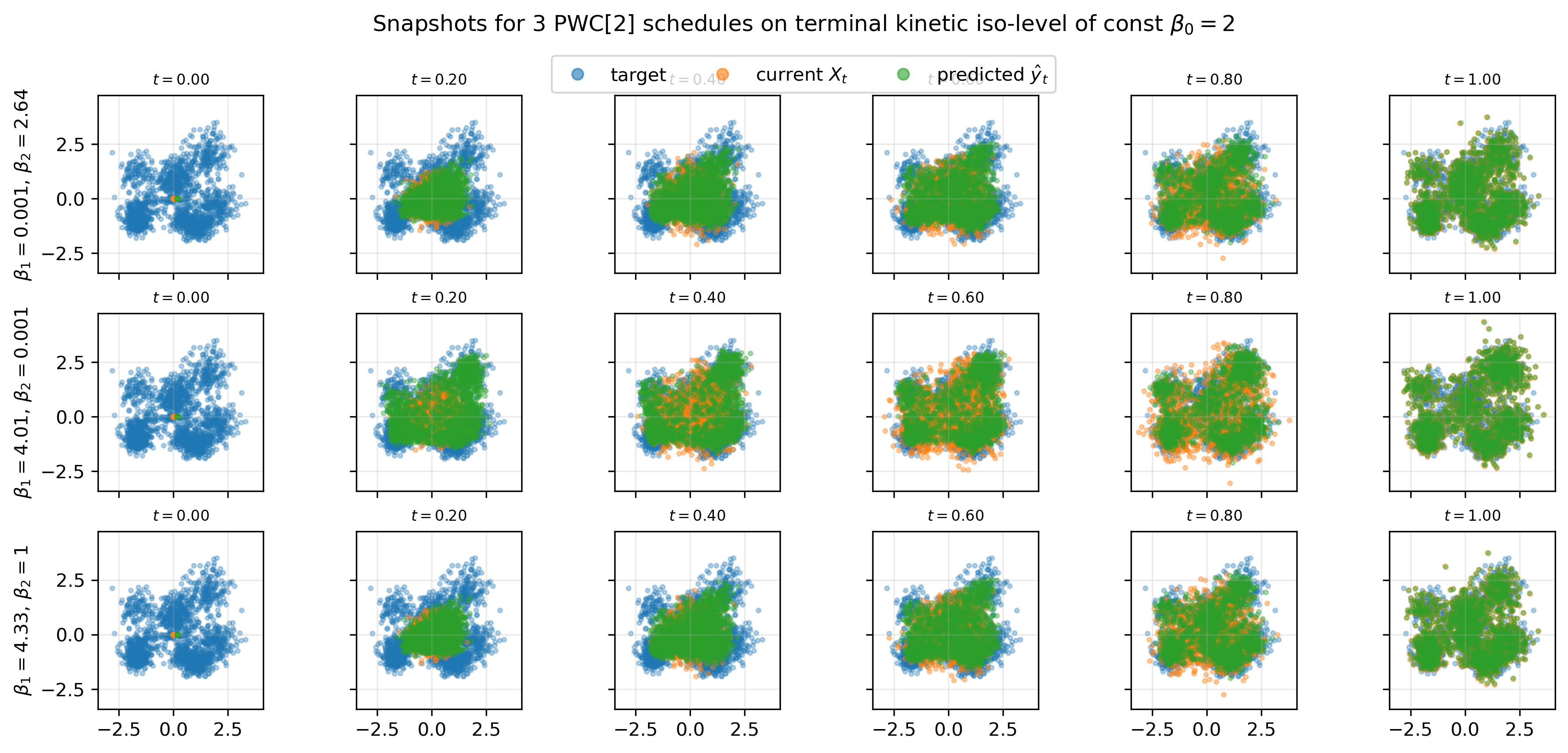}
    \label{fig:isocost-snaps}
  \end{subfigure}
  \caption{Iso--kinetic tradeoffs and trajectory–level behavior under two–piece stiffness: \emph{Left:} the kinetic iso--cost landscape $\mathcal{C}^{(\mathrm{kin})}(t=1;\beta_1,\beta_2)$ (defined in Eq.~(\ref{eq:DCG})) for two–interval PWC schedules, with the iso--level matching a constant $\beta_0=2$ highlighted. \emph{Right:} trajectory snapshots for three representative schedules on that iso--level, illustrating that markedly different temporal stiffness allocations can share the same kinetic budget while inducing distinct intermediate–time dynamics.}
  \label{fig:isocost-both}
\end{figure}

We have also conducted experiments focused on analysis of different two-interval PWC schedules corresponding to the same value of the kinetic part of the cost -- that is "cost of control". Results of the experiments are shown in Fig.~(\ref{fig:isocost-both}. Here we set target density to the ``Perturbed~A'' $3\times3$ GMM.  For each two–piece schedule with edges $\{0,\tfrac12,1\}$ and values $(\beta_1,\beta_2)$ we simulate $M$ particles for $T$ midpoint–EM steps with $X(0)=0$ and evaluate the kinetic objective $\mathcal{C}^{(\mathrm{kin})}(t=1;\beta_1,\beta_2)$ defined in Eq.~(\ref{eq:DCG}).
We then perform \emph{adaptive refinement} over $(\beta_1,\beta_2)\in[B_{\min},B_{\max}]^2$, sampling densely where the map varies rapidly, and render a colormap together with contour lines. 
The thick red level in the left panel is the set of $(\beta_1,\beta_2)$ whose $\mathcal{C}^{(\mathrm{kin})}(t=1;\beta_1,\beta_2)$ equals that of the constant schedule $\beta(t)\equiv \beta_0=2$. We pick three representative points on this iso–level (minimal $\beta_1$, minimal $\beta_2$, and a middle point along the arc) and visualize their dynamics on the right: each row shows six equally–spaced times with the target cloud (blue/gray), current samples $X_t$ (orange), and predicted means $\hat y(t;X_t)$ (green). Overall, the figure reveals that even under an \emph{equal kinetic budget} (same $\mathcal{C}^{(\mathrm{kin})}(t=1)$), redistributing stiffness across the two halves can substantially alter intermediate–time behavior.

\subsubsection{Exemplary Negative \texorpdfstring{$\beta_t$}{beta}} \label{sec:negative-beta-exp}

\begin{figure}[t]
  \centering
    \includegraphics[width=0.48\linewidth]{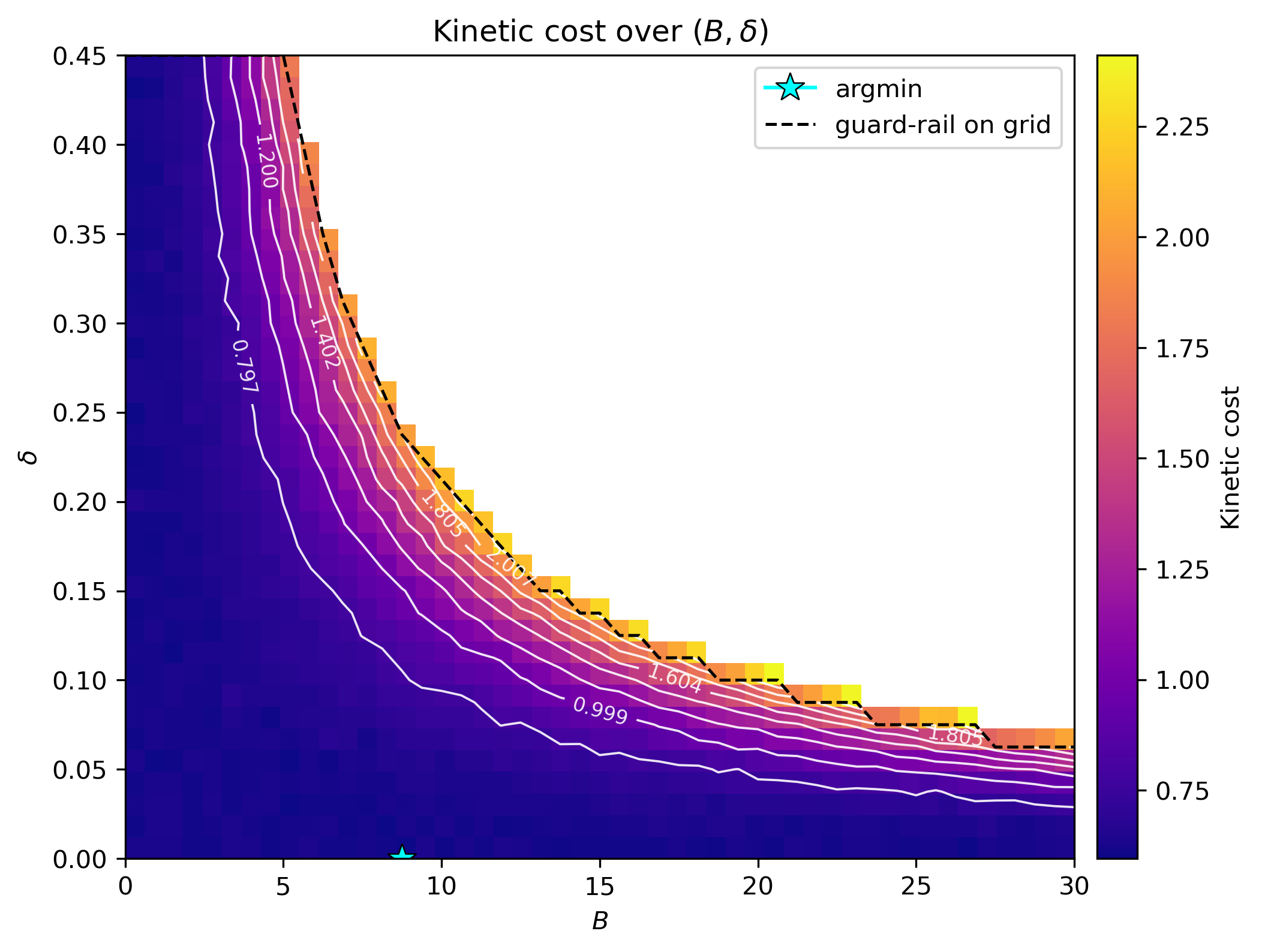}
    \includegraphics[width=0.48\linewidth]{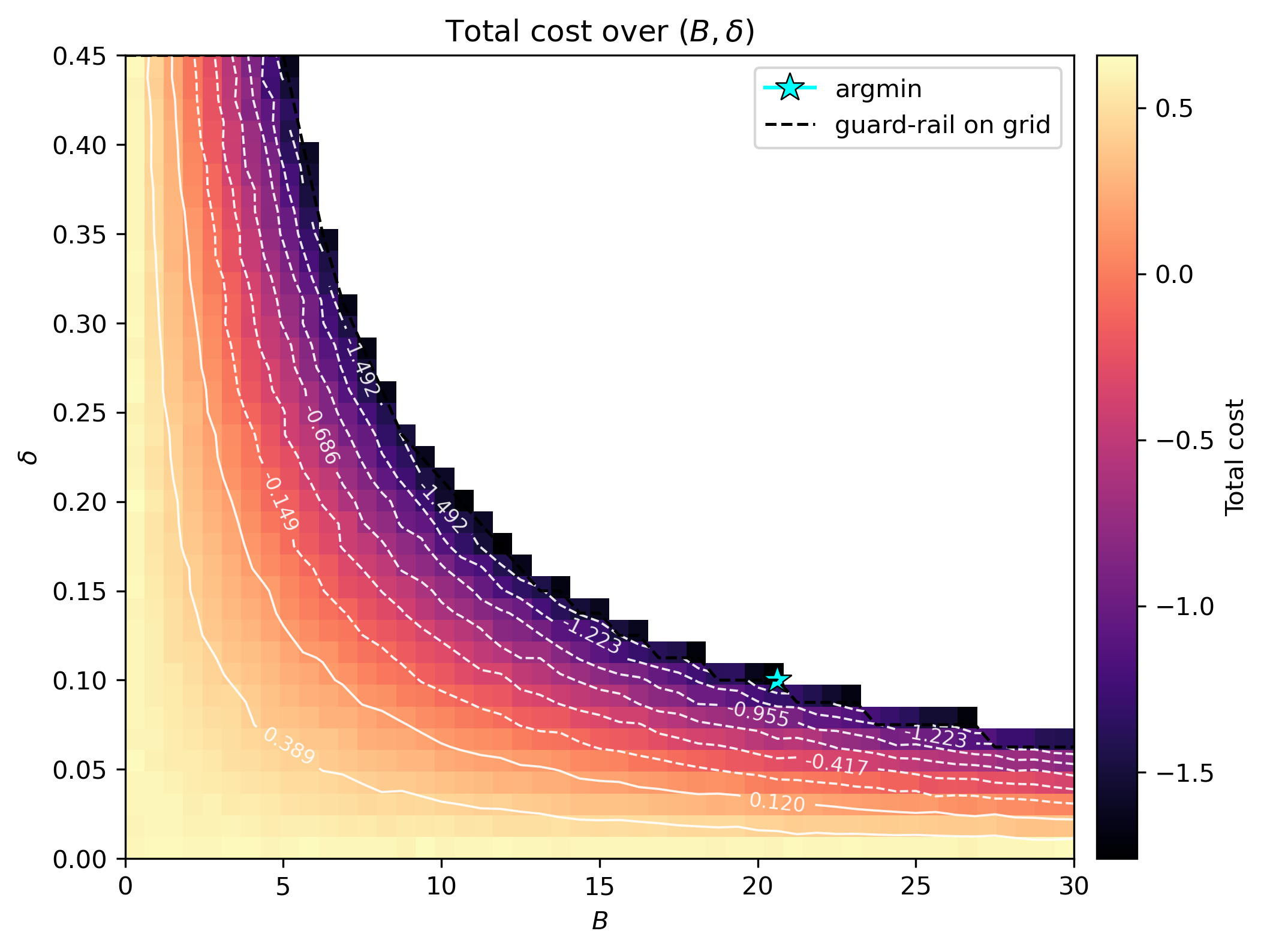}
    \includegraphics[width=\linewidth]{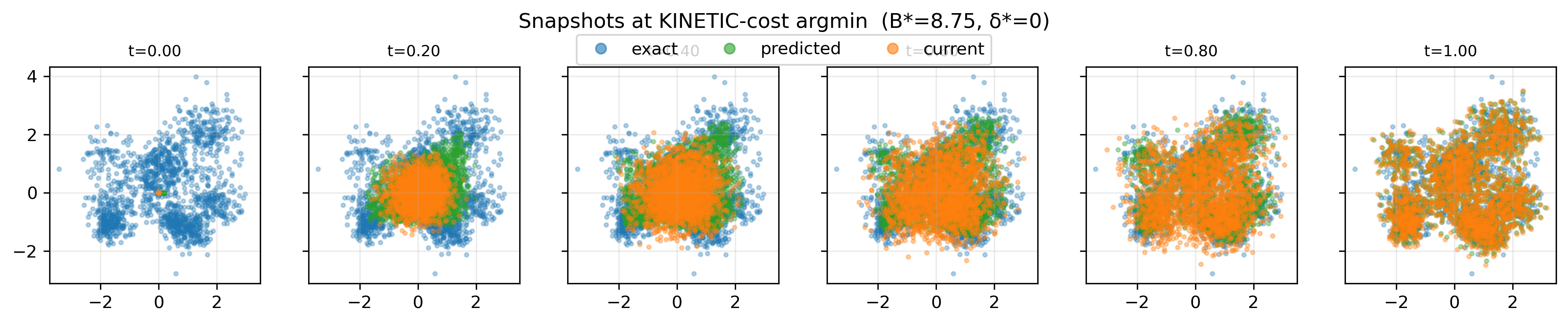}
    \includegraphics[width=\linewidth]{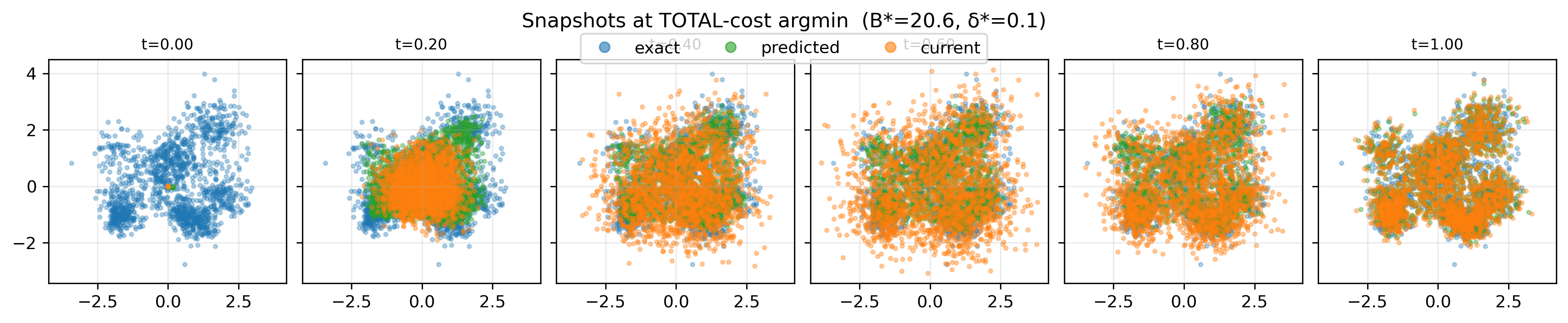}
  \caption{{\bf Top Row:} Cost landscapes for the “negative window” schedule (Appendix~\ref{sec:negative-beta}). We set $\beta_t(\delta;B)=-B$ on the window $|t-\tfrac12|<\delta$ and $\beta_t=0$ outside (cf.\ \eqref{eq:beta-negative-window}), and evaluate costs. {\bf Left Panel} shows the time\;/path averaged \emph{kinetic} budget $\int_0^1 \tfrac12\,\mathbb{E}\,\|u_t^*\|^2\,dt$ as a colormap in $(B,\delta)$, with iso-contours overlaid. {\bf Right Panel} shows the corresponding \emph{total} budget $\int_0^1 \Big(\tfrac12\,\mathbb{E}\,\|u_t^*\|^2 + \tfrac12\,\beta_t\,\mathbb{E}\,\|x_t\|^2\Big)\,dt$. In both panels, only parameters within Green-function well-posedness guard-rails  (cf.\ Eq.~(\ref{eq:nosing-plus})) are shown. {\bf Bottom Rows} show screenshots corresponding to $(B,\delta)$ protocol minimizing the kinetic cost (middle) and total cost (bottom) -- marked as stars in the top row heat-maps.}
  \label{fig:negwin-heatmaps}
\end{figure}

Fig.~\ref{fig:negwin-heatmaps} explores the kinetic and total costs for the two-parameter negative-window profile \(\beta(t)\) described in Appendix~\ref{sec:negative-beta}. The total cost (top right) is minimized by the \emph{most negative} admissible schedules, i.e., near the boundary of the Green–function well-posedness domain. In contrast, the kinetic cost (top left) is minimized by the \emph{least negative} schedules and then grows monotonically as either \(B\) (magnitude) or \(\delta\) (duration) of the negative window increases. The associated dynamics (bottom panels) differ markedly: at the total-cost optimum (which coincides with a large kinetic cost), trajectories exhibit a pronounced overshoot before collapsing to the target; at the kinetic-cost optimum the evolution is more gradual and avoids such overshoot.

\subsection{Sensitivity/Gradient Statistics}
\label{sec:grad-exp}

\subsubsection{Const-$\beta$ analysis of the velocity (optimal control) gradient statistics} 
\label{sec:VG-const}

\begin{figure}[h!]
    \centering
    \includegraphics[width=0.32\textwidth]{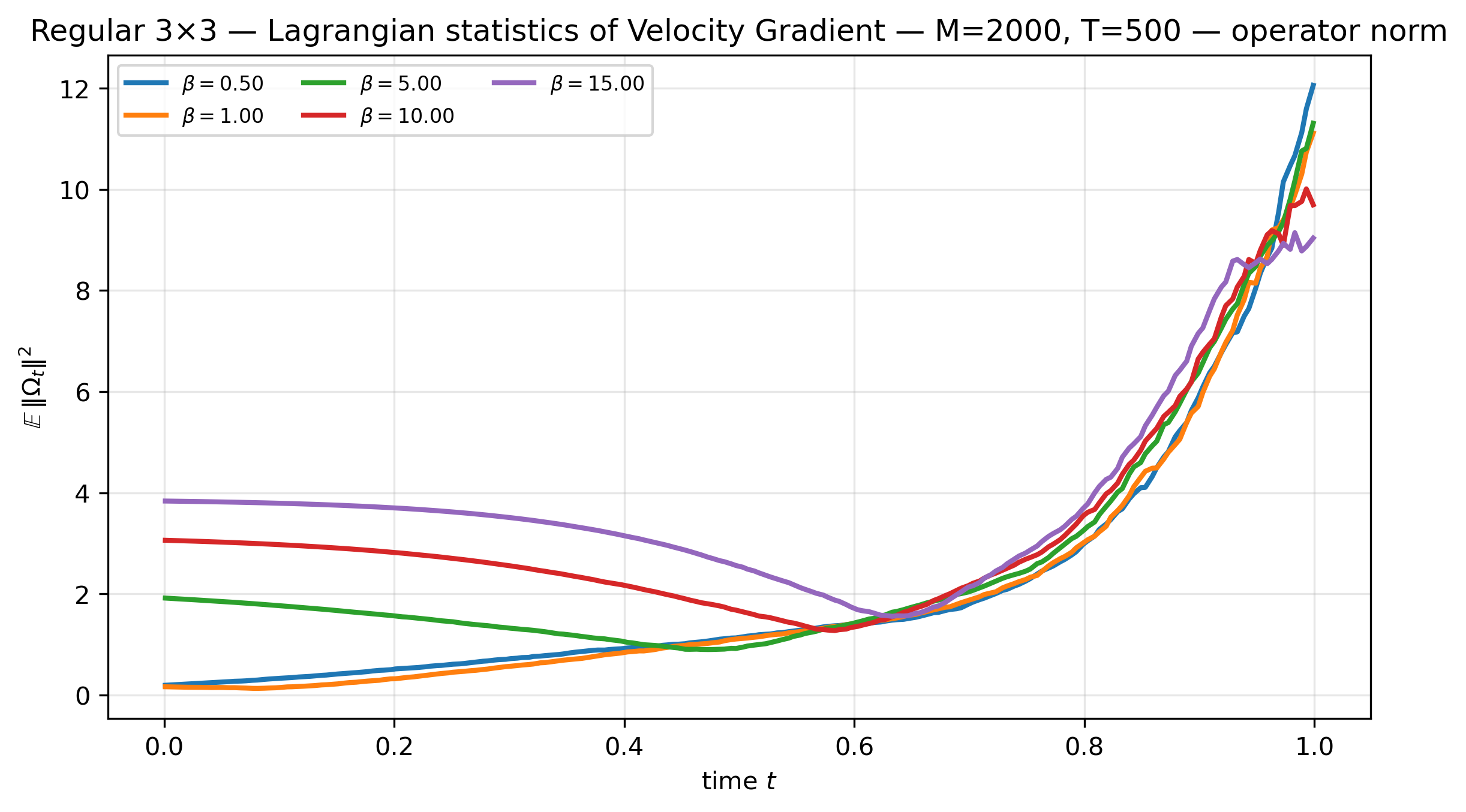}
    \includegraphics[width=0.32\textwidth]{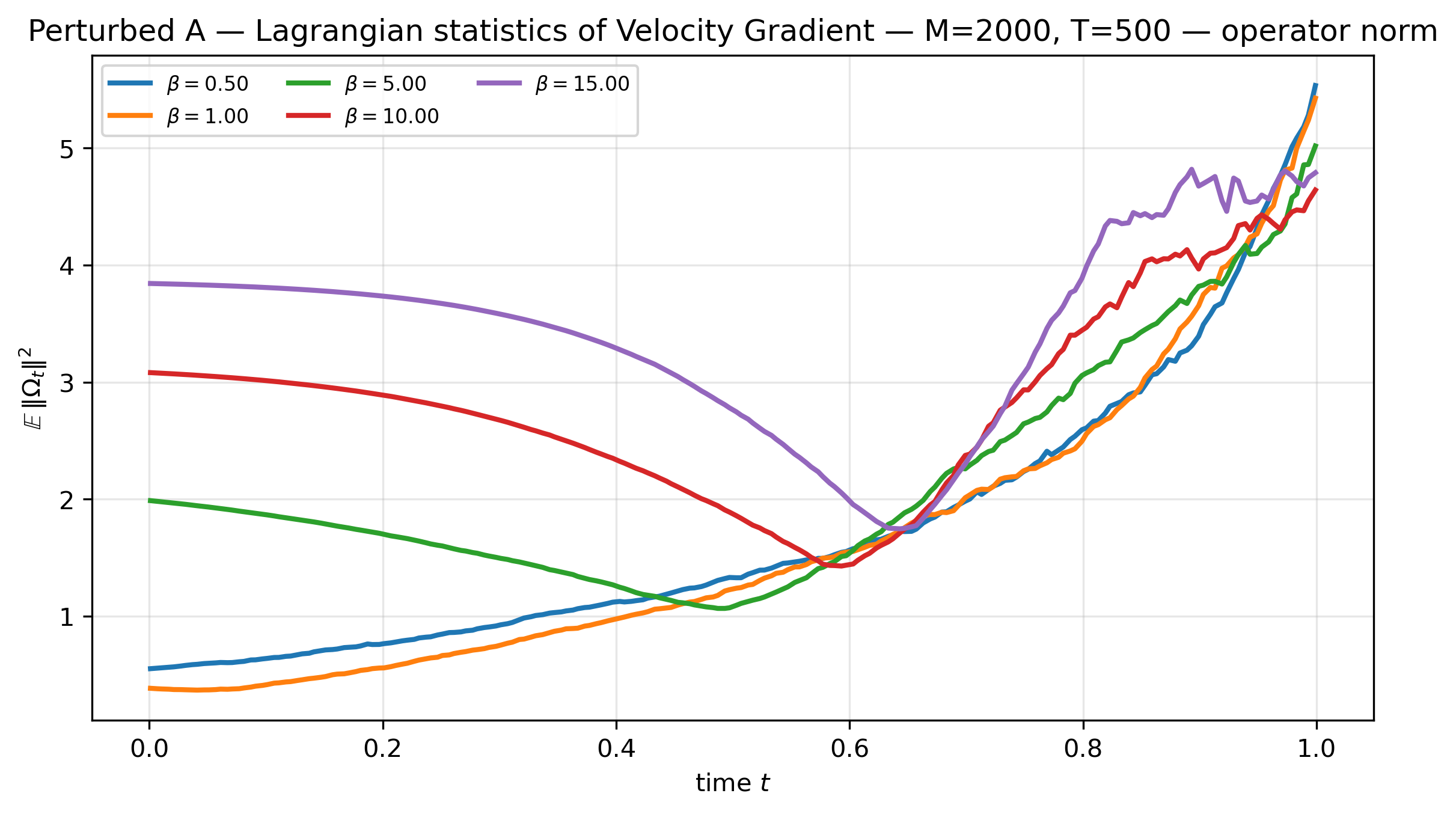}
    \includegraphics[width=0.32\textwidth]{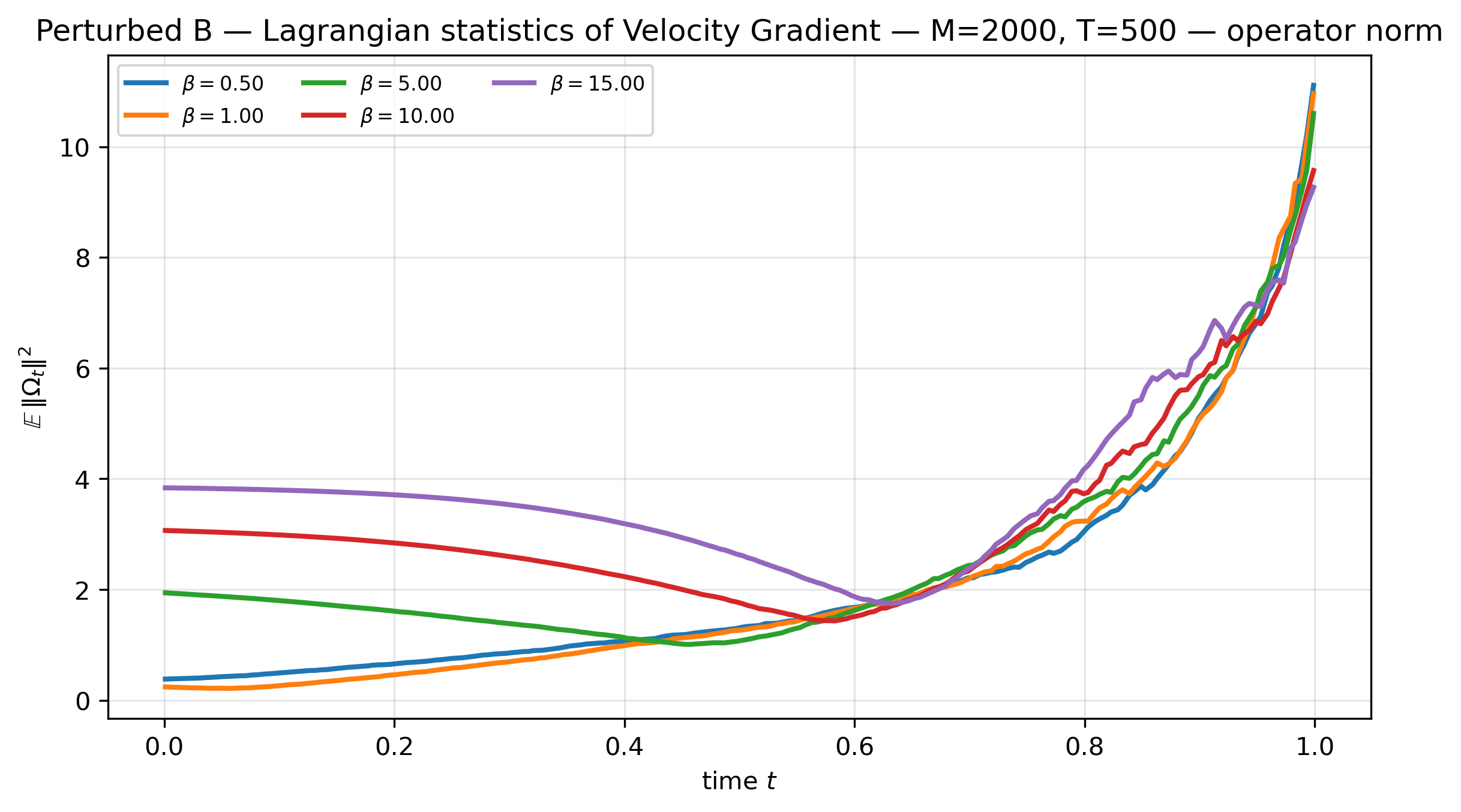}
    \includegraphics[width=0.32\textwidth]{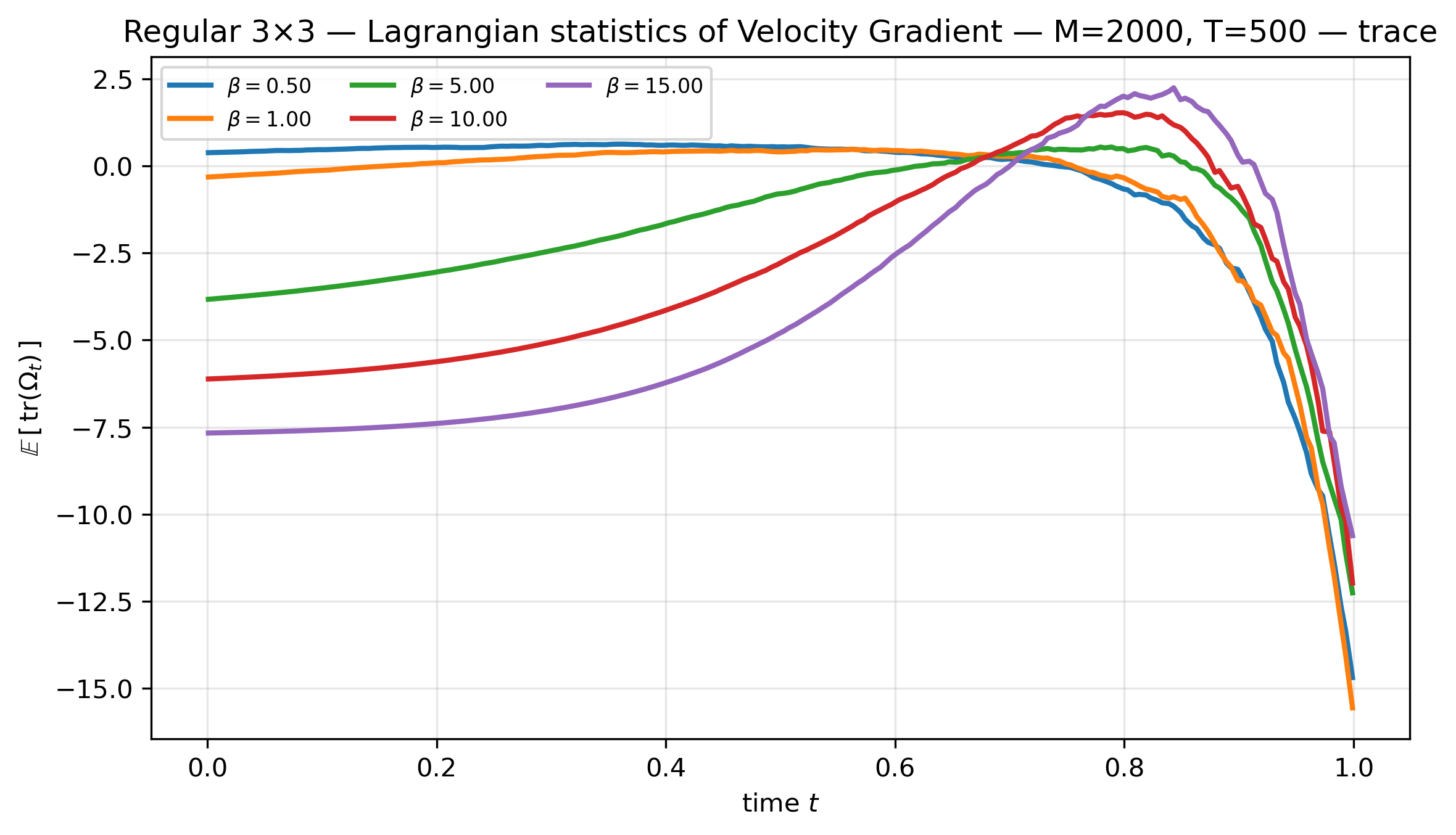}
    \includegraphics[width=0.32\textwidth]{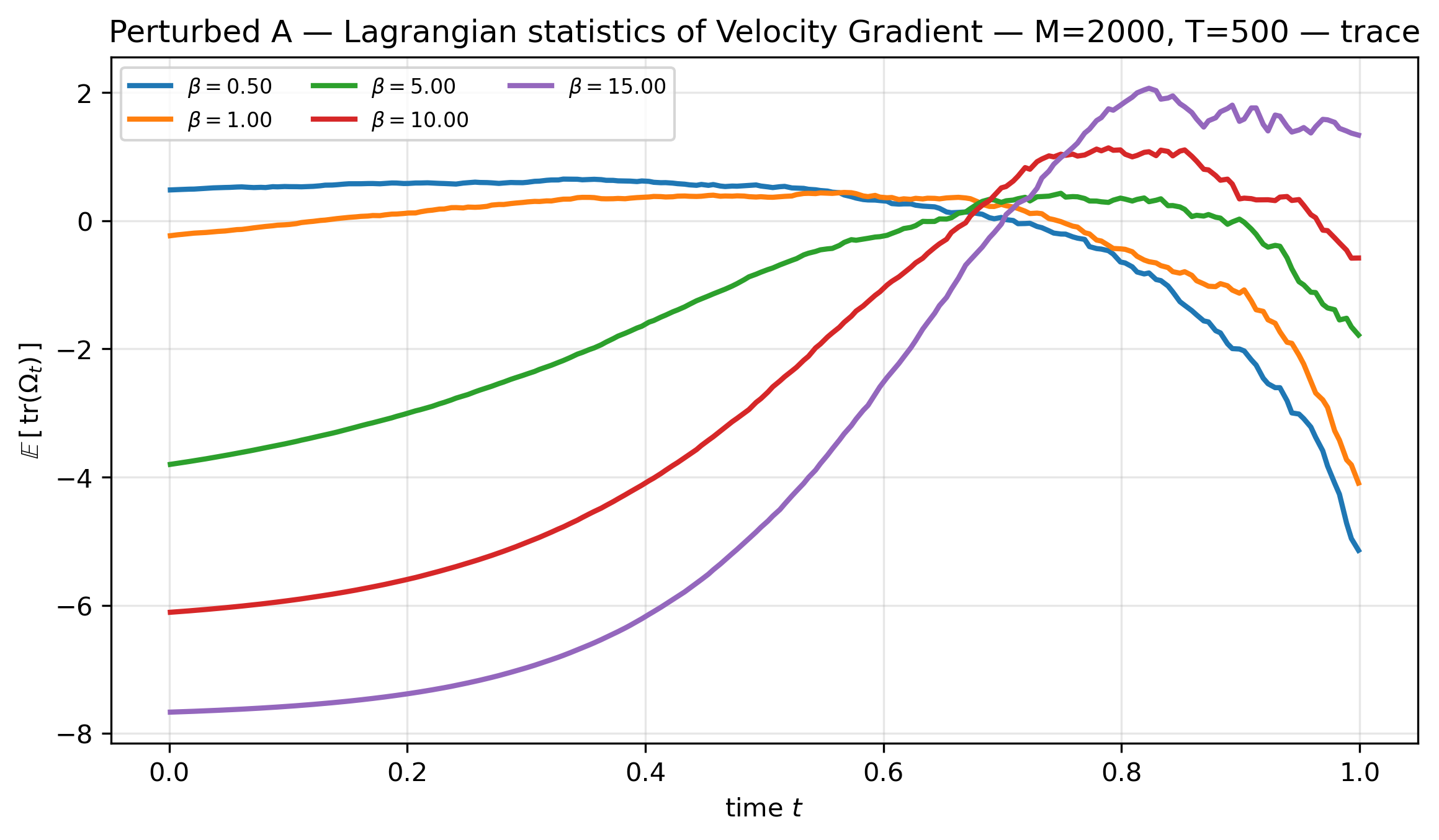}
    \includegraphics[width=0.32\textwidth]{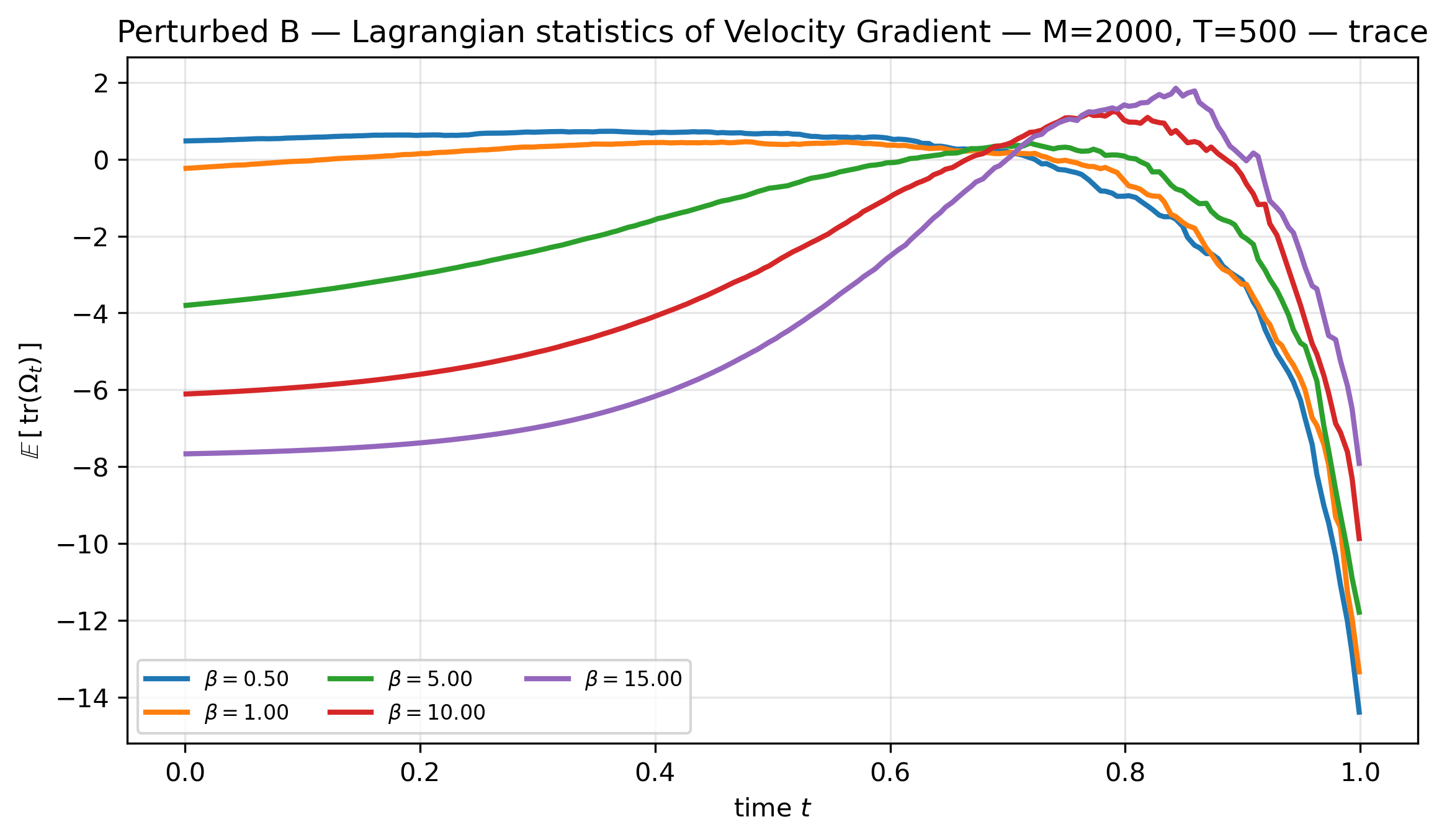}
    \includegraphics[width=0.32\textwidth]{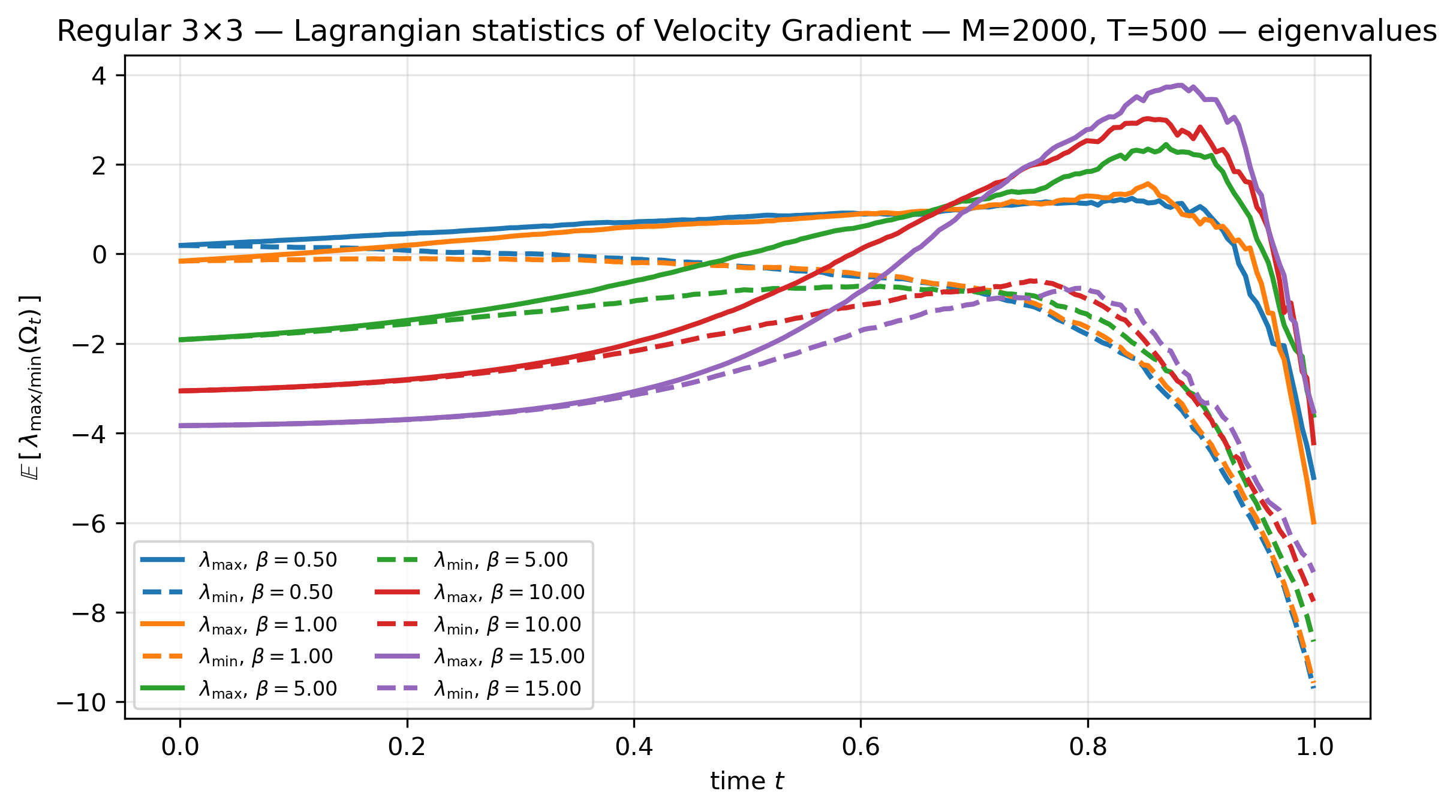}
    \includegraphics[width=0.32\textwidth]{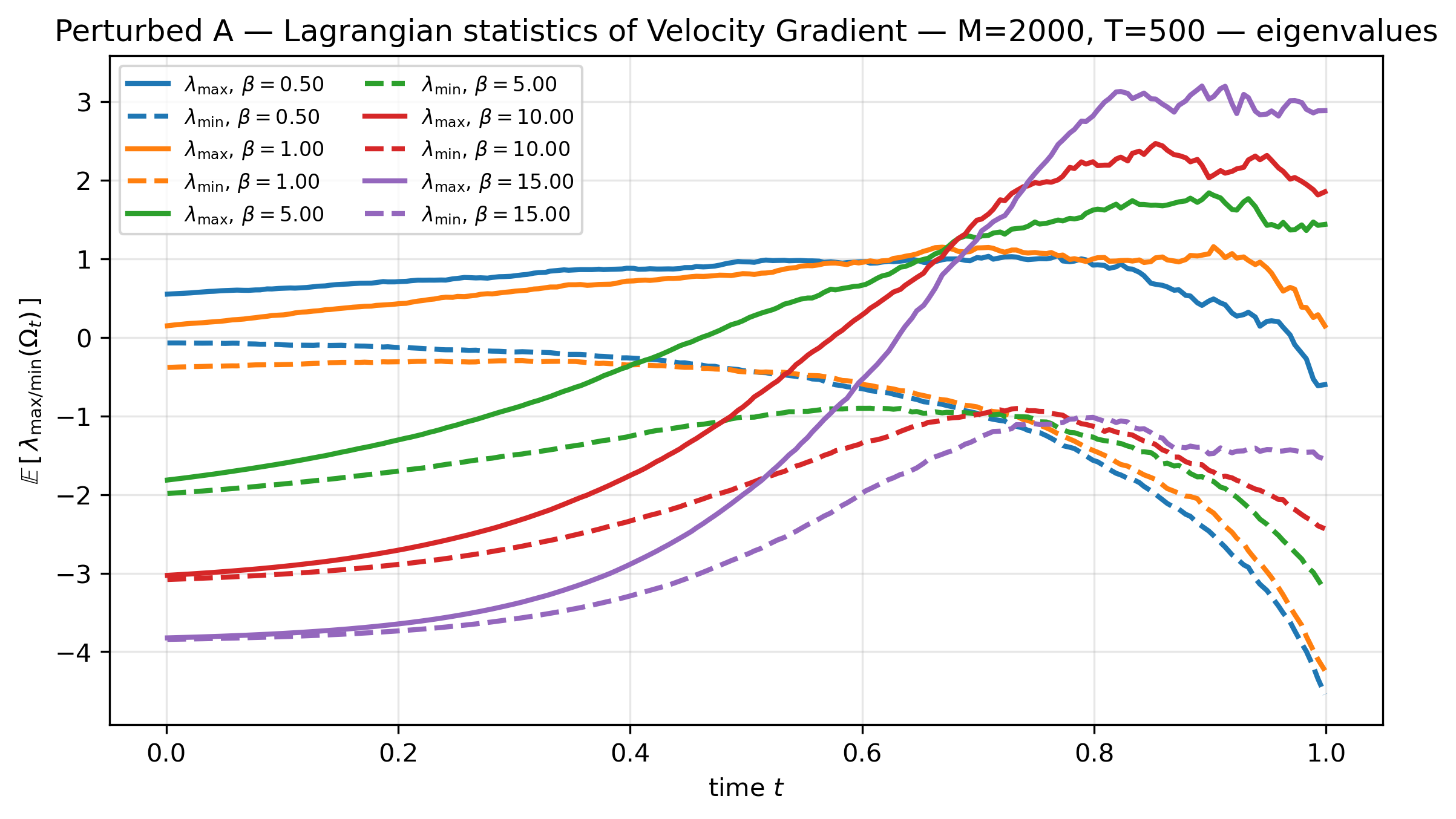}
    \includegraphics[width=0.32\textwidth]{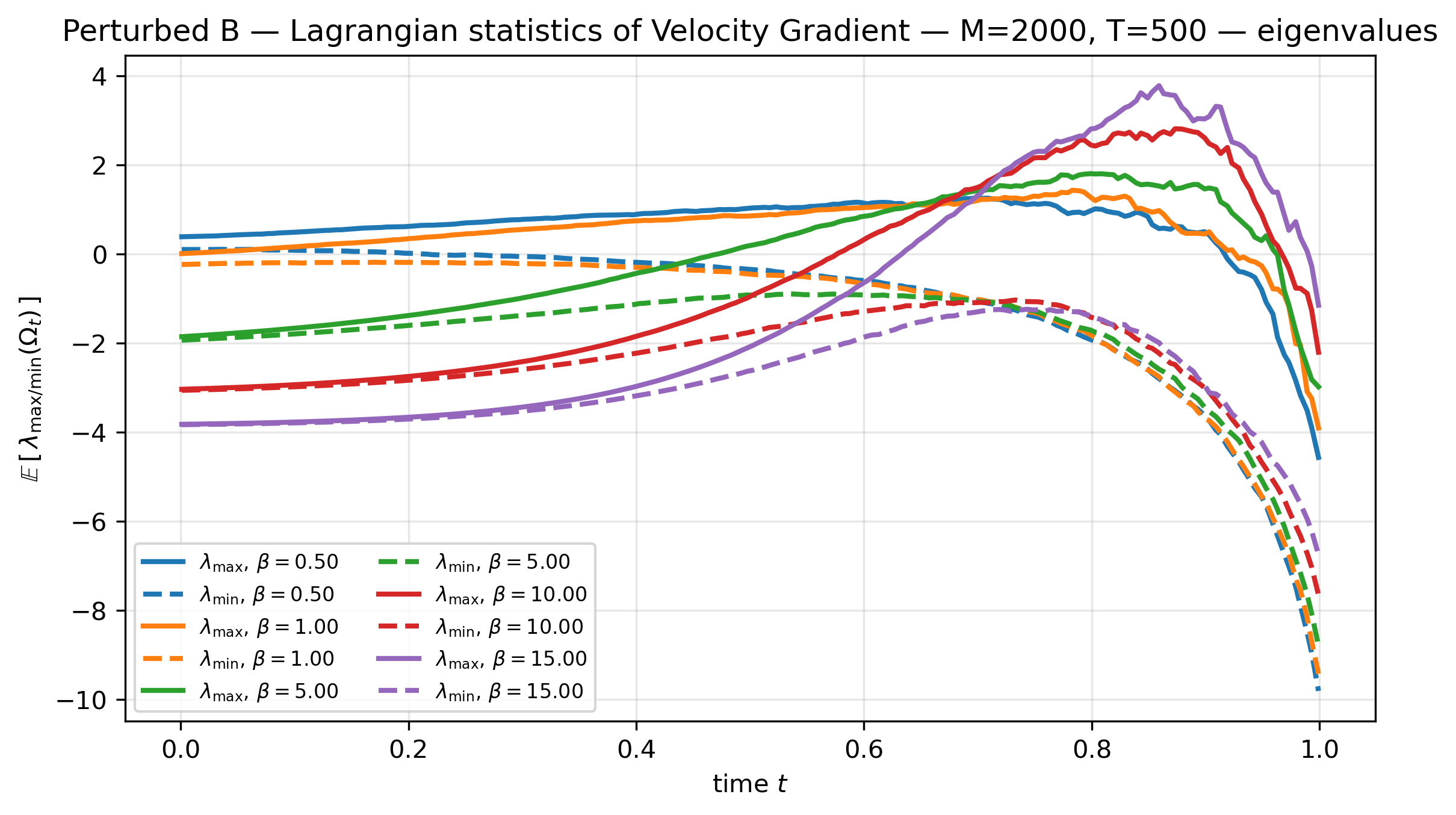}
    \caption{The figure reports dependence of the sensitivity matrix $\Omega_t$ (of velocity/optimal control gradient) defined in Eq.~(\ref{eq:Omega}) statistics on the GMM model and on const–$\beta$ schedules. Top row: Operator norm $\mathbb{E}\,\|\Omega_t\|^2$; Middle row: Trace $\mathbb{E}\,[\mathrm{tr}(\Omega_t)]$; Bottom row: Mean extremal eigenvalues: $\mathbb{E}\,[\lambda_{\max}(\Omega_t)]$ (solid) and $\mathbb{E}\,[\lambda_{\min}(\Omega_t)]$ (dashed).}
    \label{fig:omega-stats}
\end{figure}

Fig.~(\ref{fig:omega-stats}) reports analysis of the time-dependent statistics  of the sensitivity operator $\Omega_t$ defined in Eq.~(\ref{eq:Omega}), specifically mean 2-norm, mean trace and mean max/min eigenvalues.

\begin{figure}[th!]
  \begin{minipage}[t]{0.48\textwidth}
    \centering
    \subcaptionbox{Time–averaged mean operator norm vs.\ constant $\beta$.\label{fig:anytime-avgop}}{%
      \includegraphics[width=\linewidth]{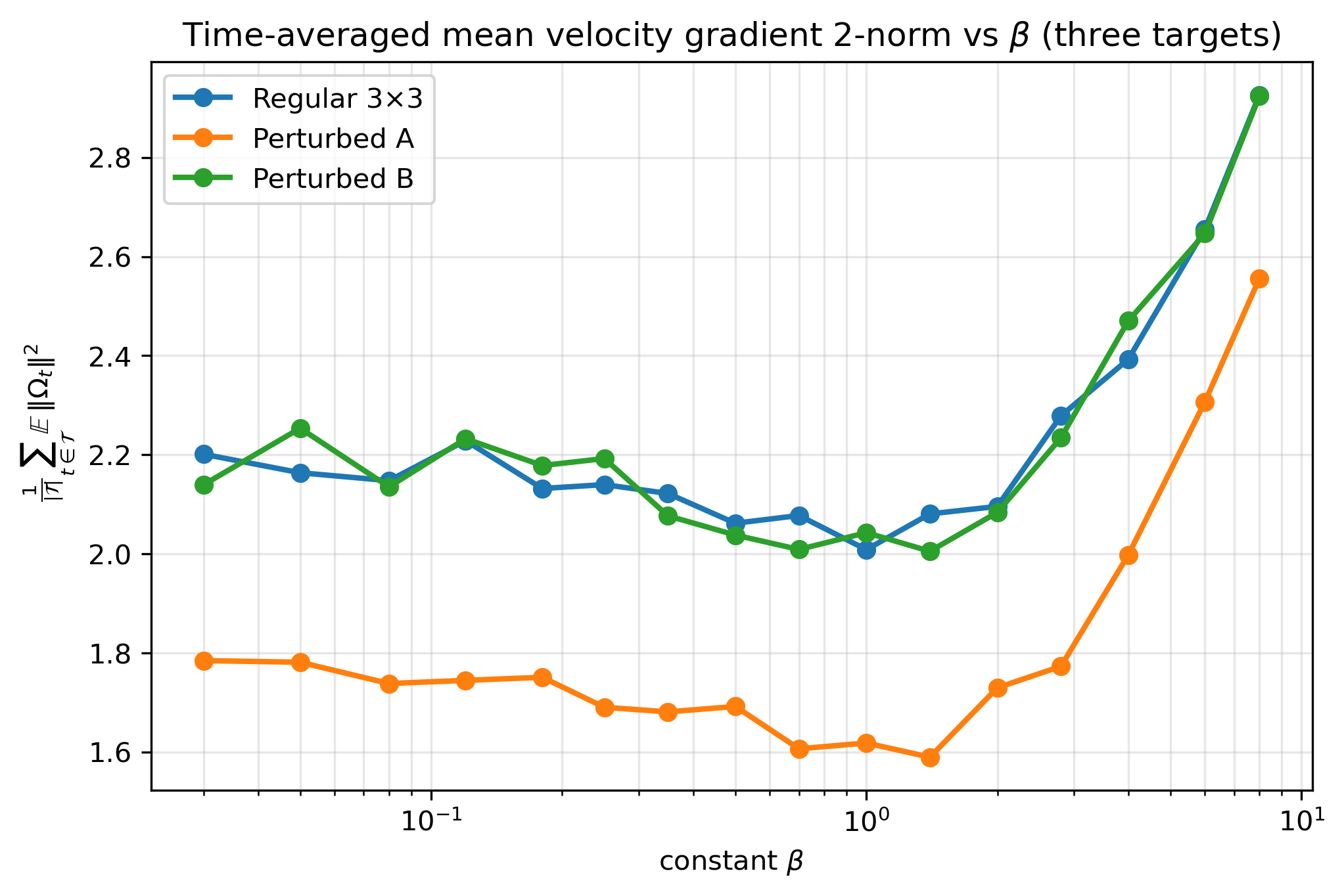}}
    \vspace{1.2ex}
    
    \subcaptionbox{Snapshot grids at $\beta^\star$ (rows: targets; columns: time).\label{fig:anytime-snaps}}{%
      \includegraphics[width=\linewidth]{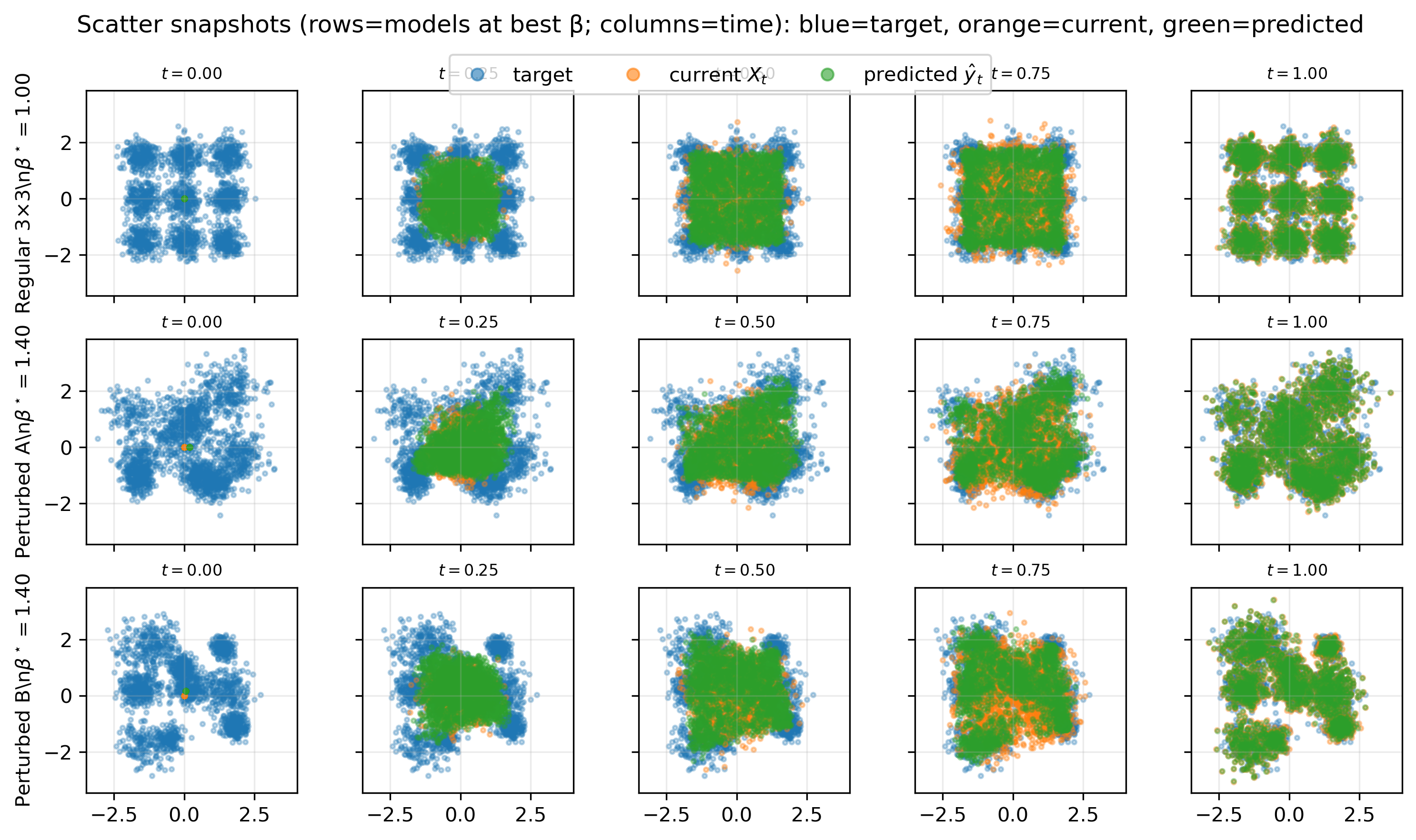}}
  \end{minipage}
  \hfill
  \begin{minipage}[t]{0.48\textwidth}
    \centering \vspace{-5.2cm}
    \subcaptionbox{Phase portraits at $\beta^\star$ (left: $x_t$, right: $\hat{x}_t$) for each target.\label{fig:anytime-phase}}{%
      \includegraphics[width=\linewidth]{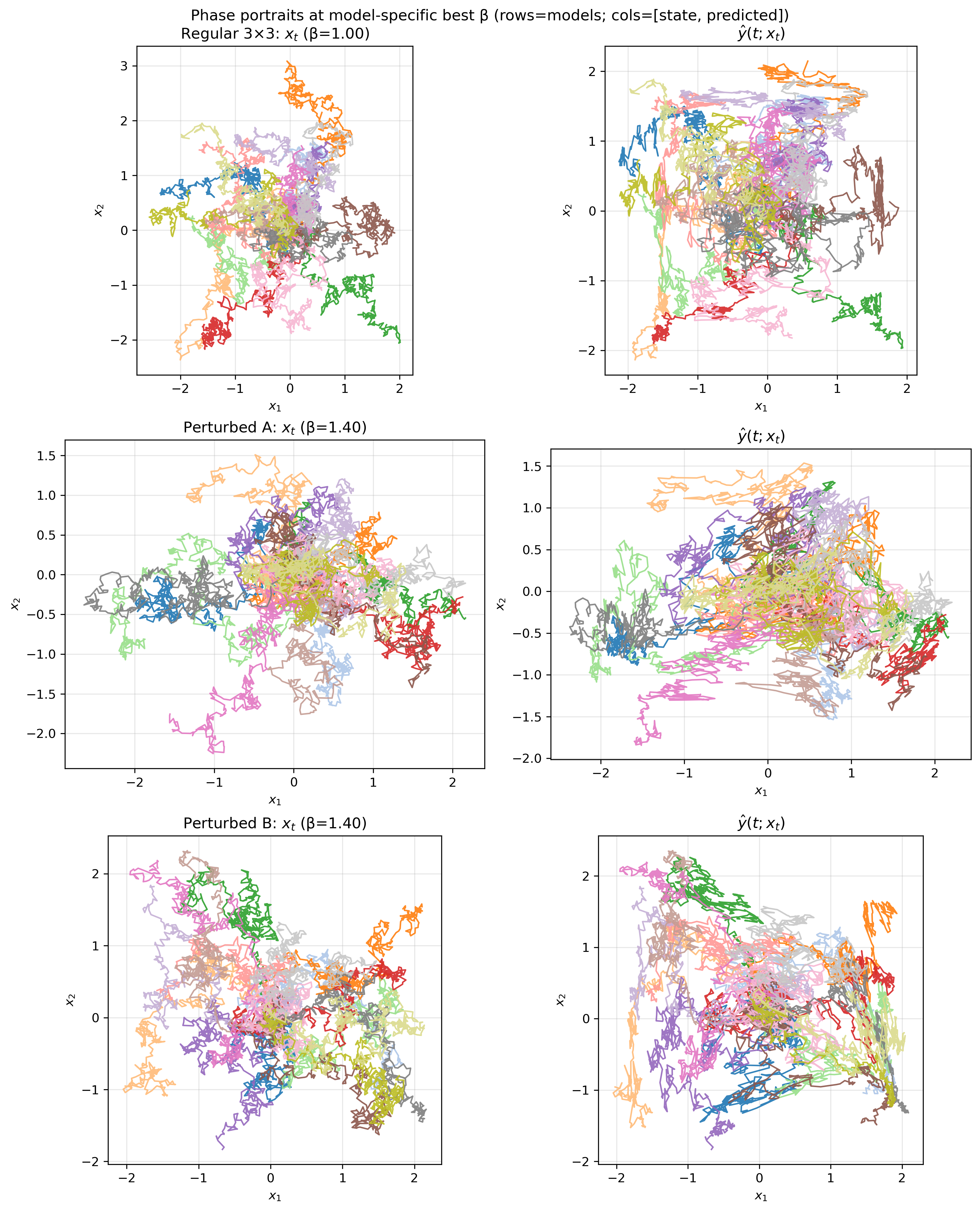}}
  \end{minipage}
  \caption{Time-integrated sensitivity across constant–$\beta$ schedules on three targets. (a) Time–averaged mean of the 2-norm velocity gradient versus constant $\beta$, for the regular 3×3 grid and two perturbed GMMs. The minimizer on each curve defines $\beta^\star$ used below. (b) Phase portraits at $\beta^\star$ (per target): state paths $x_t$ and predicted paths $\hat{y}(t;x_t)$. (c) Snapshot grids at $\beta^\star$ (rows: targets; five evenly spaced midpoints in time). Blue: exact target; orange: current $x_t$; green: predicted $\hat{y}_t$.}
  \label{fig:sensitivity-optimal-const-beta}
\end{figure}

Fig.~(\ref{fig:sensitivity-optimal-const-beta}) shows (a) dependence of the time-averaged mean 2-norm
of the velocity  gradient defined in Eq.(\ref{eq:Omega}) on const-$\beta$; (b)   respective phase-portraits; and (c) dynamic scatter plots -- all for our three exemplary GMMs. We observe that  the time-averaged mean 2-norm of the velocity  gradient has a well-defined minimum at $O(1)$ const-$\beta$. 

Notably -- of all the experiments so far (but under exception of the "negative"-$\beta$ case discussed in Section \ref{sec:negative-beta-exp}) this is the first objective which resulted in the optimality achieved not at the extremes (const-$\beta\to 0$ or $\beta\to \infty$) but at a $O(1)$ values.

Furthermore, even thought we observe variations from one GMM to another, the optimal values of const-$\beta$ are all within numerical error from each other. This last observation motivates us to examine time-dependent (PWC-) schedule of $\beta$. 

\subsubsection{Hierarchical PWC-$\beta$ Optimization}\label{sec:PWC-opt-VG}

\begin{figure}[t]
  \centering
    \includegraphics[width=\linewidth]{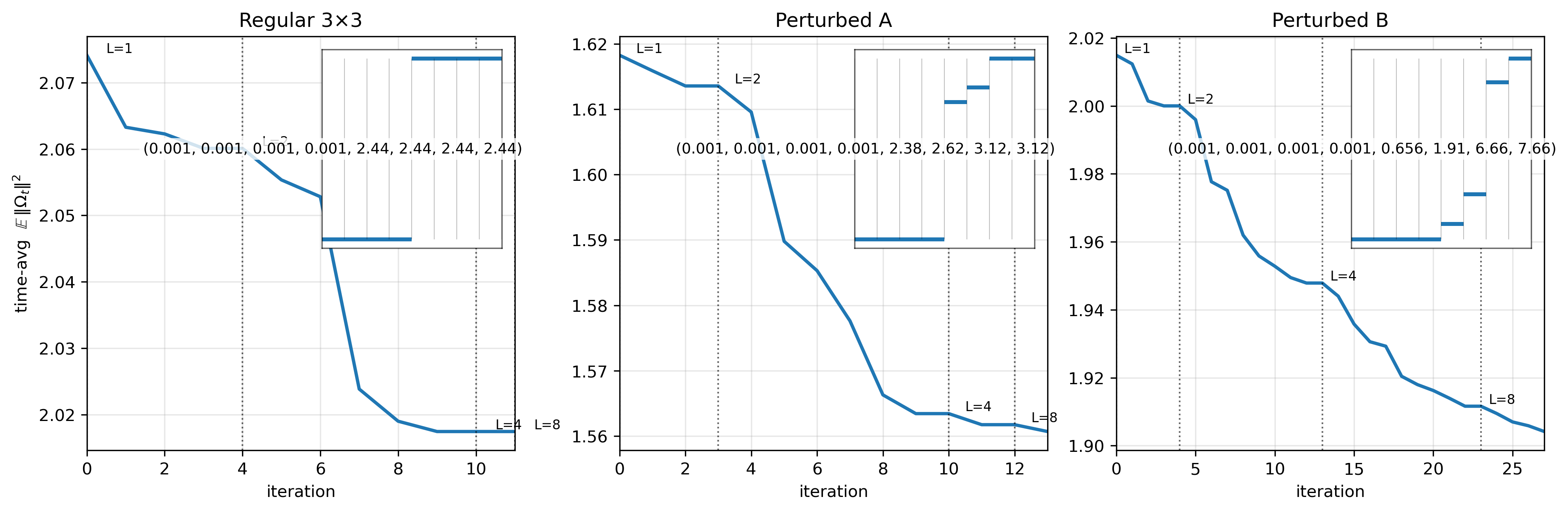}
    \includegraphics[width=\linewidth]{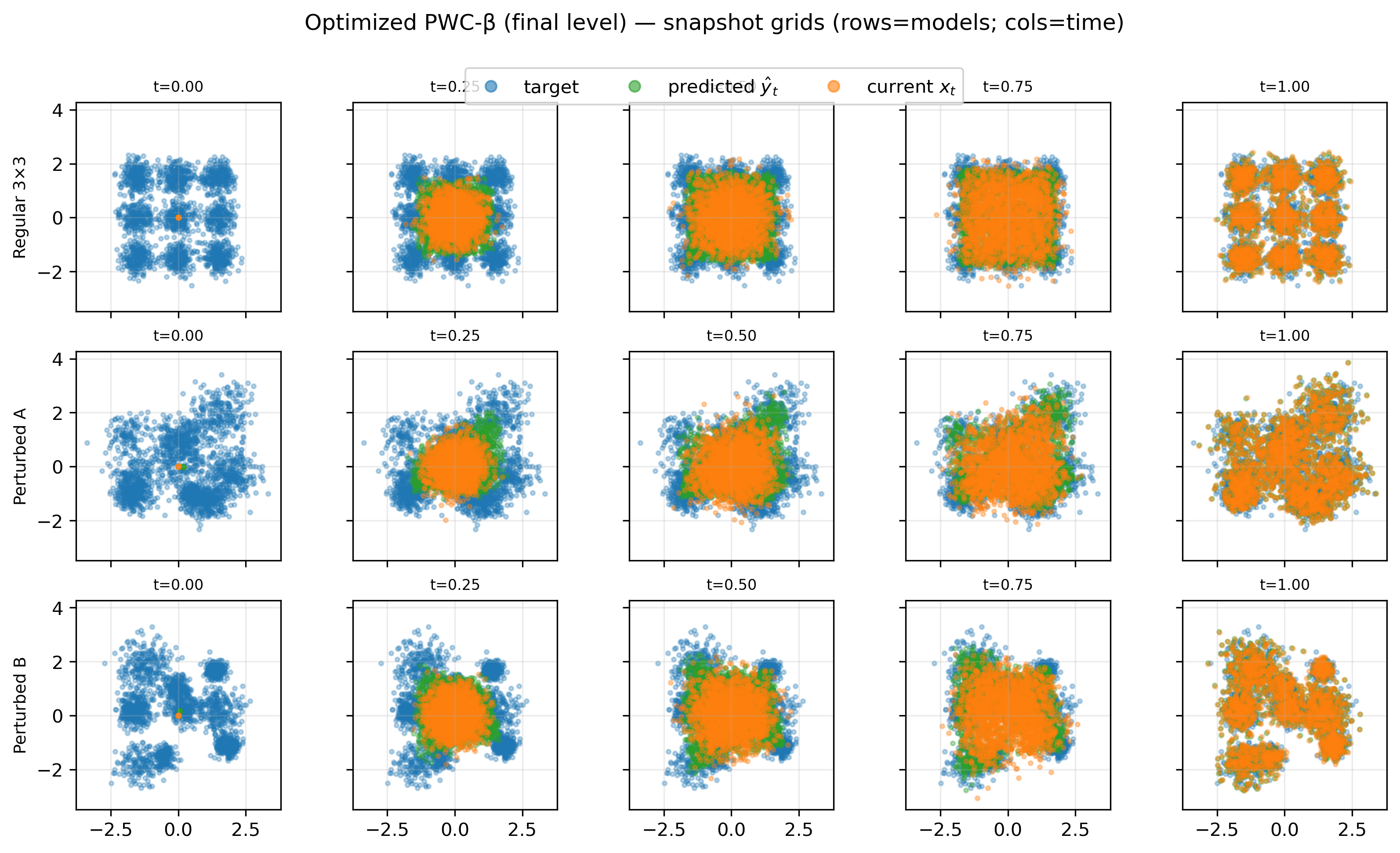}
  \caption{
  Summary of the three-model study and hierarchical PWC-$\beta$ minimization of the time-averaged mean velocity-gradient $2$-norm. Top: Convergence traces for the nested 1$\to$2$\to$4$\to$8 interval optimization, with top insets visualizing the final $\beta(t)$ steps. Bottom: Qualitative evolution (scatter snapshots) under the final schedules: for each model, we show target samples, predicted $\hat y_t$, and current states $x_t$ at five time slices.  
  }
  \label{fig:three-models-pwc-summary}
\end{figure}

Fig.~(\ref{fig:three-models-pwc-summary}) reports extension of the analysis of minimization of the time-averaged mean velocity-gradient $2$-norm -- defined in Eq.~(\ref{eq:Omega}) from the case of const-$\beta$, just discussed above in Section \ref{sec:VG-const}), to the case over a time-dependent PWC-$\beta$ (cf. the PWC-$\beta$ framework was set up in Section \ref{sec:PWC}). We optimize the PWC-$\beta(t)$ by hierarchical refinement: starting with a single constant level ($L{=}1$), we minimize the time–averaged mean velocity–gradient norm
\[
\min_{\{\beta_\ell\}_{\ell=1}^{L}} \;\; \frac{1}{|\mathcal T|}\sum_{t\in\mathcal T} \mathbb{E}\!\left[\|\Omega_t\|^2\right],
\qquad
\Omega_t \;=\; b^{-}(t)\,\partial_x\hat{y}(t;x)\;-\;a^{-}(t)\,I,
\]
using coordinate descent under the PWC parameterization $\beta(t)\equiv\beta_\ell$ for $t\in[\tau_{\ell-1},\tau_\ell)$. After convergence at level $L$, we refine $L\!\to\!2L$ by splitting each interval at its midpoint and initializing child values by their parent’s $\beta_\ell$, then re–optimize; this $1\!\to\!2\!\to\!4\!\to\!8$ cascade converges fast. The optimal PWC-$\beta$ are expressive and notably different for different GMMs (even though optimal constant-$\beta$ schedules for the same GMMs were all within a numerical accuracy from each other). In general, we observe that $\beta(t)\approx 0$ is favored at early times with eventual and monotonic in $t$ increase of $\beta(t)$ at later times. 

\subsection{Drift–Diffusion Balance}\label{sec:drift-diff-balance-exp}

\begin{figure}[h!]
    \centering
    \includegraphics[width=\textwidth]{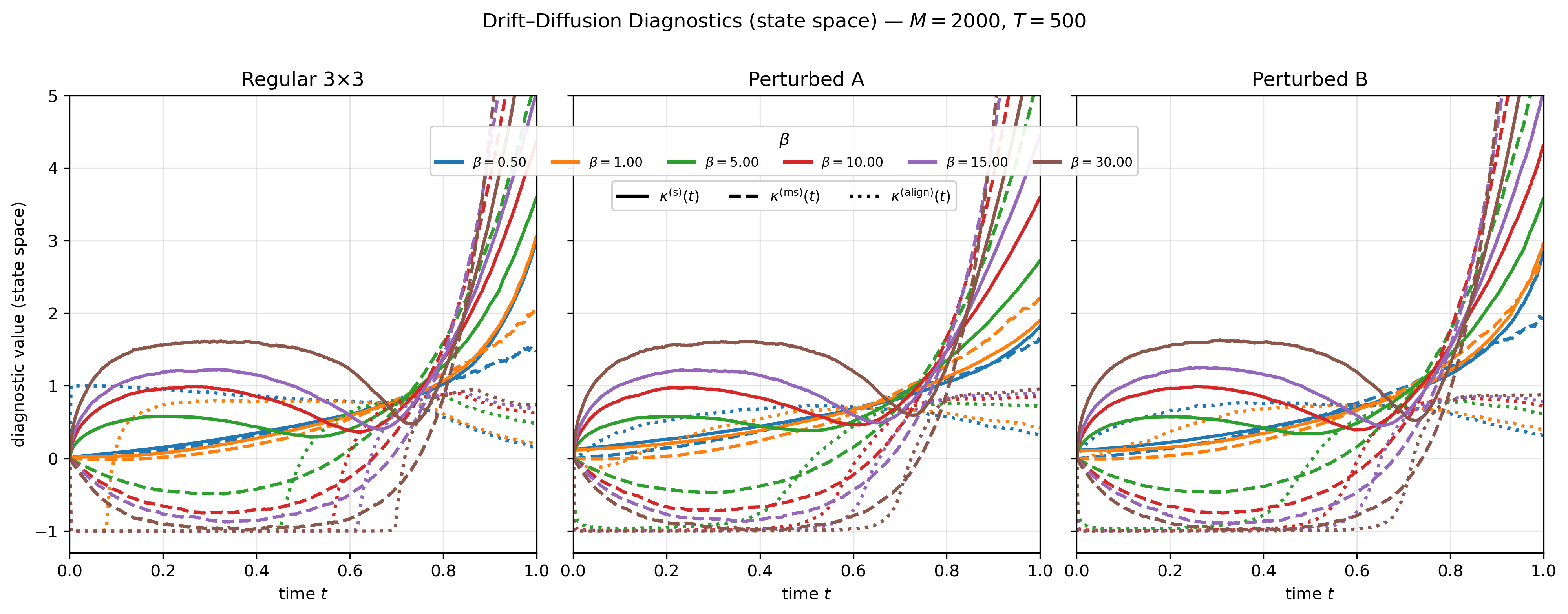}
\caption{Drift–diffusion diagnostics in state space vs.\ time for the three exemplary GMMs. Trajectories are simulated from the SDE (\ref{eq:x-SODE}) with optimal drift $u^*$. Colors encode constant–$\beta$ schedules. Solid: $\kappa^{(\mathrm{s})}(t)=\sqrt{\mathbb{E}\,\|u^*(x_t,t)\|^2/d}$; dashed: $\kappa^{(\mathrm{ms})}(t)=2\,\mathbb{E}[x_t\!\cdot u^*(x_t,t)]/d$; dotted: $\kappa^{(\mathrm{align})}(t)=\mathbb{E}\!\big[\,(x_t\!\cdot u^*(x_t,t))/(\|x_t\|\,\|u^*(x_t,t)\|)\,\big]$. (See Section \ref{sec:drift-diff-balance} for definitions.)}    
    \label{fig:drift-diff}
\end{figure}

Fig.~\ref{fig:drift-diff} reports analysis of the drift–diffusion balance discussed in Section~\ref{sec:drift-diff-balance}. The drift–to–diffusion strength $\kappa^{(\mathrm{s})}(t)$ starts at $0$ and diverges as $t\!\to\!1$ when diffusion vanishes; the way it grows, however, depends strongly on $\beta$. For small $\beta$, $\kappa^{(\mathrm{s})}(t)$ increases monotonically, and the drift is consistently aligned with the state: the alignment cosine $\kappa^{(\mathrm{align})}(t)$ begins near $1$, then decreases smoothly with $t$ while remaining positive. For large $\beta$, the early-time drift $u^*$ initially opposes $x$; consequently $\kappa^{(\mathrm{align})}(t)<0$ at small $t$, crosses zero, and turns positive later. In this regime $\kappa^{(\mathrm{s})}(t)$ is non-monotone: it exhibits a pronounced maximum followed by a dip before rising again and diverging as $t\!\to\!1$. (Consistently, the mean-square drift share $\kappa^{(\mathrm{ms})}(t)$ mirrors these patterns: near $0$ at very early times, approaching $1$ as drift dominates, and showing the same non-monotonic transient at large $\beta$.) 

Curves for different GMMs are qualitatively similar. Here we only tested const-$\beta$. However, we expect (based on analysis of Section \ref{sec:PWC-opt-VG}) that if and when any of the drift-to-diffusion indicators examined contribute the optimization objective (time-integrated, or limited to specific times) and the optimization is over a time-dependent $\beta$ -- the resulting optimal values will show a stronger dependence on the model. 

\subsection{Auto-Correlation Functions}\label{sec:AC-exp}

Here we experiment with auto-correlation functions defined in Section \ref{sec:AC-exp}.

As observed in \cite{behjoo_harmonic_2025}, $\hat{\mathcal{A}}(t;\beta_{0\to 1})$ can exhibit a sharp transition in $t$ in high dimensions (or possibly in the case of statistics with heavy tails -- which is obviously absent for the GMMs considered here). 

To enforce such transition in a scale-free manner, we map $\hat{\mathcal{A}}$ into the ``distance from the transition'' via $t\mapsto |2\,\hat{\mathcal{A}}(t)-1|\in[0,1]$, which equals $0$ at the transition level $1/2$ and approaches $1$ near $0$ or $1$.

We then define the "sharpness functional" as the time–integrated \emph{deficit} from the extremes:
\begin{equation}
\label{eq:sharpness}
\hat{S}(\beta_{0\to 1})
\;\doteq\;
\int_{0}^{1}\Bigl(1-\bigl|\,2\,\hat{\mathcal{A}}(t;\beta_{0\to 1})-1\,\bigr|\Bigr)\,dt
\;\in [0,1].
\end{equation}
Smaller $\hat{S}$ indicates sharper transition -- a perfect step transition would yield $\hat{S}\approx 0$. 

When we attempt to find optimal $\beta$ which minimizes $\hat{S}(\beta_{0\to 1})$ we discover that the largest $\beta$ -- delaying the transition as much as possible -- are favored.

This suggests introducing a regularization of $\hat{S}(\beta_{0\to 1})$:
\begin{equation}
\label{eq:sharpness-reg}
\hat{S}^{(\text{\tiny reg})}(\beta_{0\to 1})=\underbrace{\hat{S}(\beta_{0\to 1})}_{\text{encourage sharp transition}}\;+\;\underbrace{\lambda^{(\text{\tiny trans})}\,
\bigl(t^\ast(\beta_{0\to 1})-t^{(\text{\tiny trans})}\bigr)^2}_{\text{localize transition}}\,,
\end{equation}
where $t^\ast(\beta_{0\to 1})$ is defined as the time of $\hat{\mathcal{A}}(t;\beta_{0\to 1})$ first crossing of $1/2$ (since $\hat{\mathcal{A}}(t;\beta_{0\to 1})$ is normally monotonic in $t$ it will likely be the only crossing) and $\lambda^{(\text{\tiny trans})}$ is the respective regularization coefficient emphasizing tradeoff between sharpness and localization (in time) of the transition. 

\begin{figure}[t]
  \centering
  \begin{subfigure}[t]{0.49\linewidth}
    \centering
    \includegraphics[width=\linewidth]{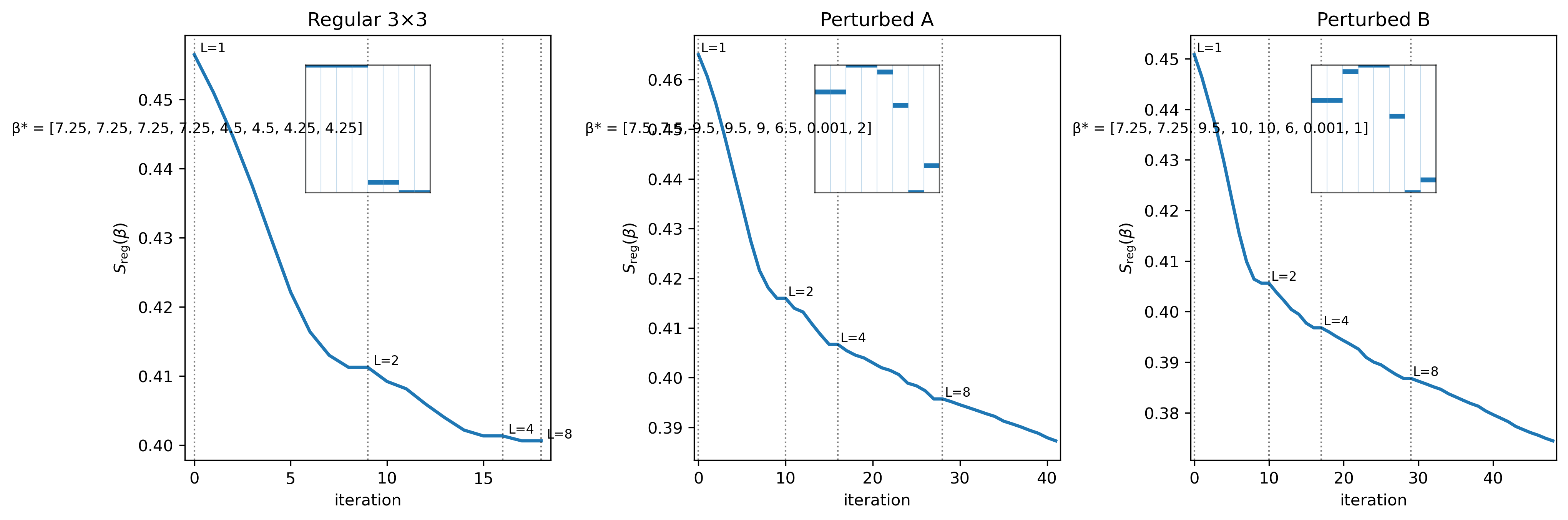}
    \caption{Convergence of the regularized sharpness objective
      \( \widehat S(\beta) + \lambda_{\mathrm{time}}\!\left(t^\ast - t^{(\mathrm{trans})}\right)^2 \)
      for the three targets (Regular 3$\times$3, Perturbed~A, Perturbed~B). Vertical dotted lines mark level changes
      (1\(\to\)2\(\to\)4\(\to\)8 intervals). Insets show the final PWC-8 \(\beta(t)\) profiles and the corresponding
      optimized \(\beta\) values.}
    \label{fig:sharp-conv}
  \end{subfigure}
  \hfill
  \begin{subfigure}[t]{0.49\linewidth}
    \centering
    \includegraphics[width=\linewidth]{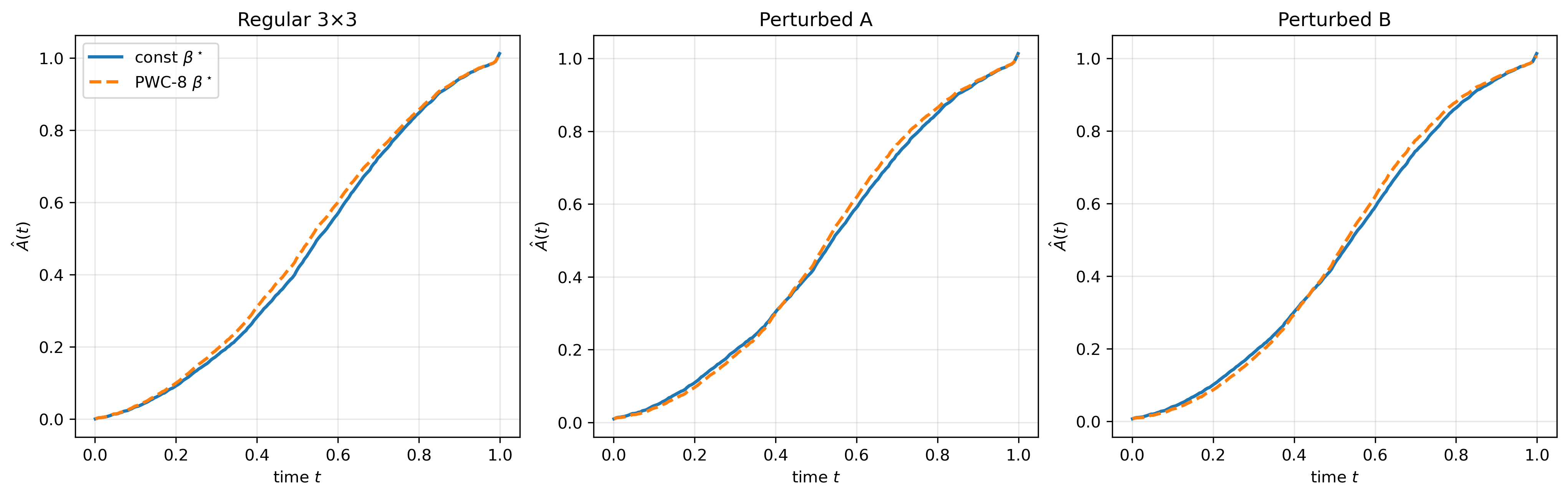}
    \caption{Predicted auto-correlation \(\hat A(t)\) for each target: constant-\(\beta^\star\) (solid) versus PWC-8 optimum (dashed).}
    \label{fig:sharp-ahat}
  \end{subfigure}

  \vspace{0.6em}

  \begin{subfigure}[t]{0.99\linewidth}
    \centering
    \includegraphics[width=\linewidth]{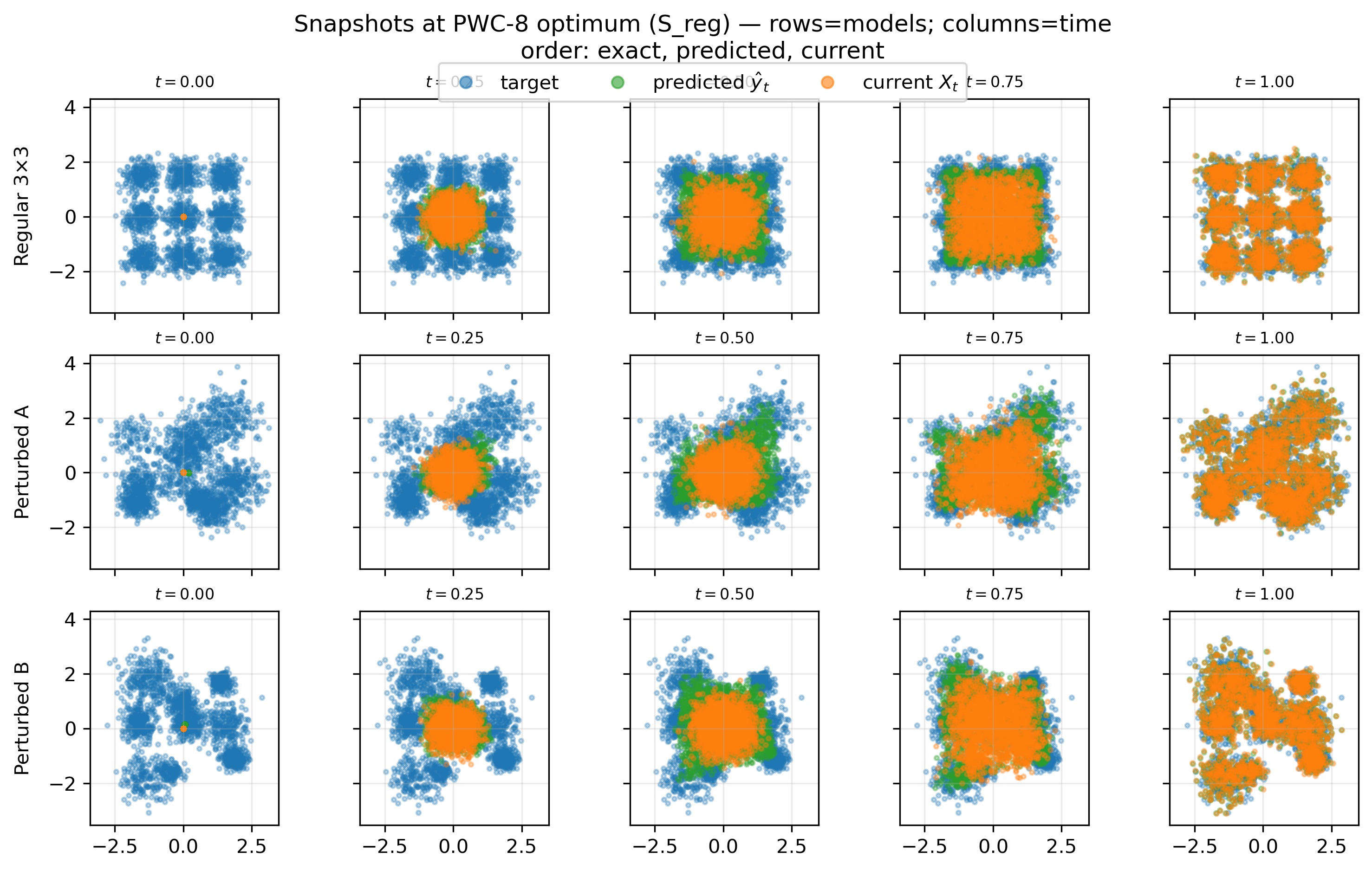}
    \caption{Snapshot grids at the PWC-8 optimum. Rows correspond to targets (Regular 3$\times$3, Perturbed~A, Perturbed~B); columns are time slices. Each panel overlays exact/target samples (blue), predicted \(\hat y_t\) (green), and current states \(x_t\) (orange), shown in this order.}
    \label{fig:sharp-snaps}
  \end{subfigure}

  \caption{Regularized sharpness design with PWC-\(\beta\) optimization via coordinated descent advanced hierarchically -- i.e. progressing optimizing Eq.~(\ref{eq:sharpness-reg}) over parameter within PWC-1 and than using optimal as a warm start for PWC-2, etc up to level $8$: 1\(\to\)2\(\to\)4\(\to\)8).}
  \label{fig:sharpness-3in1}
\end{figure}

As seen in Fig.~(\ref{fig:sharpness-3in1}) minimizing the regularized objective returns sensible, $\beta(t)=O(1)$. Notice (a) non-monotonicity in the resulting (optimal) $\beta^*(t)$ profiles; (b) relatively small change in the auto-correlation profile when we transition from $\beta$-const optimal to $\beta$-PWC-optimal; (c) variation in the optimal $\beta$-profile across the three GMMs.

\subsection{Langevin Mismatch}\label{sec:langevin-mismatch-exp}

\begin{figure}[t]
  \centering
    \includegraphics[width=\linewidth]{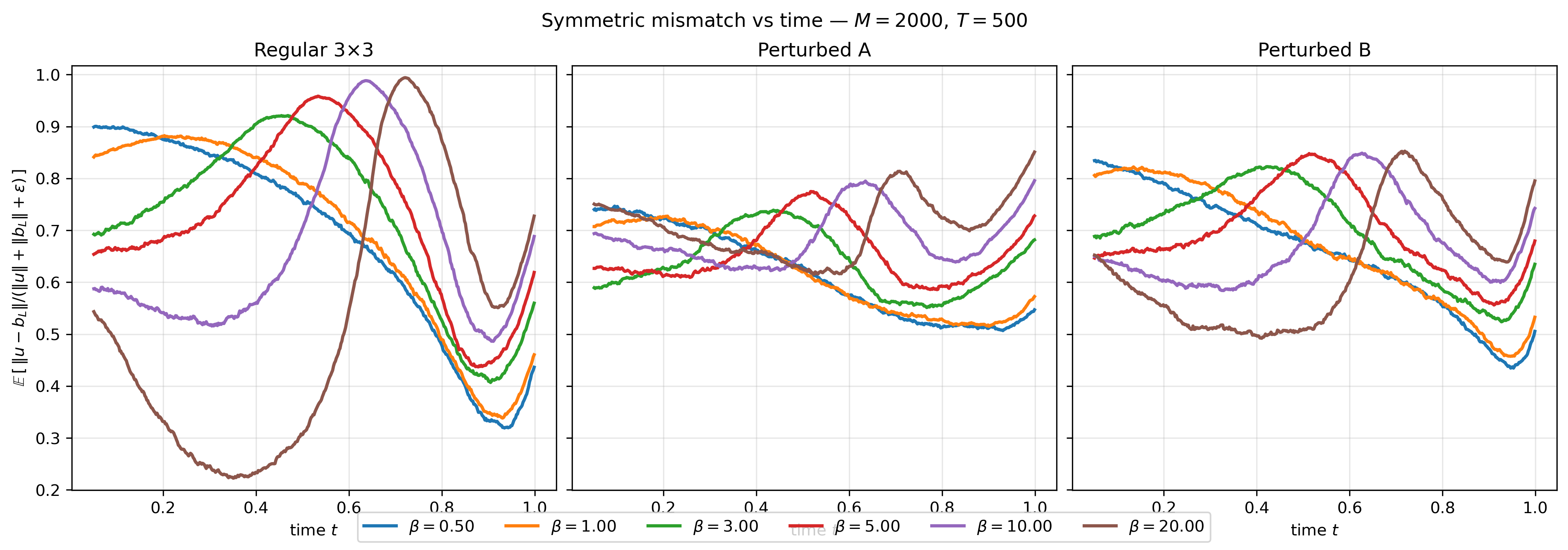}
    \includegraphics[width=\linewidth]{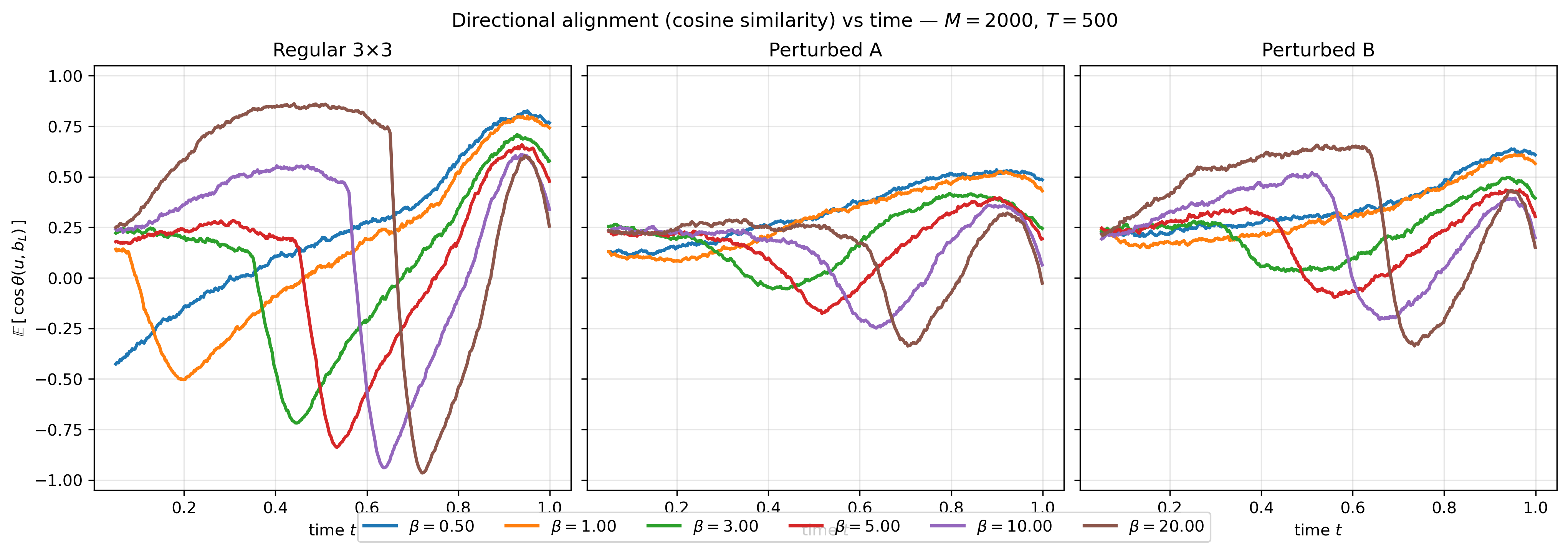}
  \caption{AdaPID--Langevin relative alignment over time for three GMMs. 
  Each curve corresponds to a different constant $\beta$ (shared color coding across subfigures). 
  {\bf Top figure} reports the \emph{symmetric mismatch} $\rho_{\mathrm{sym}}(t;\beta)\in[0,1]$, which measures the relative drift discrepancy
  $\|\Lambda_t\|/(\|u^*\|+\|b_L\|+\varepsilon)$ with $\Lambda_t=u^*-b_L$; lower is better (perfect agreement $\to 0$).
  {\bf Bottom figure} shows the \emph{directional alignment} $\mathrm{cos}(t;\beta)\in[-1,1]$, the average cosine between the AdaPID drift $u^*(x_t,t;\beta)$ and the reference Langevin drift $b_L(x_t)$; higher is better ($\to 1$ for perfect alignment). Formal definitions of $u^*$, $b_L$, and the metrics are given in Section~\ref{sec:langevin-mismatch}.}
  \label{fig:langevin-mismatch-main}
\end{figure}

Fig.~(\ref{fig:langevin-mismatch-main}) reports results of the experiments set up in Section \ref{sec:langevin-mismatch}. Our main observation here is that AdaPID and Langevin are significantly different and moreover the type of mismatch between Langevin and AdaPID is strongly dependent on $\beta$ and also on the GMM. We observe that some alignment at earlier times change to misalignment at intermediate time and weaker alignement at latest times is typical for AdaPID at large $\beta$. Behavior of AdaPID at small $\beta$ is significantly different -- AdaPID dynamics is strongly misaligned at earlier time with the misalignment decreasing with time and then turning into better and better alignement as time increases -- the alignement is the best at the terminal time in this case. 

\subsection{Energy Efficiency}\label{sec:energy-eff-experiment}

\begin{figure}[t]
  \centering
  \subfloat[Convergence and final schedules]{%
    \includegraphics[width=.95\linewidth]{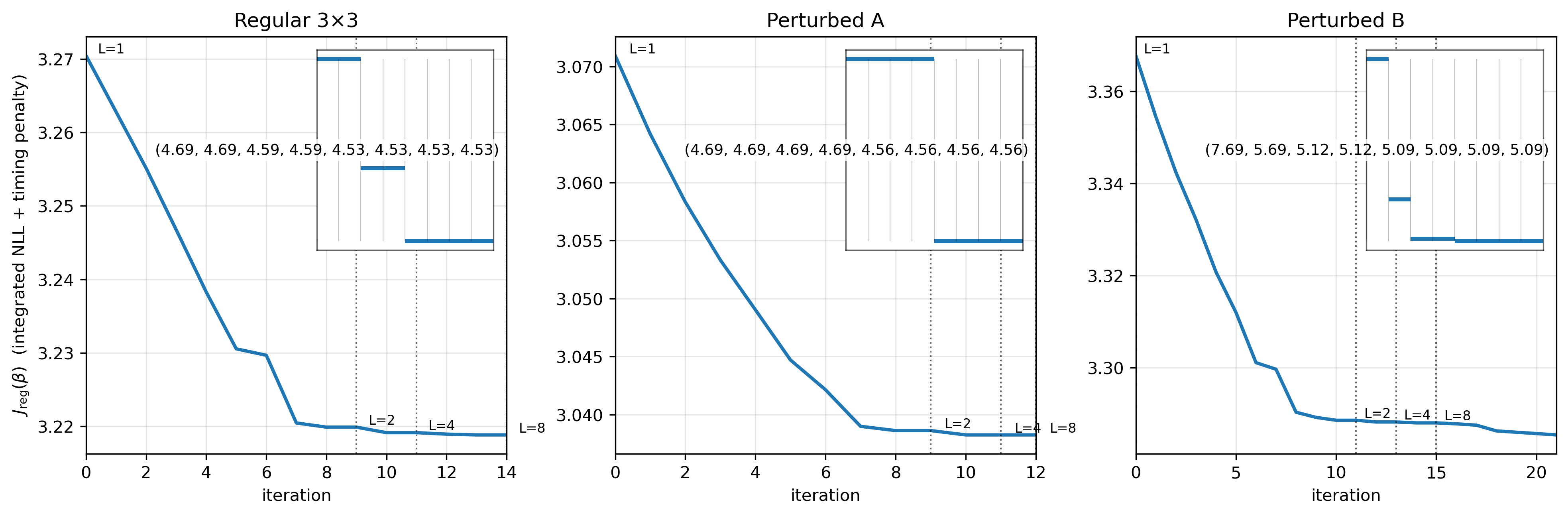}
  }\quad
  \\
  \subfloat[Optimized snapshot grids at final level]{%
    \includegraphics[width=.95\linewidth]{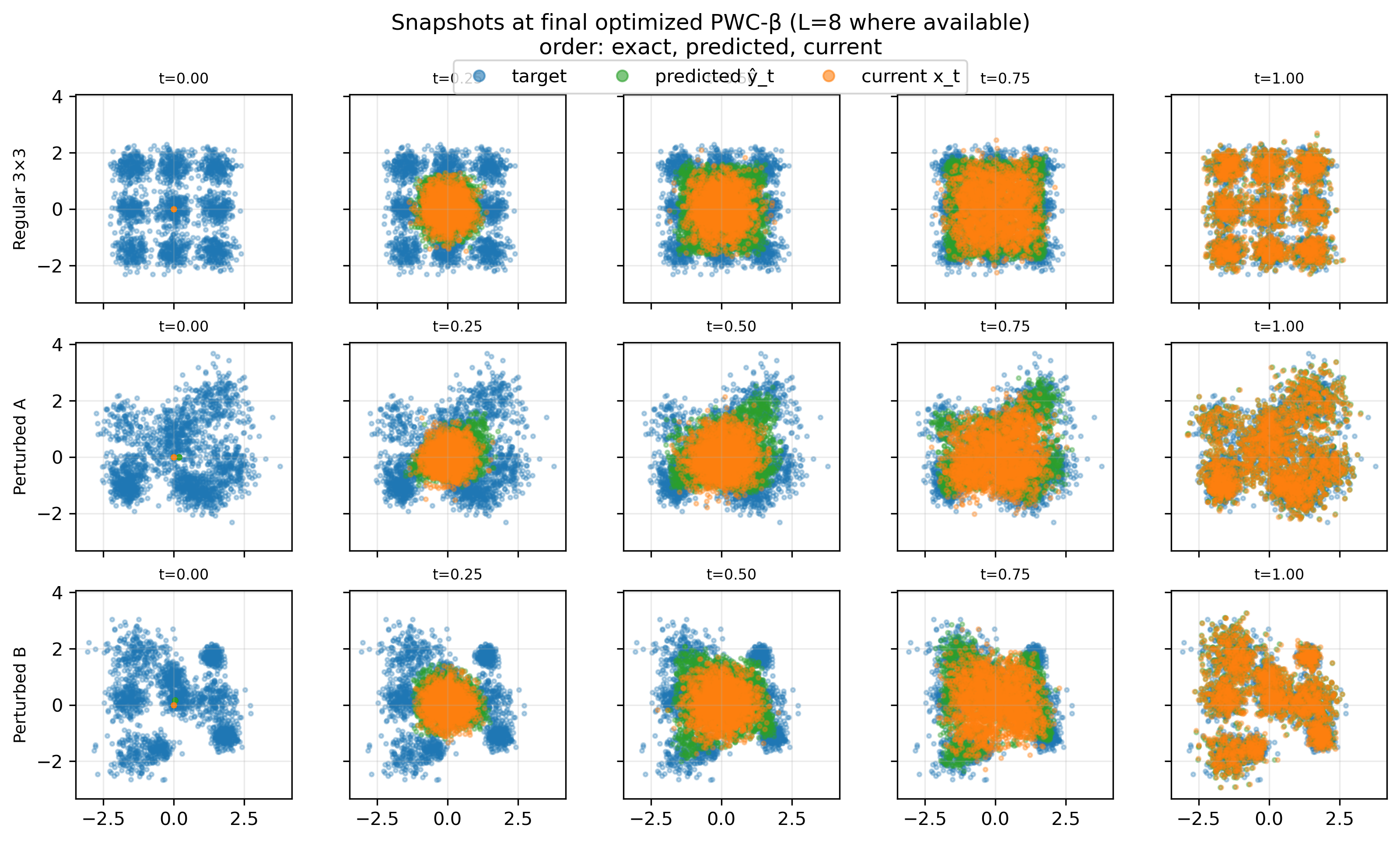}
  }
  \caption{Energy-efficiency experiments with regularized objective \eqref{eq:energy-regularized} with $\lambda_{\mathrm{time}}=10,\ \lambda_{\mathrm{time}}=1/2$ and $\mathcal A^\ast=\tfrac12$. (a) Convergence over coordinate-descent iterations at levels \(L=1,2,4,8\) (three subplots for three targets). Each subplot includes a top-right inset showing the final PWC-\(\beta\) profile at the last level.  (b) For the same final schedules, 3\(\times\)5 snapshot grids (rows=targets; columns=time) show the target samples (blue), predicted \(\hat y_t\) (green), and current \(x_t\) (orange).}
  \label{fig:energy-eff-results}
\end{figure}

We evaluate the regularized objective \eqref{eq:energy-regularized} under a hierarchical PWC schedule refinement (levels \(L=1\to2\to4\to8\)).  The \emph{energy} term \(J_E\) is computed as the time integral of the expected negative log target density along the path.  The timing penalty uses \(t^{(\mathrm{trans})}=1/2\) and a scalar weight \(\lambda_{\mathrm{time}}=10\). The results are shown in Fig.~(\ref{fig:energy-eff-results}). At the optimum \(t^\ast\) -- extracted from the predicted auto-correlation \(\hat{\mathcal A}(t;\beta_{0\to 1})\) (Eq.~\eqref{eq:A-hat}) as the first crossing of \(\mathcal A^\ast=\tfrac12\) -- are $t^\ast\approx 0.51$ for [Regular $3\times 3$] model;  $t^\ast\approx 0.49$ for [Perturbed A] model; and $t^\ast\approx 0.53$ for [Perturbed B] model.

\subsection{Speciation Transient}\label{sec:Speciation-experiment}

\begin{figure}[t]
    \includegraphics[width=\linewidth]{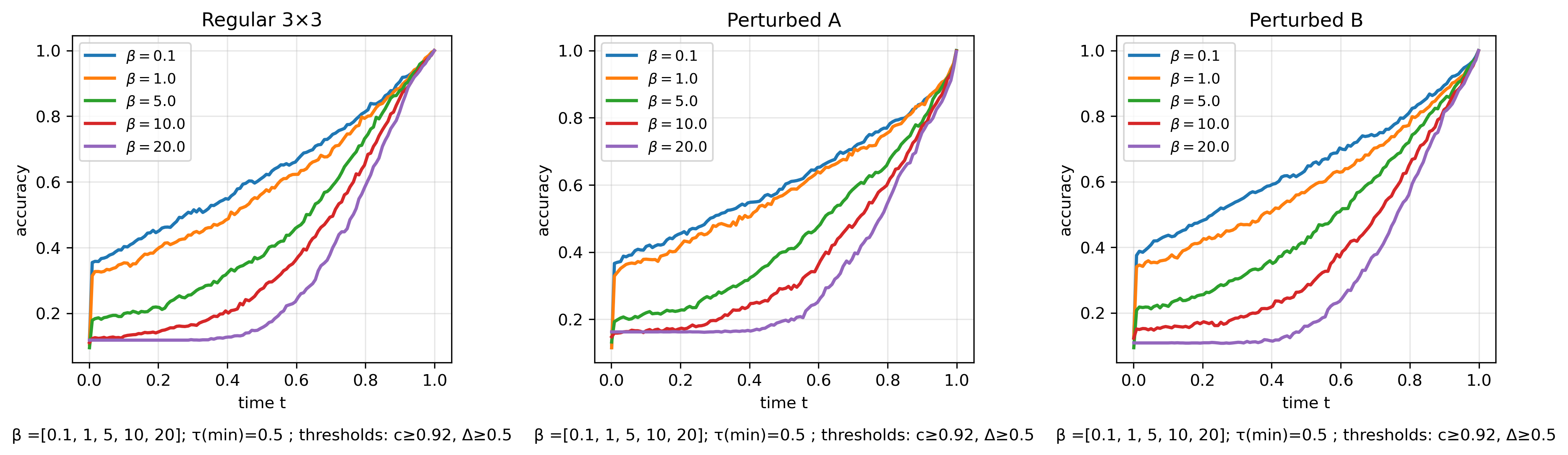}
    \caption{Accuracy vs.\ time.}
    \includegraphics[width=\linewidth]{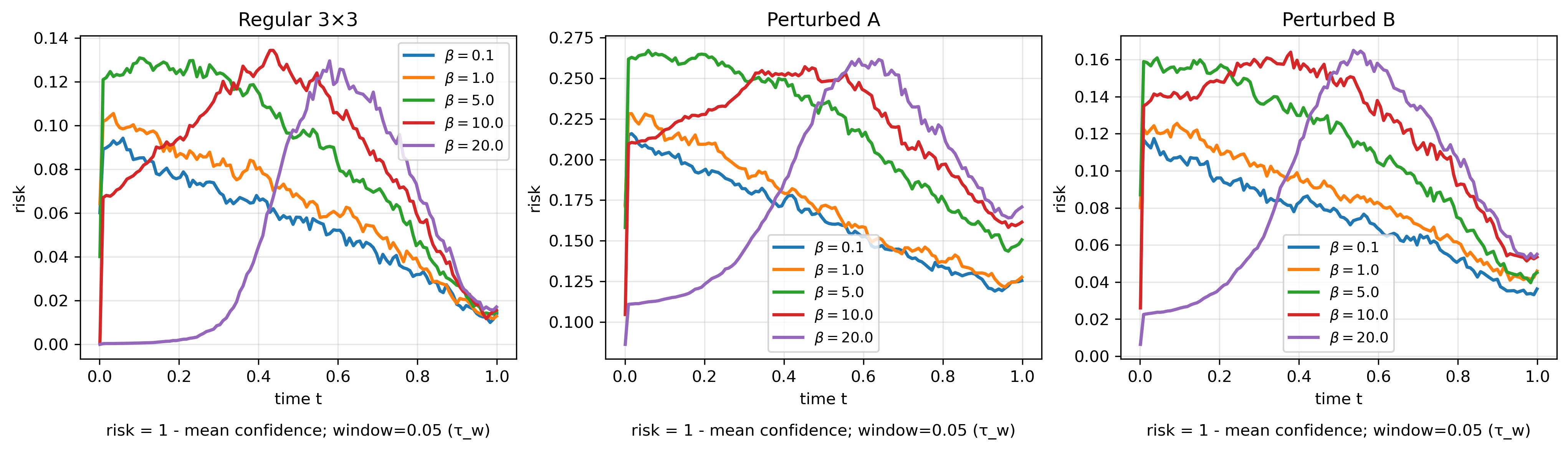}
    \caption{Risk \(=1-\) mean confidence vs.\ time.}
    \includegraphics[width=\linewidth]{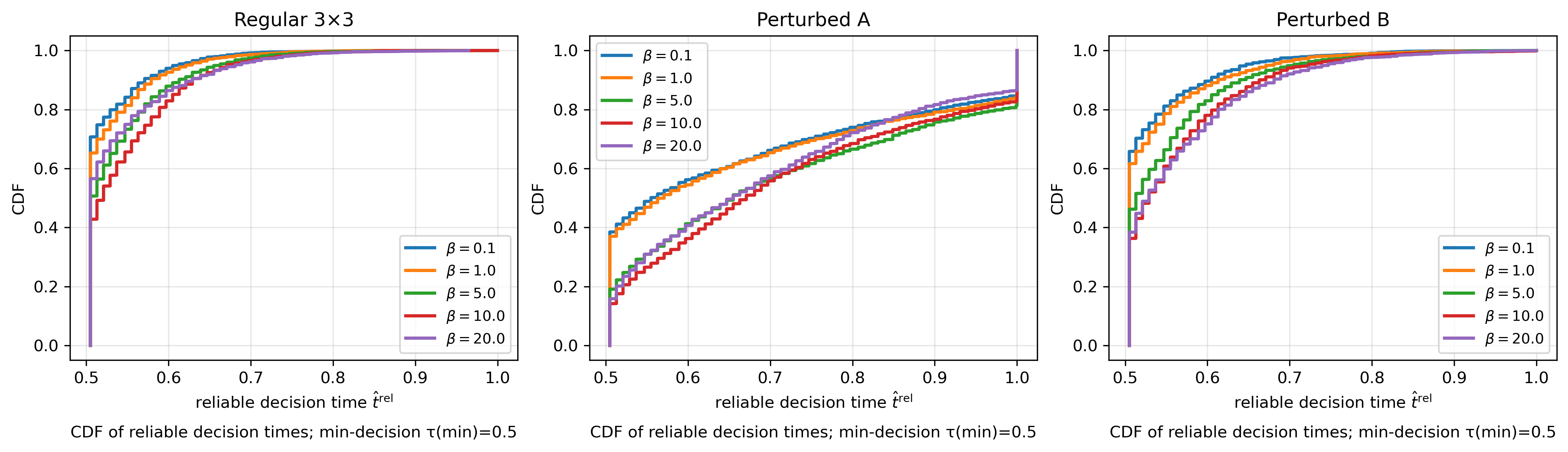}
    \caption{CDF of reliable decision time \(\hat t^{\mathrm{rel}}\).}
  \caption{%
  Early label identification under constant \(\beta\). Each row shows results for three targets (columns: Regular \(3\times 3\), Perturbed~A, Perturbed~B) with \(\beta\in\{0.1,1,5,10,20\}\) overlaid in distinct colors. Prediction–based responsibilities \(r_k^{(\mathrm{pred})}(t;x_t)\) (Sec.~\ref{sec:Speciation}) yield confidence \(c(t)=\max_k r_k^{(\mathrm{pred})}\), margin \(\Upsilon(t)\), and entropy \(H(t)\). A reliable decision is declared at \(\hat t^{\mathrm{rel}}\ge \tau\) when the predicted label is stable over a window \(\tau_w\) and thresholds are met: \(c(t)\ge c_\star\), \(\Upsilon(t)\ge \Upsilon_\star\), \(H(t)\le H_\star\). Here \(\tau=0.5\), \(\tau_w=0.05\), \(c_\star=0.92\), \(\Upsilon_\star=0.5\), and \(H_\star\) equals the entropy of \((c_\star,1-c_\star)\). (Top Row) Accuracy; (Middle Row) risk \(=1-\mathbb{E}[c(t)]\); (Bottom Row) empirical CDF of \(\hat t^{\mathrm{rel}}\). Smaller \(\beta\) accelerates early identification, while larger \(\beta\) delays it.%
  }
  \label{fig:speciation-3panels}
\end{figure}

Fig.~(\ref{fig:speciation-3panels}) reports analysis of the label (specie type) identification experiments set up in Section \ref{sec:Speciation}. We observe that smaller \(\beta\) is systematically better in early identification. In the \(d=2\) setting discussed in the paper the transition from no identification to reliable identification is not sharp -- thus we call it \emph{\bf speciation transient}. For the Regular \(3\times 3\) and Perturbed~B targets, the CDF at \(t=0.5\) is about \(0.6\) and approaches \(1\) by \(t=1\) roughly exponentially. For Perturbed~A, the CDF at \(t=0.5\) is about \(0.4\) and relaxes (also approximately exponentially) but saturates near \(\approx 0.8\) at \(t=1\),
indicating a subset of trajectories that do not satisfy the reliability thresholds within the horizon (presumably because of a significant overlap of modes).

\section{Discussions}\label{sec:discussions}

\paragraph{What did we learn?} Across all diagnostics we tested, the choice of $\beta_{0\to 1}$ matters -- both for terminal sample quality and, crucially, for how the target distribution is assembled in time. Our study benchmarked constant schedules against PWC staircases learned by a hierarchical refinement $L\!=\!1\!\to\!2\!\to\!4\!\to\!8$ under objectives drawn from the Quality-of-Sampling (QoS) suite (Wasserstein/CE-type endpoints and several dynamical surrogates). The optimization recipe is simple in practice: start from a strong constant baseline, warm-start the PWC schedule with that constant value, and refine dyadically while reusing common randomness to stabilize comparisons. This protocol is computationally modest and proved robust across all three targets (Regular $3\!\times\!3$, Perturbed A, Perturbed B). 

\paragraph{Time matters as much as endpoints.} Entropic QoS criteria (terminal comparisons to the target law) are informative but incomplete: two schedules with similar terminal metrics can differ widely in their path-wise behavior, e.g., alignment, stiffness, or transition timing. The paper’s diagnostics explicitly separate endpoint quality from temporal organization; this is consistent with our framework where trajectory-level quantities are primary and time aggregation is a design choice. 

\paragraph{Hierarchical PWC schedules reliably improve dynamical surrogates.} On multiple objectives we consistently observed that the learned PWC-$8$ schedules outperform constant $\beta$ baselines on time-aware metrics while preserving (or improving) terminal quality. The CRN-based search and stepwise refinement keep the procedure stable and inexpensive; we recommend reporting both the learned staircase and the associated time-series diagnostics next to the constant-$\beta^\star$ reference. 

\paragraph{AdaPID is \emph{not} Langevin -- and the mismatch depends on $\beta$.} Alignment experiments against the unit-diffusion Langevin drift show a non-trivial $\beta$-dependence. For large $\beta$ we often see early alignment that degrades mid-horizon and weakens again near the end; for small $\beta$ the pattern flips: strong early misalignment gives way to steadily improved alignment by $t\!=\!1$. These systematic trends confirm that AdaPID follows a different transport strategy than detailed-balance MCMC, and that schedule tuning can steer where and when we desire the the two methods to agree or disagree. 

\paragraph{Energy/timing trade-offs can be shaped explicitly.} When we optimize a regularized objective that couples path ``energy'' with a simple timing penalty (favoring a mid-horizon transition), the learned PWC schedules place the transition close to $t\!\approx\!0.5$ across all targets ($t^\ast\!\approx\!0.49$–$0.53$). This shows we can \emph{program} when the major structural change in the flow occurs without sacrificing sample quality. 

\paragraph{Speciation: earlier is easier with smaller $\beta$.} Label-identification experiments (reliable decision times) reveal that smaller $\beta$ systematically enables earlier speciation. In $d\!=\!2$ the transition is more of a gradual transient than a sharp phase change: the Regular and Perturbed B models reach $\mathrm{CDF}(\hat t^{\mathrm{rel}}\!\le\!0.5)\!\approx\!0.6$ and approach one by $t\!=\!1$, while Perturbed A saturates near $\approx\!0.8$—consistent with stronger mode overlap. These results highlight a practical knob: if early labeling is valuable, favor schedules with lower effective stiffness early on. 

\paragraph{Takeaways for practice.} (i) \textbf{Start constant, then refine.} A good constant $\beta$ is a strong baseline and a reliable warm-start for PWC. (ii) \textbf{Report time series, not just endpoints.} Many improvements appear in the \emph{path}, not in the terminal snapshot. (iii) \textbf{Match schedule to task.} If you care about early speciation, prefer lower early $\beta$; if you aim to synchronize a structural transition, add a timing term and learn the staircase accordingly. (iv) \textbf{Reuse randomness.} Common-randomness comparisons are essential for stable optimization and fair reporting. 

\paragraph{Limitations and outlook.} Our experiments focused on 2D GMMs to leverage an oracle for careful validation, and on PWC schedules for simplicity and exact per-interval Green-function evaluation. Extending to higher dimensions and richer targets will require scalable estimators and possibly smoother parameterizations of $\beta_t$; nevertheless, the linearly-solvable PID machinery and PWC matching arguments carry over to time-varying quadratic potentials, keeping the approach tractable. See Section \ref{sec:outlook} for deeper outlook. 

\paragraph{Broader perspective.} By framing sampling as a linearly-solvable stochastic control problem, AdaPID provides a \emph{programmable} way to shape both terminal quality and the temporal assembly of the distribution. Schedules are levers: they do not merely speed up or slow down dynamics—they determine \emph{when} and \emph{how} structure emerges. The evidence across our QoS suite suggests that schedule design—constant for baselines, PWC for control—should be treated as a first-class component of practical samplers.

\section{Outlook}\label{sec:outlook}

\paragraph{Scaling in dimension and multi-modality.}
Two natural axes for stress–testing PID are higher ambient dimension $d\gg 1$ and larger numbers of modes $N$ in the target mixture. A first line of inquiry is scaling of QoS with sample size: for Wasserstein distances (and CE-type criteria), characterize finite-sample rates $W_p(\hat p_M,p^{(\mathrm{tar})})$ as functions of $(d,N,M)$ under PID dynamics, with explicit constants that depend on the schedule $\beta_{0\to 1}$. Empirically, we propose controlled sweeps in $(d,N)$ at fixed compute to quantify how steep the sensitivity to $\beta_t$ becomes as geometry and overlap worsen, and whether the optimal schedule families (constant vs.\ PWC) change regime with $d$ or $N$.

\paragraph{How complex should schedules be?} A central design question is the schedule complexity needed for accuracy. Beyond our hierarchical $L\!=\!1\!\to\!2\!\to\!4\!\to\!8$ refinement, we envision two complementary studies: (i)~an ablation/decimation experiment, starting from an over–parameterized PWC-$K$ and greedily pruning intervals while monitoring validation QoS (e.g., Wasserstein and dynamical surrogates); and (ii)~ smooth parameterizations of $\beta_t$ (e.g., via neural networks) trained with the same objectives used in the paper. This can clarify whether the staircase is merely a convenient optimizer or intrinsically well matched to linearly-solvable PID.

\paragraph{Theory: optimal schedules and Robust Wasserstein Profile Inference (RWPI).} Two complementary theory thrusts naturally extend this work. First, \emph{statistical guarantees and optimal paths}: viewing PID as a controllable transport induces finite-sample rates and stability bounds that depend explicitly on schedule complexity and path regularity. Recent analyses for flow/diffusion learners provide non-asymptotic guarantees and show that restricting temporal degrees of freedom and minimizing spatial Lipschitz constants along the path (see Sections \ref{sec:grad},\ref{sec:grad-exp}) can sharpen constants and even improve rates; this motivates schedule design (e.g., Lipschitz-minimizing or variance-balancing protocols) and ablations on PWC resolution \cite{marzouk_data_2025,chen_lipschitz-guided_2025,benton_error_2023}. Second, \emph{RWPI} supplies a statistically calibrated, task-aware evaluation and correction layer: by projecting the terminal law onto hypothesis manifolds (moments, calibration, conditional targets) in Wasserstein geometry, one obtains valid tests with $n^{-1}$ scaling for the induced projection radius, interpretable dual certificates, and principled DRO radii—amenable to causal/adapted extensions for pathwise constraints \cite{blanchet_robust_2019,acciaio_causal_2020,backhoff-veraguas_adapted_2020}. Together, these directions aim at (i) schedule/path designs with provably optimal (or near-optimal) sample complexity under PID, and (ii) RWPI-driven stress tests and corrections that integrate seamlessly with PID’s control to deliver dimension-aware guarantees and decision-relevant robustness.

\paragraph{Modeling the target beyond analytic GMMs.} GMMs are invaluable because they admit closed-form oracles (score, posteriors), enabling precise diagnostics and ablations. The next step is to test when a \emph{learned} mixture (Gaussian or from a broader exponential family) remains a competitive \emph{surrogate} for complex targets. Two practical questions arise: 
(i)~is automatic differentiation on the fitted mixture sufficient for efficient PID learning, or (ii)~do we benefit from \emph{NN-based surrogates} (score nets, normalizing flows) when the mixture fit is poor? A hybrid pipeline -- mixture fits where accurate, neural surrogates elsewhere—may preserve interpretability while extending coverage.

\paragraph{Positioning relative to neural score pipelines.}
\textbf{Classical annealing and tempering.} Simulated–annealing MCMC (SA–MCMC) \cite{kirkpatrick_optimization_1983} and annealed/tempered importance–sampling chains \cite{neal_annealed_2001} cool a temperature (or anneal a bridge) to move from an easy reference to the target (see also \cite{doucet_tutorial_2011,chopin_introduction_2020}). They are broadly applicable but notoriously suffer from slow equilibration, mode–hopping barriers, and delicate schedule design (temperature ladders, swap rates); guarantees typically hold only under strong mixing assumptions. Sequential/tempered MCMC \cite{del_moral_sequential_2006} extends this idea by mutating and reweighting a particle population along a likelihood or temperature path; it improves robustness but still inherits path–design and mixing bottlenecks.

\textbf{AI–era score/flow pipelines.} Path–integral samplers \cite{zhang_diffusion_2021}, stochastic–interpolant flows \cite{albergo_stochastic_2023}, and IDEM (Integrated Denoising–Energy Matching) \cite{akhound-sadegh_iterated_2024} learn a continuous field that transports a reference to the data.  Path–integral and stochastic–interpolant methods parameterize an interpolation (or velocity) between endpoints and fit it by regression of score/velocity functionals; they deliver strong empirical performance and amortized sampling, yet the learned vector fields are typically unconstrained (non–integrable, hard to analyze), and stability/accuracy hinges on network capacity and training losses.  IDEM explicitly ties the vector field to an energy model, enforcing conservativity by learning $E_\theta$ whose gradient matches denoising targets across noise scales; this yields a principled potential but still relies on large neural networks and multi–scale training, with theoretical guarantees mostly asymptotic.

\textbf{Where PID sits.} PID provides \emph{integrable, linear–operator} drift–diffusions with explicit Green–function ratios, closed–form schedule–to–drift maps, and risk–sensitive tilting. This yields (i) tractable correctness proofs, (ii) non–asymptotic sample–complexity analysis, and (iii) protocol design tools ($\beta_{0\to1}$ as a control knob) that are hard to obtain for generic neural score fields. In our experiments, these properties translate into predictable trade–offs (e.g., between kinetic cost, Wasserstein error, and speciation timing) and reproducible improvements from piecewise–constant schedules.

\textbf{Critical comparison and outlook.} \emph{Strengths of PID:} analytic structure (no black–box scores), stable integrators, cheap per–step evaluation, and interpretable protocols; one can \emph{optimize} $\beta_t$ for a stated objective (Wasserstein, sharpness, speciation timing), rather than hope the network implicitly discovers it. 
\emph{Limitations today:} quadratic baseline potentials and schedule families can be misspecified for highly anisotropic or strongly non–Gaussian targets; performance depends on protocol tuning (number of PWC knots, guard–rails), and some results lean on mixture–oracle structure (readily available for GMMs but not for arbitrary data). 
\emph{Bridging the gap:} promising directions include (i) richer yet still–integrable potentials (guided quadratics, piecewise–quadratic in space or time), (ii) hybrid PID–score schemes that keep PID’s Green–ratio backbone while learning low–dimensional corrections to the velocity field, (iii) data–driven protocol learning (neural parametrizations of $\beta_t$ with PID constraints), and (iv) systematic ablations on PWC resolution to quantify how many knots suffice for a target QoS (e.g., Wasserstein). Compared to SA/tempering/sequential MCMC, PID retains explicit path–design and often achieves faster, smoother transitions; compared to neural score pipelines, PID currently trades universal approximation for verifiability and control---a trade we plan to relax by expanding the integrable family and adding light–weight learned components under PID’s stability guarantees.

Going forward, we will compare \emph{end-to-end cost} (wall-clock to target QoS), \emph{data/compute footprint}, and \emph{robustness} under domain shift. Because PID's schedules are interpretable levers, head-to-head studies can also report \emph{when} structure emerges (speciation timing, alignment), not just terminal fit.

\paragraph{Methodological extensions of PID.}
We see several high-priority directions:
\begin{enumerate}
\item \textbf{Shifted quadratic guidance.} Extend to $V_t(x)=\beta_t \|x-\nu_t\|^2/2$ with time-varying centers $\nu_t$ (data-driven or learned). This retains integrability while enabling guided transport and path sampling in $2$D/3D (e.g., conditioning or tracking moving manifolds).

\item \textbf{Interacting particles.} Study PID with weak interactions to couple trajectories (e.g., repulsive terms for diversity or consensus terms for variance control). Analyze mean-field limits and their effect on QoS and energy budgets.

\item \textbf{Beyond harmonic, still integrable.} Explore spatially piecewise-constant or piecewise-quadratic potentials where interval-wise Green’s functions remain explicit. This tests how far integrability can be pushed without neural surrogates.

\item \textbf{Non-zero drifts and gauges.} Incorporate known drift fields (e.g., physics priors) or gauge freedoms to bias transport while preserving solvability. 
\end{enumerate}

\paragraph{Evaluation protocols and compute realism.} To make results portable, we advocate reporting: (i)~terminal and temporal QoS with confidence bands; (ii)~speciation timing distributions; (iii)~energy/actuation and stability diagnostics; and (iv)~wall-clock vs.\ accuracy curves under fixed hardware budgets. Such standardized reporting would clarify the trade-offs between schedule complexity, model surrogates, and ultimate sample quality.

\paragraph{Summary.}
The agenda above aims to keep the distinctive advantages of PID -- closed-form operators and programmable schedules -- while expanding along three fronts: \emph{scale} (large $d,N$), \emph{expressivity} (richer $\beta_t$ and target surrogates), and \emph{realism} (compute-aware evaluation and interaction effects). We view schedule design as a first-class control problem whose complexity should be commensurate with the target’s geometry and the practitioner’s objective.

\section*{Acknowledgments}

We are thankful to Jos\'e Blanchet and Youssef Marzouk for stimulating discussions and for their guidance on the broader development of PID and related methodologies. This work was partially supported by MC’s start-up funds at the University of Arizona. A substantial portion of the research was conducted while MC was partially supported by the Mini-Sabbatical Faculty Fellow Program at Lawrence Livermore National Laboratory (LLNL). MC gratefully acknowledges the support, scientific engagement, and encouragement of colleagues in the Center for Applied Scientific Computing (CASC) at LLNL.

In the interest of reproducibility, all Python/Jupyter code used to generate the figures and experiments reported in this paper is openly available at \href{https://github.com/mchertkov/AdaPID}{github.com/mchertkov/AdaPID}.

The authors used large language models to assist with editing, code refactoring, and the organization of computational experiments. All mathematical derivations, scientific claims, and final code were independently checked, validated, and approved by the authors.

\printbibliography

\newpage

\appendix 

\section{Representations of Probability Densities}\label{sec:densities}

To set up a proper metric for this paper main goal -- {\bf choosing optimal $\beta_{0\to 1}$} -- we utilize and juxtapose to each other the following (probability) densities -- all defined for a given (fixed) $\beta_{0\to 1}$:
\begin{itemize}
    \item The {\bf target} density itself -- $p^{(\text{\tiny tar})}(y)$ defined  according to Eq.~(\ref{eq:tar}) up to the normalization factor (partition function) by the explicitly given energy function $E(\cdot)$.
    
    \item Probability density of a {\bf path} $p^{(\text{\tiny path})}(x_{0\to 1}|u^*_{0\to 1}(\cdot;\beta_{0\to 1}))$ generated according to the SODE (\ref{eq:x-SODE}) with $u_{0\to 1}(x_{0\to 1})\doteq (u_t(x_t)| t\in[0,1])$ substituted by the optimal protocol (conditioned to $\beta_{0\to 1}$) and defined in Eq.~(\ref{eq:u*}). We will also be working with {\bf marginal} density of the state $x_t$ observed at the moment of time $t$ and thus defined according to the following path integral
    \begin{equation}\label{eq:fin-PathInt}
    p^{(\text{\tiny mar})}(x_t|t;u^*_{0\to t}(\cdot;\beta_{0\to 1}))=\int p^{(\text{\tiny path})}(x_{0\to 1}|u^*_{0\to 1}(\cdot;\beta_{0\to 1})) {\cal D}x_{0^+\to t^{-}},
    \end{equation}
    where  we utilize the standard (in theoretical physics) notations for the path-integral differential 
    \[{\cal D}x_{0^+\to t^{-}}=\lim_{N\to \infty} \prod_{n=1}^{N-1} d x_{t_n},\quad t_n=t\frac{n}{N}.\]
    Given dependence of the definitions of the path density and its marginals on the SODE (\ref{eq:x-SODE}) -- the densities will be represented empirically via generated sample paths.

    \item Density of the final state as {\bf probe} at the time $t$, given that the system is observed in the state $x$, is another important object of the H-PID theory:
\begin{equation}\label{eq:tar-probe}
    p^{(\text{\tiny probe})}(\cdot |t;x;\beta_{0\to 1})=\frac{q^{(\text{\tiny probe})}(\cdot|t;x;\beta_{0\to 1})}{Z^{(\text{\tiny probe})}(t;x;\beta_{0\to 1})}.
\end{equation}
Note that according to Eqs.~(\ref{eq:cond},\ref{eq:Delta}): 
\[\lim_{t\to 1} p^{(\text{\tiny probe})}(y|t;x;\beta_{0\to 1})=\delta(x-y),\quad \lim_{t\to 0} p^{(\text{\tiny probe})}(y|t;x;\beta_{0\to 1})=p^{(\text{\tiny tar})}(y).\] 
In other words this density is certainly not a good approximation for the target density at larger $t$, and especially at $t\to 1$. This suggests to also consider {\bf integrated probe} density 
\begin{equation}\label{eq:i-probe}
    p^{(\text{\tiny i-probe})}(\cdot |t;\beta_{0\to 1})=\int p^{(\text{\tiny mar})}(x|t;u^*_{0\to t}(\cdot;\beta_{0\to 1}))\frac{q^{(\text{\tiny probe})}(\cdot|t;x;\beta_{0\to 1})}{Z^{(\text{\tiny probe})}(t;x;\beta_{0\to 1})} dx,
\end{equation}
for which the asymptotics become
\[\lim_{t\to 1} p^{(\text{\tiny i-probe})}(y|t;\beta_{0\to 1})=\lim_{t\to 0} p^{(\text{\tiny i-probe})}(y|t;\beta_{0\to 1})=p^{(\text{\tiny tar})}(y).\] 

\item Let us define {\bf expectation} (average value) of the final state (conditioned to the observation of $x$ at time $t$):
\begin{equation}\label{eq:y-hat}
    \hat y(t;x;\beta_{0\to 1})\doteq  \mathbb{E}_{y\sim p^{(\text{\tiny probe})}(\cdot|t;x;\beta_{0\to 1})} \left[y\right]=
    \int_{\mathbb R^{d}} y p^{(\text{\tiny probe})}(y|t;x;\beta_{0\to 1})dy.
\end{equation}
The expected final state was called in \cite{behjoo_harmonic_2025} an "order-parameter" because in analysis of auto-correlations it shows sharper (than for the original $x$) transition from de-correlated case to what seems to be relatively close structurally to the finite generated sample. 

Next, it is natural to introduce what we call {\bf predicted} density of the {\bf expected} state 
\begin{align}\label{eq:pred-expected}
    p^{(\text{\tiny pred-e})}(y |t;\beta_{0\to 1}) & =\int \delta\left(y-\hat y(t;x;\beta_{0\to 1})\right)p^{(\text{\tiny mar})}(x|t;u^*_{0\to t}(\cdot;\beta_{0\to 1})) dx 
\end{align}
We will see below that these densities will be useful for understanding quality of the final state as we {\bf grow } it in time with H-PID. Let us now discuss how we can sample from $p^{(\text{\tiny pred-e})}(y |t;\beta_{0\to 1})$ (or from $p^{(\text{\tiny pred-r-e})}(y |t;\beta_{0\to 1})$). The sampling is done in two steps: first, sample a path and record the state $x$  where it arrives at the moment of time $t$, and then, second, evaluate the function $\hat y(t;x;\beta_{0\to 1})$ (or $\hat y(t;x;\beta_{0\to 1})$) at these $t$ and $x$.

\item Gaussian density built from $\Delta(t;x;y;\beta_{0\to 1})$ -- which is a quadratic form in $x$ and $y$ with coefficients dependent on time defined in Eq.~(\ref{eq:Delta}) becomes
\begin{align}
\label{eq:rho}
\rho(y\!\mid\! t;x;\beta_{0\to 1}) & = \exp\!\bigl(-\Delta(t;x;y;\beta_{0\to 1})\bigr)=\mathcal{N}\!\Bigl(y\;\Big| \nu_t(x)\doteq \frac{b^{(-)}_t}{{\cal K}_t}x, \frac{1}{{\cal K}_t}\mathrm{I}_{d\times d} \Bigr)\\ \nonumber & =\exp\left(-\frac{{\cal K}_t}{2}\left\|y-\frac{b_t^{(-)}}{{\cal K}_t} x\right\|^2+\frac{d}{2}\log\left(\frac{{\cal K}_t}{2\pi}\right)\right),\quad {\cal K}_t\doteq c^{(-)}_t - a^{(+)}_1.
\end{align}
Here $a^{(+)}$, $a^{(-)}$, $b^{(-)}$, and $c^{(-)}$ are the time-(and $\beta_{0\to 1}$) dependent coefficients obtained by solving the coupled Riccati-type ODEs for the harmonic path integral (cf. Sec.~\ref{sec:PID}).  As discussed in  Section \ref{sec:PID-beta-t} in the case of piecewise-constant $\beta_{0\to 1}$ these coefficients are computed exactly in each interval and matched continuously at the interval boundaries.

We call the auxiliary Gaussian density $\rho(\cdot\mid\! \cdot;\cdot;\cdot)$ -- the {\bf re-weighting} density because of its role played in Eq.~(\ref{eq:cond}) -- re-weighting from the target to probe densities. 

\end{itemize}

\section{Use Case: Gaussian Mixture}\label{sec:G-Mix}

We consider the special case when the target distribution is a Gaussian mixture with
component–specific full covariances,
\begin{equation}\label{eq:G-Mix-full}
p^{(\text{\tiny tar})}(y)
\;\to\;
p^{(\text{\tiny tar-G-Mix})}(y)
\;\doteq\;
\sum_{n=1}^N \varrho_n\,
\mathcal{N}\!\big(y;\mu_n,\Sigma_n\big),
\qquad
\sum_{n=1}^N \varrho_n = 1,
\end{equation}
where each \(\Sigma_n \in \mathbb{R}^{d\times d}\) is symmetric positive–definite.

The {\bf re-weighting density} \(\rho(y\!\mid\! t;x)\), defined in Eqs.~(\ref{eq:rho}) (restated here for convenience),
\[
\rho(y\!\mid\! t;x)
=
\mathcal{N}\!\Big(y\;\Big|\;\frac{b_t^{(-)}}{K_t}x,\;\frac{1}{K_t}I_d\Big),
\]
is isotropic with scalar gain \(b_t^{(-)}\) and precision \(K_t=c_t^{(-)}-a_1^{(+)}>0\) determined by the chosen
\(\beta_{0\to 1}\)–protocol, with \(a^{(\pm)}_t,b^{(-)}_t,c_t^{(-)}\) solving ODEs~(\ref{eq:abcpm}) subject to the boundary conditions (\ref{eq:-asympt},\ref{eq:+asympt}).

\paragraph{Probe density as a product of experts.}
The {\bf probe} (a product of experts: target \(\times\) re-weighting) is defined by
Eqs.~(\ref{eq:cond},\ref{eq:tar-probe}) and restated as
\[
p^{(\text{\tiny probe})}(y\mid t;x)
= \frac{q^{(\text{\tiny probe})}(y\mid t;x)}{Z^{(\text{\tiny probe})}(t;x)}
= \frac{p^{(\text{\tiny tar})}(y)\,\rho(y\!\mid\! t;x)}
       {\int p^{(\text{\tiny tar})}(y')\,\rho(y'\!\mid\! t;x)\,dy'}.
\]
Plugging in \eqref{eq:G-Mix-full} and \(\rho\) yields the (unnormalized) mixture–Gaussian product
\begin{align} \nonumber 
q^{(\text{\tiny probe-G-Mix})}(y\mid t;x)
\; & \propto\;
\mathcal{N}\!\Big(y\;\Big|\;\nu(t;x)=\frac{b_t^{(-)}}{K_t}x,\;\frac{1}{K_t}I_d\Big)
\;\sum_{n=1}^N \varrho_n\,\mathcal{N}\!\big(y;\mu_n,\Sigma_n\big),\\
\label{eq:probe-G-Mix} & \propto \sum_{n=1}^N \varrho_n\,w_n(t;x)\,\mathcal{N}\!\big(y;\tilde{\mu}_n(t;x),\tilde{\Sigma}_n(t)\big),
\end{align}
where the newly introduced variance, mean and mixing weights are
\begin{align} \label{eq:Gmix-post-cov-full}
\tilde\Sigma_n(t) & 
\;\doteq\;\Big(\Sigma_n^{-1} + K_t\,I_d\Big)^{-1},
\quad \tilde\mu_n(t;x)\;\doteq\;\tilde\Sigma_n(t)\Big(\Sigma_n^{-1}\mu_n + b_t^{(-)}\,x\Big),\\ \label{eq:Gmix-weight-full}
w_n(t;x)\; & \doteq\;
\mathcal{N}\!\Big(\nu(t;x);\,\mu_n,\;\Sigma_n+\tfrac{1}{K_t}I_d\Big)\\ \nonumber &
=\frac{\exp\!\Big(-\tfrac{1}{2}\big(\nu(t;x)-\mu_n\big)^\top
\big(\Sigma_n+\tfrac{1}{K_t}I_d\big)^{-1}\big(\nu(t;x)-\mu_n\big)\Big)}
{\sqrt{(2\pi)^d\,\det\!\big(\Sigma_n+\tfrac{1}{K_t}I_d\big)}}.
\end{align}
(Eq.~(\ref{eq:probe-G-Mix}) is a an example of the Gaussian product identity in action.)

Then {\bf predicted (expected) state map}, defined in \eqref{eq:y-hat} becomes
\begin{align}\label{eq:hat-x-G-mix-full}
\hat{y}(t;x)\; =\; \frac{\int y\,q^{(\text{\tiny probe-G-Mix})}(y\mid t;x)\,dy}
     {\int \;q^{(\text{\tiny probe-G-Mix})}(y\mid t;x)\,dy}\; =\;
\frac{\sum_{n=1}^N \varrho_n\,w_n(t;x)\,\tilde\mu_n(t;x)}{\sum_{n=1}^N \varrho_n\,w_n(t;x)}.
\end{align}

\paragraph{Numerical note (stable evaluation when $d$ is large $K_t$ close to singular and $\Sigma$ is ill-conditioned).} For robustness, avoid explicit matrix inverses and work in Cholesky / log–space.

\smallskip
\emph{Mixture weights in log–space.}
Let \(m_t(x)\!\doteq\!\tfrac{b_t^{(-)}}{K_t}x\) and
\(S_n \doteq \Sigma_n + K_t^{-1} I_d\).
Compute the Cholesky factor \(L_n L_n^\top = S_n\), solve
\(L_n s_n = m_t(x)-\mu_n\), then
\[
\log w_n(t;x)
= -\tfrac{1}{2}\,\|s_n\|_2^2 \;-\; \sum_{i=1}^d \log (L_n)_{ii}
\;-\; \tfrac{d}{2}\log(2\pi).
\]
Normalize the weights with a log-sum-exp:
\(\;\tilde{w}_n \propto \exp\big(\log w_n - \max_j \log w_j\big)\),
so that only ratios of weights are needed (the additive constant cancels in~\eqref{eq:hat-x-G-mix-full}).

\smallskip
\emph{Posterior mean/covariance without \(\Sigma_n^{-1}\).}
Use the identities
\[
\tilde{\Sigma}_n(t)=\big(\Sigma_n^{-1}+K_t I_d\big)^{-1}
=\Sigma_n\big(I_d + K_t \Sigma_n\big)^{-1},\qquad
\tilde{\mu}_n(t;x)
=\big(I_d + K_t \Sigma_n\big)^{-1}\!\big(\mu_n + b_t^{(-)} \Sigma_n x\big).
\]
Factor \(A_n \doteq I_d + K_t \Sigma_n = R_n R_n^\top\) (Cholesky) and obtain
\(\tilde{\mu}_n\) via two triangular solves with \(R_n\); avoid forming
\(\tilde{\Sigma}_n\) explicitly unless needed (then use
\(\tilde{\Sigma}_n = R_n^{-\top} R_n^{-1}\Sigma_n\)).

\section{Exemplary Negative \texorpdfstring{$\beta_t$}{beta}}
\label{sec:negative-beta}

Here we examine the case of harmonic time dependent potential (\ref{eq:V-t-beta-t}) which includes a time sub-interval on which \(\beta_t<0\). The Green functions \(G_t^{(\pm)}\) remain Gaussian with time–dependent quadratic/linear coefficients governed by the Riccati Eqs.~(\ref{eq:Gpm},\ref{eq:abcpm}) which obeys the delta–boundary conditions Eqs.~(\ref{eq:-asympt},\ref{eq:+asympt}).

\subsection*{A three-piece “negative window” profile}

Fix \(B>0\) and \(\delta\in(0,1)\), and consider
\begin{equation}
\beta_t(\delta;B)=
\begin{cases}
 -B,& |t-\tfrac12|<\delta,\\
 0,& \text{otherwise},
\end{cases}
\qquad
t_L\doteq  \frac{1-\delta}{2},\quad
t_R\doteq  \frac{1+\delta}{2}.
\label{eq:beta-negative-window}
\end{equation}
Below we list closed forms for the forward and backward branches, matched continuously at \(t_L,t_R\).

\paragraph{Forward branch \(a_t^{(+)}\).}
\begin{itemize}
\item On \([0,t_L]\) (here \(\beta=0\)): \(a_t^{(+)}=\frac{1}{t}\).

\item On \([t_L,t_R]\) (here \(\beta=-B\)): Integrating \(\dot a=-(B+a^2)\) forward in time, with the initial condition \(a(t_L)=\dfrac{2}{1-\delta}\) (assuming continuity between branches),  gives
\begin{equation}
a_t^{(+)} \;=\; \sqrt{B}\tan\left(C_L- t\sqrt{B}\right),\quad C_L\doteq \arctan\left(\frac{1}{t_L\sqrt{B}}\right)+t_L \sqrt{B}.
\label{eq:a-plus-mid}
\end{equation}
\item On \([t_R,1]\) (back to \(\beta=0\)): 
\[
a_t^{(+)}=\frac{a_{t_R}^{(+)}}{1+a_{t_R}^{(+)}(t-t_R)},
\]
\end{itemize}
with the initial condition for $a_{t_R}^{(+)}$ set (due to continuity) by Eq.~(\ref{eq:a-plus-mid}). 

Well-posedness of the Green-function (positive branch) requires that $a_t^{(+)}>0$ for all $t\in [0,1]$, which is achieved if $a_{t_R}^{(+)}>0$ -- therefore translating in  the following sufficient condition
\begin{equation}
1> t_L\sqrt{B}\tan\left((t_R-t_L)\sqrt{B}\right)= \frac{1-\delta}{2}\sqrt{B}\tan\left(\delta\sqrt{B}\right).
\label{eq:nosing-plus}
\end{equation}

\paragraph{Backward branch \(a_t^{(-)}, b_t^{(-)}, c_t^{(-)}\).}
\begin{itemize}
\item On \([t_R,1)\) (here \(\beta=0\)), the terminal asymptotics yield,  
\(a_t^{(-)}=b_t^{(-)}=c_t^{(-)}=\frac{1}{1-t}\).

\item On \([t_L,t_R]\) (here \(\beta=-B\)): Solving \(\dot a=a^2+B\), \(\dot b=a\,b\), \(\dot c=b^2\) backward from \(t_R\) -- with the initial conditions \(a_{t_R}^{(-)}=b_{t_R}^{(-)}=c_{t_R}^{(-)}=\frac{1}{1-t_R}\) -- gives
\begin{align} \label{eq:a-minus-mid}
    a_t^{(-)} & = \sqrt{B}\tan\left(C_R\!-\! (1\!-\!t)\sqrt{B}\right),\ C_R
    \!\doteq\! \arctan\left(\frac{1}{(1\!-\!t_R)\sqrt{B}}\right)\!+\!(1\!-\!t_R)\sqrt{B},\\
b_t^{(-)} &= \frac{\cos\left(C_R- (1-t_R)\sqrt{B}\right)}{(1-t_R)\cos\left(C_R- (1-t)\sqrt{B}\right)}, \label{eq:b-minus-mid}\\ \nonumber
c_t^{(-)} &= \frac{1}{1-t_R}+\frac{1}{(1-t_R)^2\sqrt{B}}\Big(\cos^2\left(C_R- (1-t_R)\sqrt{B}\right) \tan\left(C_R- (1-t)\sqrt{B}\right)\\  & -\frac{\sin\left(2(C_R- (1-t_R)\sqrt{B})\right)}{2}\Big).
\label{eq:c-minus-mid}
\end{align}

\item On \([0,t_L]\) (here \(\beta=0\)): Solving, \(\dot a=a^2\), \(\dot b=a\,b\), \(\dot c=b^2\), backward from \(t_L\), and requiring continuity with Eqs.~(\ref{eq:b-minus-mid}, \ref{eq:b-minus-mid},\ref{eq:c-minus-mid}) at $t_L$, yields
\[
a_t^{(-)} =\frac{a_{t_L}^{(-)}}{1+a_{t_L}^{(-)}(t_L-t)}, \ 
b_t^{(-)} =\frac{b_{t_L}^{(-)}}{1+a_{t_L}^{(-)}(t_L-t)}, \ 
c_t^{(-)} =c_{t_L}-\frac{(b_{t_L}^{(-)})^2}{1+a_{t_L}^{(-)}(t_L-t)}. 
\]
\end{itemize}

A sufficient  well-posedness of the Green function condition here  is $a_{t_L}^{(-)}>0$, thus resulting in
the same Eq.~(\ref{eq:nosing-plus}).

\end{document}